\documentclass{article}
\usepackage[preprint]{neurips_2024}
\usepackage[utf8]{inputenc} 
\usepackage[T1]{fontenc}    
\usepackage{hyperref}       
\usepackage{url}            
\usepackage{booktabs}       
\usepackage{amsfonts}       
\usepackage{nicefrac}       
\usepackage{microtype}      
\usepackage{xcolor}         
\usepackage{graphicx}
\usepackage{enumitem}
\usepackage{amsmath}        
\usepackage{amssymb}        
\usepackage[capitalise,nameinlink]{cleveref}
\usepackage{tcolorbox}      
\usepackage{wrapfig}
\usepackage{makecell}
\usepackage{fontawesome5}
\usepackage{float}  

\usepackage{tabularx}

\newcommand{\videoicon}{\raisebox{-0.2ex}{\normalsize\faIcon[regular]{file-video}}}
\newcommand{\audioicon}{\raisebox{-0.2ex}{\normalsize\faIcon[regular]{file-audio}}}
\newcommand{\docicon}{\raisebox{-0.2ex}{\normalsize\faIcon[regular]{file-alt}}}
\newcommand{\ppticon}{\raisebox{-0.2ex}{\normalsize\faIcon[regular]{file-powerpoint}}}

\title{A Definition of AGI}

\author{%
\textbf{Dan Hendrycks\textsuperscript{1}}, \textbf{Dawn Song\textsuperscript{2,3}}, \textbf{Christian Szegedy\textsuperscript{4}}, \textbf{Honglak Lee\textsuperscript{5,6}}, \textbf{Yarin Gal\textsuperscript{7}}, \\
\textbf{Erik Brynjolfsson\textsuperscript{8}}, \textbf{Sharon Li\textsuperscript{9}}, \textbf{Andy Zou\textsuperscript{1,10,11}}, \textbf{Lionel Levine\textsuperscript{12}}, \textbf{Bo Han\textsuperscript{13}}, \\
\textbf{Jie Fu\textsuperscript{14}}, \textbf{Ziwei Liu\textsuperscript{15}}, \textbf{Jinwoo Shin\textsuperscript{16}}, \textbf{Kimin Lee\textsuperscript{16}}, \textbf{Mantas Mazeika\textsuperscript{1}}, \\
\textbf{Long Phan\textsuperscript{1}}, \textbf{George Ingebretsen\textsuperscript{1}}, \textbf{Adam Khoja\textsuperscript{1}}, \textbf{Cihang Xie\textsuperscript{17}}, \\
\textbf{Olawale Salaudeen\textsuperscript{18}}, \textbf{Matthias Hein\textsuperscript{19}}, \textbf{Kevin Zhao\textsuperscript{20}}, \textbf{Alexander Pan\textsuperscript{2}}, \\
\textbf{David Duvenaud\textsuperscript{21,22}}, \textbf{Bo Li\textsuperscript{3,23}}, \textbf{Steve Omohundro\textsuperscript{24}}, \textbf{Gabriel Alfour\textsuperscript{25}}, \\
\textbf{Max Tegmark\textsuperscript{18}}, \textbf{Kevin McGrew\textsuperscript{26}}, \textbf{Gary Marcus\textsuperscript{27}}, \textbf{Jaan Tallinn\textsuperscript{28}}, \\
\textbf{Eric Schmidt\textsuperscript{18}}, \textbf{Yoshua Bengio\textsuperscript{29,30}} \\
\\
\textsuperscript{1}Center for AI Safety \quad
\textsuperscript{2}University of California, Berkeley \quad
\textsuperscript{3}Virtue AI \\
\textsuperscript{4}Morph Labs \quad
\textsuperscript{5}University of Michigan \quad
\textsuperscript{6}LG AI Research \\
\textsuperscript{7}University of Oxford \quad
\textsuperscript{8}Stanford University \quad
\textsuperscript{9}University of Wisconsin–Madison \\
\textsuperscript{10}Gray Swan AI \quad
\textsuperscript{11}Carnegie Mellon University \quad
\textsuperscript{12}Cornell University \\
\textsuperscript{13}Hong Kong Baptist University \quad
\textsuperscript{14}HKUST \quad
\textsuperscript{15}Nanyang Technological University \\
\textsuperscript{16}KAIST \quad
\textsuperscript{17}University of California, Santa Cruz \quad
\textsuperscript{18}Massachusetts Institute of Technology \\
\textsuperscript{19}University of Tübingen \quad
\textsuperscript{20}University of Washington \quad
\textsuperscript{21}University of Toronto \\
\textsuperscript{22}Vector Institute \quad
\textsuperscript{23}University of Illinois Urbana-Champaign \quad
\textsuperscript{24}Beneficial AI Research \\
\textsuperscript{25}Conjecture \quad
\textsuperscript{26}Institute for Applied Psychometrics \quad
\textsuperscript{27}New York University \\
\textsuperscript{28}CSER \quad
\textsuperscript{29}Université de Montréal \quad
\textsuperscript{30}LawZero
}

\begin{document}
\maketitle

\begin{abstract}
The lack of a concrete definition for Artificial General Intelligence (AGI) obscures the gap between today's specialized AI and human-level cognition. This paper introduces a quantifiable framework to address this, defining AGI as matching the cognitive versatility and proficiency of a well-educated adult. To operationalize this, we ground our methodology in Cattell-Horn-Carroll theory, the most empirically validated model of human cognition. The framework dissects general intelligence into ten core cognitive domains—including reasoning, memory, and perception—and adapts established human psychometric batteries to evaluate AI systems. Application of this framework reveals a highly ``jagged'' cognitive profile in contemporary models. While proficient in knowledge-intensive domains, current AI systems have critical deficits in foundational cognitive machinery, particularly long-term memory storage. The resulting AGI scores (e.g., GPT-4 at 27\%, GPT-5 at 57\%) concretely quantify both rapid progress and the substantial gap remaining before AGI.
\end{abstract}


\section{Introduction}

Artificial General Intelligence (AGI) may become the most significant technological development in human history, yet the term itself remains frustratingly nebulous, acting as a constantly moving goalpost. As specialized AI systems master tasks once thought to require human intellect—from mathematics to art—the criteria for ``AGI'' continually shift. This ambiguity fuels unproductive debates, hinders discussions about how far AGI is, and ultimately obscures the gap between today's AI and AGI.

This paper provides a comprehensive, quantifiable framework to cut through the ambiguity. Our framework aims to concretely specify the informal definition:
\begin{quote}
    \textbf{AGI is an AI that can match or exceed the cognitive versatility and proficiency of a well-educated adult.}
\end{quote}
This definition emphasizes that general intelligence requires not just specialized performance in narrow domains, but the breadth (versatility) and depth (proficiency) of skills that characterize human cognition.

To operationalize this definition, we must look to the only existing example of general intelligence: humans. Human cognition is not a monolithic capability; it is a complex architecture composed of many distinct abilities honed by evolution. These abilities enable our remarkable adaptability and understanding of the world.

To systematically investigate whether AI systems possess this spectrum of abilities, we ground our approach in the Cattell-Horn-Carroll (CHC) theory of cognitive abilities \citep{chc_theory, mcgrew2009chc, schneider2018chc, mcgrew2023carroll, mcgrew2023psychometric}, the most empirically validated model of human intelligence. CHC theory is primarily derived from the synthesis of over a century of iterative factor analysis of diverse collections of cognitive ability tests. In the late 1990’s to 2000’s almost all major clinical, individually administered tests of human intelligence have iterated towards test revisions that were either explicitly or implicitly based on CHC model test design blueprints \citep{Keith2010CattellHornCarrollAA, schneider2018chc}. CHC theory provides a hierarchical taxonomic map of human cognition. It breaks down general intelligence into distinct broad abilities and numerous narrow abilities (such as induction, associative memory, or spatial scanning). Readers interested in the strengths and limitations of the CHC framework are directed to further scholarly discussions \citep{wasserman2019, canivez2019}.

\begin{wrapfigure}{r}{0.5\textwidth}
    \centering
    \includegraphics[width=0.5\textwidth]{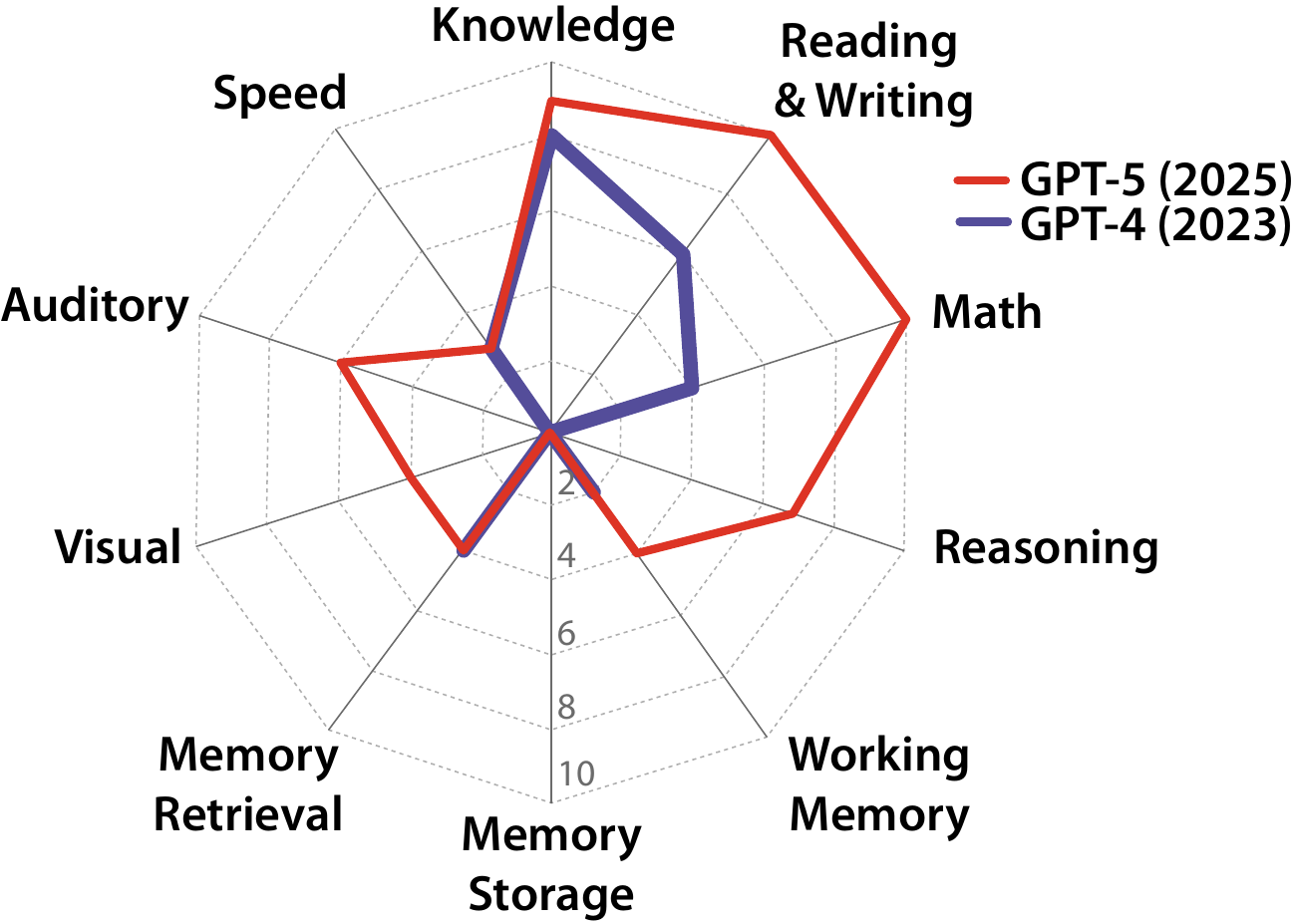}
    \caption{The capabilities of GPT-4 and GPT-5. Here GPT-5 answers questions in `Auto' mode.}
    \label{fig:gpt4_gpt5_capability}
\end{wrapfigure}

Decades of psychometric research have yielded a vast battery of tests specifically designed to isolate and measure these distinct cognitive components in individuals. Our framework adapts this methodology for AI evaluation. Instead of relying solely on generalized tasks that might be solved through compensatory strategies, we systematically investigate whether AI systems possess the underlying CHC narrow abilities that humans have. To determine whether an AI has the cognitive versatility and proficiency of a well-educated adult, we test the AI system with the gauntlet of cognitive batteries used to test people. This approach replaces nebulous concepts of intelligence with concrete measurements, resulting in a standardized ``AGI Score'' (0\% to 100\%), in which 100\% signifies AGI.

The application of this framework is revealing. By testing the fundamental abilities that underpin human cognition—many of which appear simple for humans—we find that contemporary AI systems can solve roughly half of these often-simple assessments. This indicates that despite impressive performance on complex benchmarks, current AI lacks many of the core cognitive capabilities essential for human-like general intelligence. Current AIs are narrower than well-educated humans overall but far smarter on some specific tasks.

The framework comprises ten core cognitive components, derived from CHC broad abilities and weighted equally (10\%) to prioritize breadth and cover major areas of cognition:
\begin{itemize}
    \item \textbf{General Knowledge (K):} The breadth of factual understanding of the world, encompassing commonsense, culture, science, social science, and history.
    \item \textbf{Reading and Writing Ability (RW):} Proficiency in consuming and producing written language, from basic decoding to complex comprehension, composition, and usage.
    \item \textbf{Mathematical Ability (M):} The depth of mathematical knowledge and skills across arithmetic, algebra, geometry, probability, and calculus.
    \item \textbf{On-the-Spot Reasoning (R):} The flexible control of attention to solve novel problems without relying exclusively on previously learned schemas, tested via deduction and induction.
    \item \textbf{Working Memory (WM):} The ability to maintain and manipulate information in active attention across textual, auditory, and visual modalities.
    \item \textbf{Long-Term Memory Storage (MS):} The capability to continually learn new information (associative, meaningful, and verbatim).
    \item \textbf{Long-Term Memory Retrieval (MR):} The fluency and precision of accessing stored knowledge, including the critical ability to avoid confabulation (hallucinations).
    \item \textbf{Visual Processing (V):} The ability to perceive, analyze, reason about, generate, and scan visual information.
    \item \textbf{Auditory Processing (A):} The capacity to discriminate, recognize, and work creatively with auditory stimuli, including speech, rhythm, and music.
    \item \textbf{Speed (S):} The ability to perform simple cognitive tasks quickly, encompassing perceptual speed, reaction times, and processing fluency.
\end{itemize}

This operationalization provides a holistic and multimodal (text, visual, auditory) assessment, serving as a rigorous diagnostic tool to pinpoint the strengths and profound weaknesses of current AI systems.

\begin{table}[h!]
    \centering
    \label{tab:agi_score_summary}
    \begin{tabularx}{\columnwidth}{@{}l *{11}{>{\centering\arraybackslash}X}@{}}
        \toprule
        \textbf{Model} & \textbf{K} & \textbf{RW} & \textbf{M} & \textbf{R} & \textbf{WM} & \textbf{MS} & \textbf{MR} & \textbf{V} & \textbf{A} & \textbf{S} & \textbf{Total} \\ 
        \midrule
        GPT-4 & 8\% & 6\% & 4\% & 0\% & 2\% & 0\% & 4\% & 0\% & 0\% & 3\% & \textbf{27\%} \\
        GPT-5 & 9\% & 10\% & 10\% & 7\% & 4\% & 0\% & 4\% & 4\% & 6\% & 3\% & \textbf{57\%} \\ 
        \bottomrule
    \end{tabularx}
    \vspace{2pt}
    \caption{AGI Score Summary for GPT-4 (2023) and GPT-5 (2025).}
\end{table}

\textbf{Scope.} Our definition is not an automatic evaluation nor a dataset. Rather, it specifies a large collection of well-scoped tasks that test specific cognitive abilities. Whether AIs can solve these tasks can be manually assessed by anyone, and people could supplement their testing using the best evaluations available at the time. This makes our definition more broad and more robust than fixed automatic AI capabilities datasets. Secondly, our definition focuses on capabilities frequently possessed by well-educated individuals, not a superhuman aggregate of all well-educated individuals' combined knowledge and skills. Therefore, our AGI definition is about human-level AI, not economy-level AI; we measure cognitive abilities rather than specialized economically valuable know-how, nor is our measurement a direct predictor of automation or economic diffusion. We leave economic measurements of advanced AI to other work. Last, we deliberately focus on core cognitive capabilities rather than physical abilities such as motor skills or tactile sensing, as we seek to measure the capabilities of the mind rather than the quality of its actuators or sensors. We discuss more limitations in the \hyperref[sec:discussion]{Discussion}.

\section{Overview of Abilities Needed for AGI}

This document outlines a framework for evaluating Artificial General Intelligence (AGI) by adopting and adapting the Cattell-Horn-Carroll (CHC) theory of human intelligence. The framework decomposes general intelligence into ten core cognitive components (broad abilities) and numerous narrow cognitive abilities. Solving all the tasks corresponding to these abilities implies an AGI Score of 100\%.\looseness=-1

Here is a comprehensive list of each cognitive ability.
\begin{enumerate}[leftmargin=*]
    \item \textbf{General Knowledge (K):} Knowledge that is familiar to most well-educated people or is important enough that most of them have been exposed to it.
    \begin{itemize}
        \item \textbf{Commonsense:} The vast set of shared, obvious background knowledge about how the world works.
        \item \textbf{Science:} Knowledge of the natural and physical sciences.
        \item \textbf{Social Science:} Understanding of human behavior, societies, and institutions.
        \item \textbf{History:} Knowledge of past events and objects.
        \item \textbf{Culture:} Cultural literacy and awareness.
    \end{itemize}

    \item \textbf{Reading and Writing Ability (RW):} Captures all of the declarative knowledge and procedural skills a person uses to consume and produce written language. 
    \begin{itemize}
        \item \textbf{Letter-Word Ability:} The ability to recognize letters and decode words.
        \item \textbf{Reading Comprehension:} The ability to understand connected discourse during reading.
        \item \textbf{Writing Ability:} The ability to write with clarity of thought, organization, and good sentence structure.
        \item \textbf{English Usage Knowledge:} Knowledge of writing in the English language with respect to capitalization, punctuation, usage, and spelling.
    \end{itemize}

    \item \textbf{Mathematical Ability (M):} The depth and breadth of mathematical knowledge and skills.
    \begin{itemize}
        \item \textbf{Arithmetic:} The manipulation of numbers using basic operations and solving word problems.
        \item \textbf{Algebra:} The study of symbols and the rules for combining them to express general relationships and solve equations.
        \item \textbf{Geometry:} The study of shapes, space, size, position, and distance.
        \item \textbf{Probability:} The quantification of uncertainty by assigning numbers from 0 to 1 to events.
        \item \textbf{Calculus:} The mathematics of change and accumulation.
    \end{itemize}

    \item \textbf{On-the-Spot Reasoning (R):} The deliberate but flexible control of attention to solve novel ``on the spot'' problems that cannot be performed by relying exclusively on previously learned habits, schemas, and scripts.
    \begin{itemize}
        \item \textbf{Deduction:} Reasoning from general statements or premises to reach a logically guaranteed conclusion.
        \item \textbf{Induction:} Discovering the underlying principles or rules that determine a phenomenon's behavior.
        \item \textbf{Theory of Mind:} Attributing mental states to others and understanding how those states may differ from one's own.
        \item \textbf{Planning:} Devising a sequence of actions to achieve a specific goal.
        \item \textbf{Adaptation:} The ability to infer unstated classification rules from a simple performance feedback sequence.
\end{itemize}

    \item \textbf{Working Memory (WM):} The ability to maintain, manipulate, and update information in active attention. (Often referred to as short-term memory.)
    \begin{itemize}
        \item \textbf{Textual Working Memory:} The ability to hold and manipulate sequences of verbal information presented textually.
            \begin{itemize}
                \item \textit{Recall:} The ability to remember a short sequence of elements (digits, letters, words, and nonsense words) and answer basic questions about them.
                \item \textit{Transformation Sequence:} The ability to remember and update a short list of digits or lists of digits following a sequence of operations.
            \end{itemize}
        \item \textbf{Auditory Working Memory:} The ability to hold and manipulate auditory information, including speech, sounds, and music.
            \begin{itemize}
                \item \textit{Recall:} The ability to remember a collection of voices, utterances, and sound effects and answer basic questions about them.
                \item \textit{Transformation Sequence:} The ability to remember and modify a short utterance with a variety of transformations.
            \end{itemize}
        \item \textbf{Visual Working Memory:} The ability to hold and manipulate visual information, including images, scenes, spatial layouts, and video.
            \begin{itemize}
                \item \textit{Recall:} The ability to remember a collection of images and answer basic questions about them.
                \item \textit{Transformation Sequence:} The ability to transform a visual input following a sequence of operations.
                \item \textit{Spatial Navigation Memory:} The ability to represent a sense of location in an environment.
                \item \textit{Long Video Q\&A:} The ability to watch a long video or a movie and answer basic questions about it.
            \end{itemize}
        \item \textbf{Cross-Modal Working Memory:} The ability to maintain and modify information presented across different modalities.
    \end{itemize}

    \item \textbf{Long-Term Memory Storage (MS):} The ability to stably acquire, consolidate, and store new information from recent experiences.
    \begin{itemize}
        \item \textbf{Associative Memory:} The ability to link previously unrelated pieces of information.
            \begin{itemize}
                \item \textit{Cross-Modal Association:} The ability to form a link between two previously unrelated stimuli, such that the subsequent presentation of one of the stimuli serves to activate the recall of the other stimuli. 
                \item \textit{Personalization Adherence:}The ability to associate specific rules, preferences, or corrections with a distinct interaction context and apply them consistently and unprompted over time.
                \item \textit{Procedural Association:} The ability to acquire and retain a sequence of associated steps or rules (a procedure) and execute them when cued with the name of the procedure.
            \end{itemize}
        \item \textbf{Meaningful Memory:} The ability to remember narratives and other forms of semantically related information.
            \begin{itemize}
                \item \textit{Story Recall:} The ability to remember the gist of stories.
                \item \textit{Movie Recall:} The ability to remember the gist of movies.
                \item \textit{Episodic Context Recall:} The ability to remember specific events or experiences, including their context (the ``what, where, when, and how'').
            \end{itemize}
        \item \textbf{Verbatim Memory:} The ability to recall information exactly as it was presented, requiring precise encoding of specific sequences, sets, or designs, often independent of the information's meaning.
            \begin{itemize}
                \item \textit{Short Sequence Recall:} The ability to exactly recall short sequences of text after a delay.
                \item \textit{Set Recall:} The ability to recall a set (the order of recall does not matter).
                \item \textit{Design Recall:} The ability to recall the spatial arrangement and structure of visual information.
            \end{itemize}
    \end{itemize}

    \item \textbf{Long-Term Memory Retrieval (MR):} The fluency and precision with which individuals can access long-term memory. 
    \begin{itemize}
        \item \textbf{Retrieval Fluency (Fluency):} The speed and ease of generating ideas, associations, and solutions based on stored knowledge.
            \begin{itemize}
                \item \textit{Ideational Fluency:} This is the ability to rapidly produce a series of ideas, words, or phrases related to a specific condition, category, or object.
                \item \textit{Expressional Fluency:} This is the ability to rapidly think of different ways of expressing an idea.
                \item \textit{Alternative Solution Fluency:} This is the ability to rapidly think of several alternative solutions to a practical problem.
                \item \textit{Word Fluency:} This is the ability to rapidly produce words that share a non-semantic feature.
                \item \textit{Naming Facility:} This is the ability to rapidly call common objects by their names.
                \item \textit{Figural Fluency:} This is the ability to rapidly draw or sketch as many things as possible.
            \end{itemize}
        \item \textbf{Retrieval Precision (Hallucinations):} The accuracy of accessed knowledge, including the critical ability to avoid confabulation (hallucinations).
    \end{itemize}

    \item \textbf{Visual Processing (V):} The ability to analyze and generate natural and unnatural images and videos. 
    \begin{itemize}
        \item \textbf{Perception:} The ability to accurately interpret and understand visual input.
            \begin{itemize}
                \item \textit{Image recognition:} The ability to classify images of commonplace objects, places, or facial expressions including distorted images. 
                \item \textit{Image captioning:} The ability to generate a concise, human-like text description for the visual content of an image. 
                \item \textit{Image anomaly detection:} includes detecting whether there is something anomalous in an image, or what is missing from an object.
                \item \textit{Clip captioning:} The ability to generate a concise, human-like text description of a short video segment. 
                \item \textit{Video anomaly detection:} The ability to detect whether a short video segment is anomalous or implausible.
            \end{itemize}
        \item \textbf{Visual Generation:} The ability to synthesize images and short videos.
        \item \textbf{Visual Reasoning:} The ability to solve problems and make inferences using spatial logic and visual abstractions.
        \item \textbf{Spatial Scanning:} The speed and accuracy of visually exploring a complex field.
    \end{itemize}

    \item \textbf{Auditory Processing (A):} The ability to discriminate, remember, reason, and work creatively on auditory stimuli, which may consist of tones and speech units.
    \begin{itemize}
        \item \textbf{Phonetic Coding:} The ability to hear phonemes distinctly, blend sounds into words, and segment words into parts, sounds, or phonemes.
        \item \textbf{Speech Recognition:} The ability to transcribe a spoken audio signal into its corresponding sequence of text.
        \item \textbf{Voice:} The quality and responsiveness of the AI's synthesized voice.
            \begin{itemize}
                \item \textit{Natural speech:} The ability to utter sentences or paragraphs that sound natural and not robotic. 
                \item \textit{Natural conversation:} The ability to maintain conversational fluidity without long delays or excessive interruptions.
            \end{itemize}
        \item \textbf{Rhythmic Ability:} The ability to recognize and maintain a musical beat, including reproducing rhythms, detecting differences between rhythms, and synchronizing by playing or humming along.
        \item \textbf{Musical Judgment:} The ability to discriminate and judge simple patterns in music.
    \end{itemize}

    \item \textbf{Speed (S):} The ability to perform simple cognitive tasks quickly.
    \begin{itemize}
        \item \textbf{Perceptual Speed–Search:} The speed of scanning a visual or textual field to find specific targets.
        \item \textbf{Perceptual Speed–Compare:} The speed of comparing two or more stimuli to identify similarities or differences.
        \item \textbf{Reading Speed:} The rate at which text can be processed with full comprehension.
        \item \textbf{Writing Speed:} The rate at which text can be generated or copied.
        \item \textbf{Number Facility:} The speed and accuracy of performing basic arithmetic operations.
        \item \textbf{Simple Reaction Time:} The time taken to respond to a single, anticipated stimulus.
        \item \textbf{Choice Reaction Time:} The time taken to respond correctly when presented with one of several possible stimuli.
        \item \textbf{Inspection Time:} The speed at which subtle differences between visual or auditory stimuli can be perceived.
        \item \textbf{Comparison Speed:} The time taken to make a judgment comparing two stimuli on a specific attribute.
        \item \textbf{Pointer Fluency:} The speed and accuracy of moving a pointer, such as a virtual mouse.
    \end{itemize}
\end{enumerate}

\begin{figure}[h]
    \centering
    \includegraphics[width=1.0\linewidth]{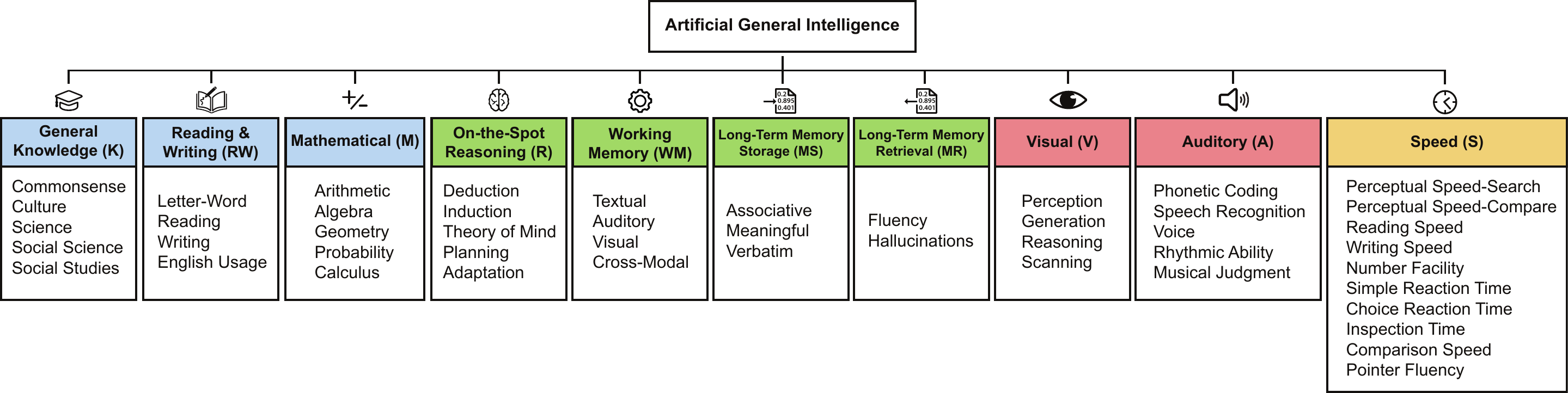}
    \caption{The ten core cognitive components of our AGI definition.}
    \label{fig:AGI_tree_diagram}
\end{figure}

\Cref{fig:AGI_tree_diagram} summarizes the broad and narrow capabilities tested.
\newpage
\section{General Knowledge (K)}
\label{sec:knowledge}

\begin{figure}[H]
    \centering
    \vspace{-20pt}
    \includegraphics[width=\textwidth]{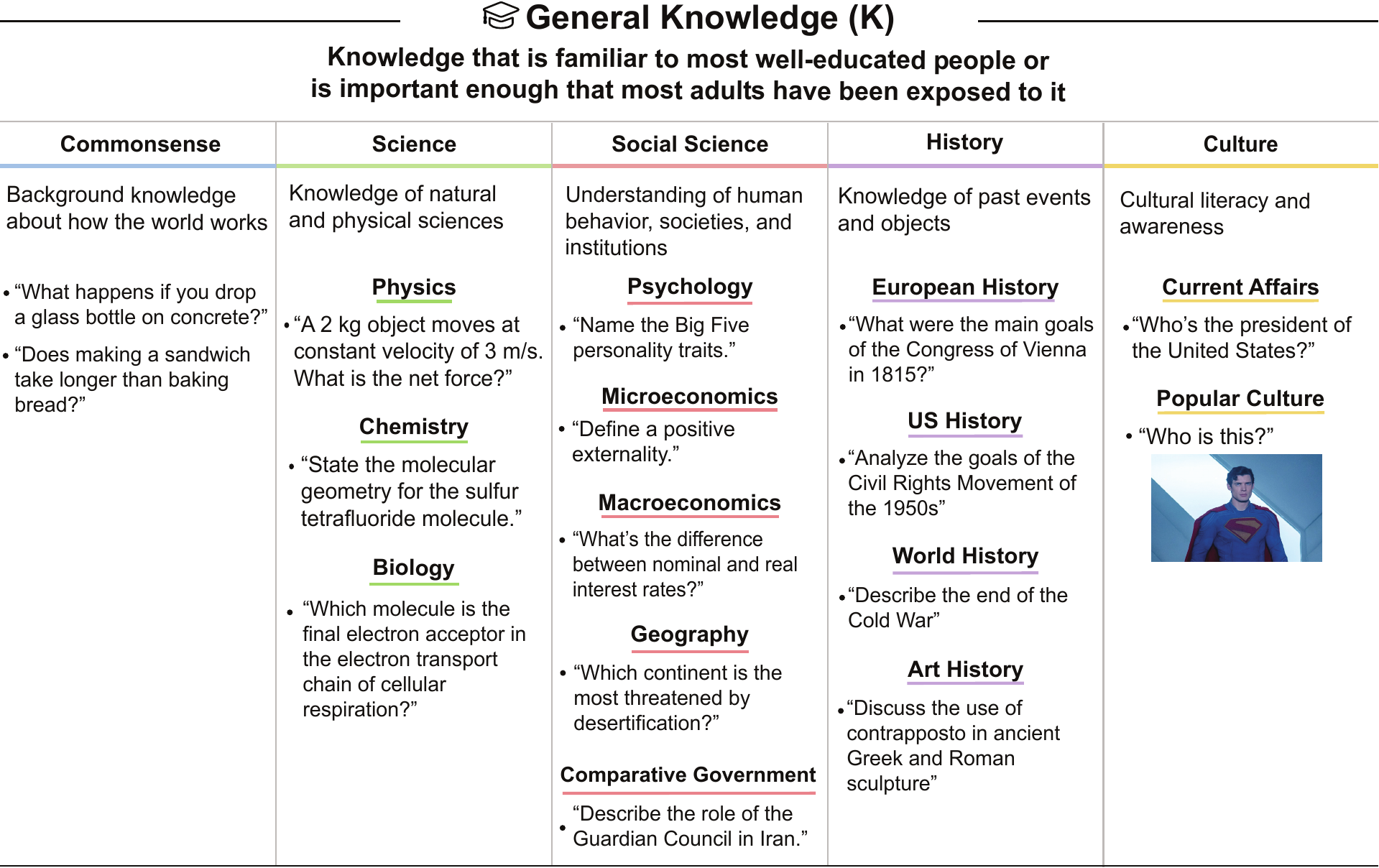}
    \vspace{-20pt}
    \label{fig:k_examples}
\end{figure}

\paragraph{Assessment Details.} See \Cref{app:knowledge} for further details on how to assess general knowledge capabilities concretely.

\paragraph{AI System Performance.} The table summarizes current AI system performance on General Knowledge (K) tasks. GPT-4 has substantial general knowledge, and GPT-5 partially fills in its remaining gaps.


\begin{table}[h!]
    \centering
    \label{tab:knowledge_scores}
    \begin{tabular*}{\columnwidth}
    {@{\extracolsep{\fill}}lcccccc@{}}
        \toprule
        \textbf{Model} &
        \makecell{\textbf{Commonsense} \\ \textbf{(2\%)}} &
        \makecell{\textbf{Science} \\ \textbf{(2\%)}} &
        \makecell{\textbf{Social} \\ \textbf{Science (2\%)}} &
        \makecell{\textbf{History} \\ \textbf{(2\%)}} &
        \makecell{\textbf{Culture} \\ \textbf{(2\%)}} &
        \textbf{Total} \\
        \midrule
        GPT-4 & 2\% & 2\% & 2\% & 2\% & 0\% & \textbf{8\%} \\
        GPT-5 & 2\% & 2\% & 2\% & 2\% & 1\% & \textbf{9\%} \\
        \bottomrule
    \end{tabular*}
\end{table}
\section{Reading and Writing Ability (RW)}
\label{sec:rw}

\begin{figure}[H]
    \centering
    \vspace{-10pt}
    \includegraphics[width=\linewidth]{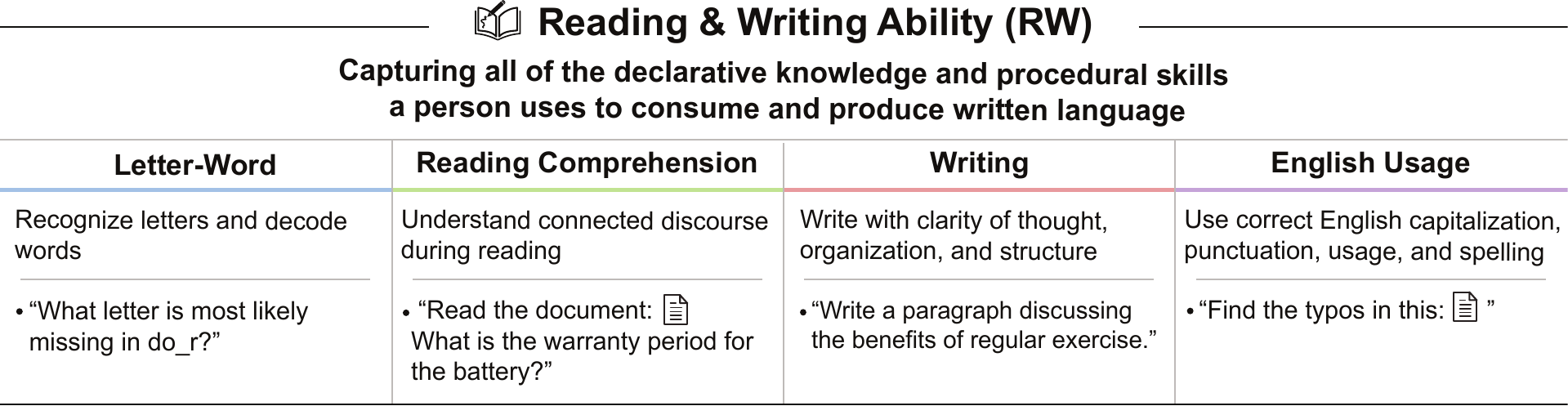}
    \vspace{-25pt}
    \label{fig:rw_examples}
\end{figure}

\paragraph{Assessment Details.} See \Cref{app:rw} for further details on how to assess reading and writing capabilities concretely.

\paragraph{AI System Performance.} The table summarizes current AI system performance on Reading and Writing Ability (RW) tasks. GPT-4's difficulty with token-level understanding, its small context window, and its imprecise working memory limit its ability to analyze substrings of words, to read long documents, and to carefully proofread text. GPT-5 addresses these issues.


\begin{table}[h!]
    \centering
    \label{tab:rw_scores}
    \begin{tabular*}{\columnwidth}{@{\extracolsep{\fill}}lccccc@{}}
        \toprule
        \textbf{Model} &
        \textbf{Letters (1\%)} &
        \textbf{Reading (3\%)} &
        \textbf{Writing (3\%)} &
        \textbf{Usage (3\%)} &
        \textbf{Total} \\
        \midrule
        GPT-4 & 0\% & 2\% & 3\% & 1\% & \textbf{6\%} \\
        GPT-5 & 1\% & 3\% & 3\% & 3\% & \textbf{10\%} \\
        \bottomrule
    \end{tabular*}
\end{table}
\section{Mathematical Ability (M)}
\label{sec:math}

\begin{figure}[H]
    \centering
    \vspace{-10pt}
    \includegraphics[width=\textwidth]{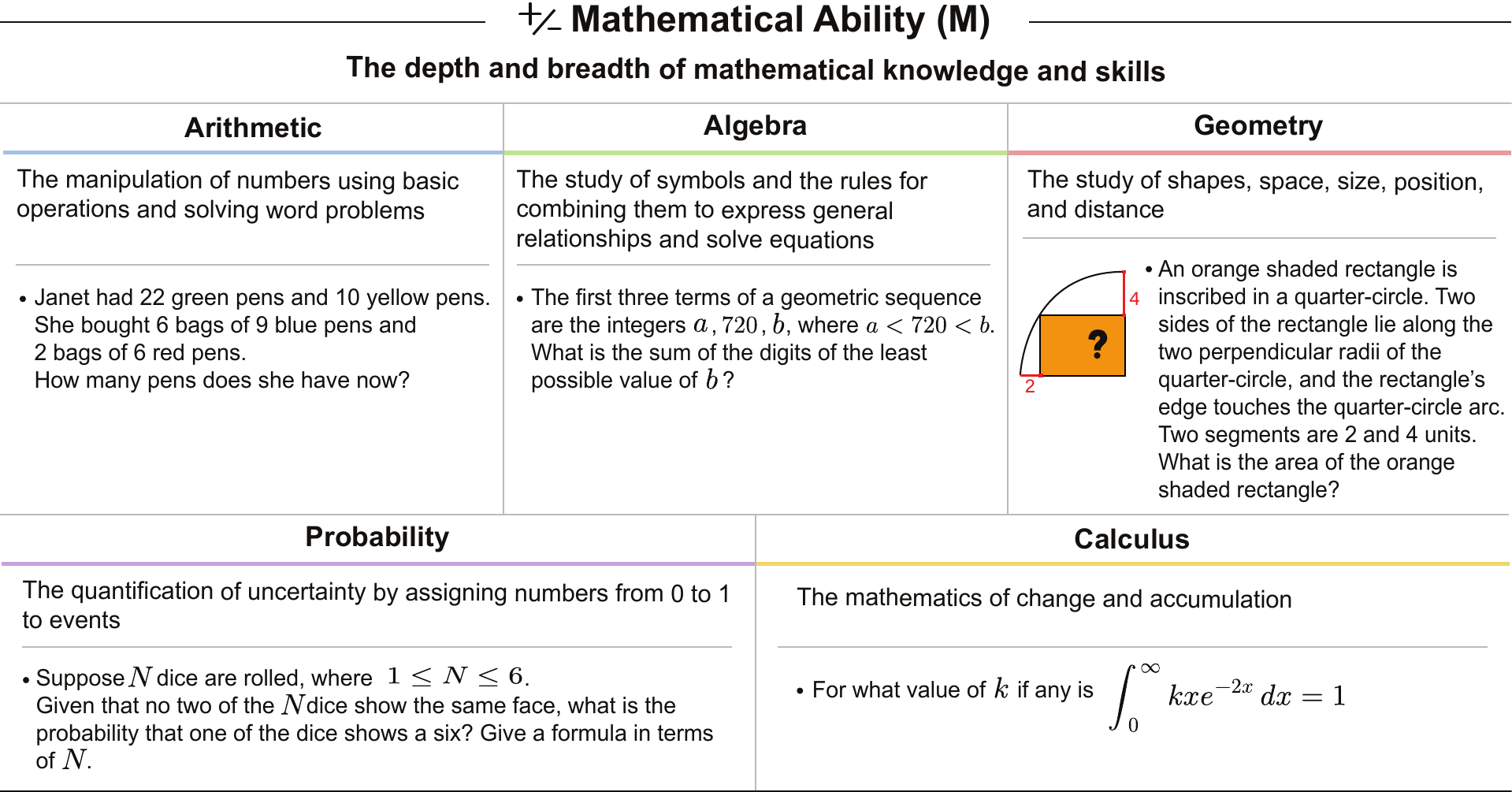}
    \vspace{-15pt}
    \label{fig:math}
\end{figure}
\paragraph{Assessment Details.} See \Cref{app:math} for further details on how to assess mathematical capabilities concretely.

\paragraph{AI System Performance.} The table summarizes current AI system performance on Mathematical Ability (M) tasks. GPT-4 has limited mathematical capabilities, while GPT-5 has exceptional mathematical capabilities.


\begin{table}[h]
    \centering
    \label{tab:math_scores}
    \begin{tabular*}{\columnwidth}{@{\extracolsep{\fill}}lcccccc@{}}
        \toprule
        \textbf{Model} &
        \makecell{\textbf{Arithmetic} \\ \textbf{(2\%)}} &
        \makecell{\textbf{Algebra} \\ \textbf{(2\%)}} &
        \makecell{\textbf{Geometry} \\ \textbf{(2\%)}} &
        \makecell{\textbf{Probability} \\ \textbf{(2\%)}} &
        \makecell{\textbf{Calculus} \\ \textbf{(2\%)}} &
        \textbf{Total} \\
        \midrule
        GPT-4 & 2\% & 1\% & 0\% & 1\% & 0\% & \textbf{4\%} \\
        GPT-5 & 2\% & 2\% & 2\% & 2\% & 2\% & \textbf{10\%} \\
        \bottomrule
    \end{tabular*}
\end{table}
\section{On-the-Spot Reasoning (R)}
\label{sec:reasoning}

\begin{figure}[h]
    \centering
    \vspace{-10pt}
    \includegraphics[width=\textwidth]{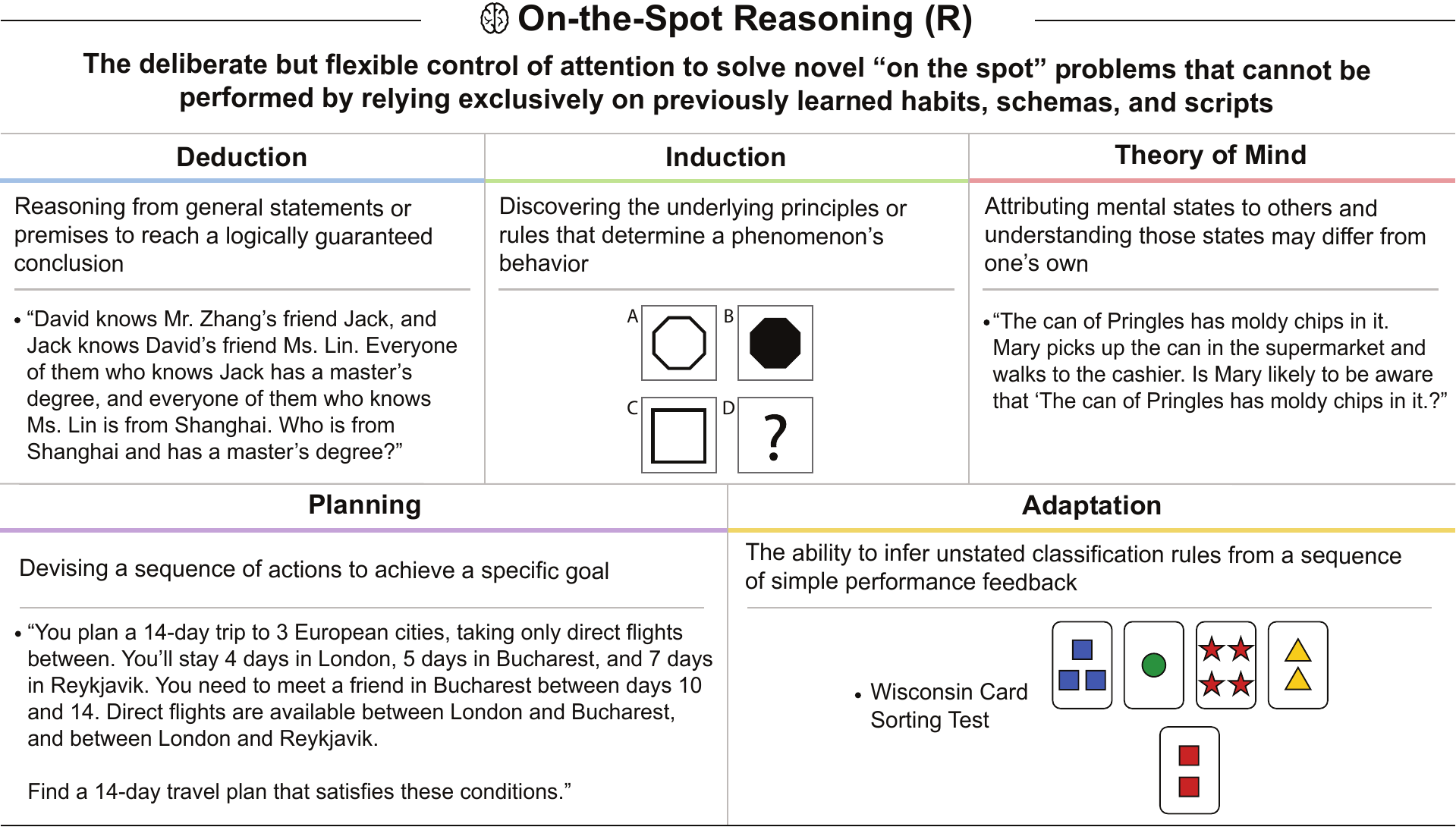}
    \vspace{-20pt}
    \label{fig:r_examples}
\end{figure}

\paragraph{Assessment Details.} See \Cref{app:reasoning} for further details on how to assess on-the-spot reasoning capabilities concretely.

\paragraph{AI System Performance.} The table summarizes current AI system performance on On-the-Spot Reasoning (R) tasks. GPT-4 has negligible on-the-spot reasoning capabilities, while GPT-5 only has some remaining gaps.


\begin{table}[h!]
    \centering
    \label{tab:reasoning_scores}
    \begin{tabular*}{\columnwidth}{@{\extracolsep{\fill}}lcccccc@{}}
        \toprule
        \textbf{Model} &
        \makecell{\textbf{Deduction} \\ \textbf{(2\%)}} &
        \makecell{\textbf{Induction} \\ \textbf{(4\%)}} &
        \makecell{\textbf{Theory of} \\ \textbf{Mind (2\%)}} &
        \makecell{\textbf{Planning} \\ \textbf{(1\%)}} &
        \makecell{\textbf{Adaptation} \\ \textbf{(1\%)}} &
        \textbf{Total} \\
        \midrule
        GPT-4 & 0\% & 0\% & 0\% & 0\% & 0\% & \textbf{0\%} \\
        GPT-5 & 2\% & 2\% & 2\% & 1\% & 0\% & \textbf{7\%} \\
        \bottomrule
    \end{tabular*}
\end{table}
\section{Working Memory (WM)}
\label{sec:wm}

\begin{figure}[H]
    \centering
    \vspace{-15pt}
    \includegraphics[width=1.\linewidth]{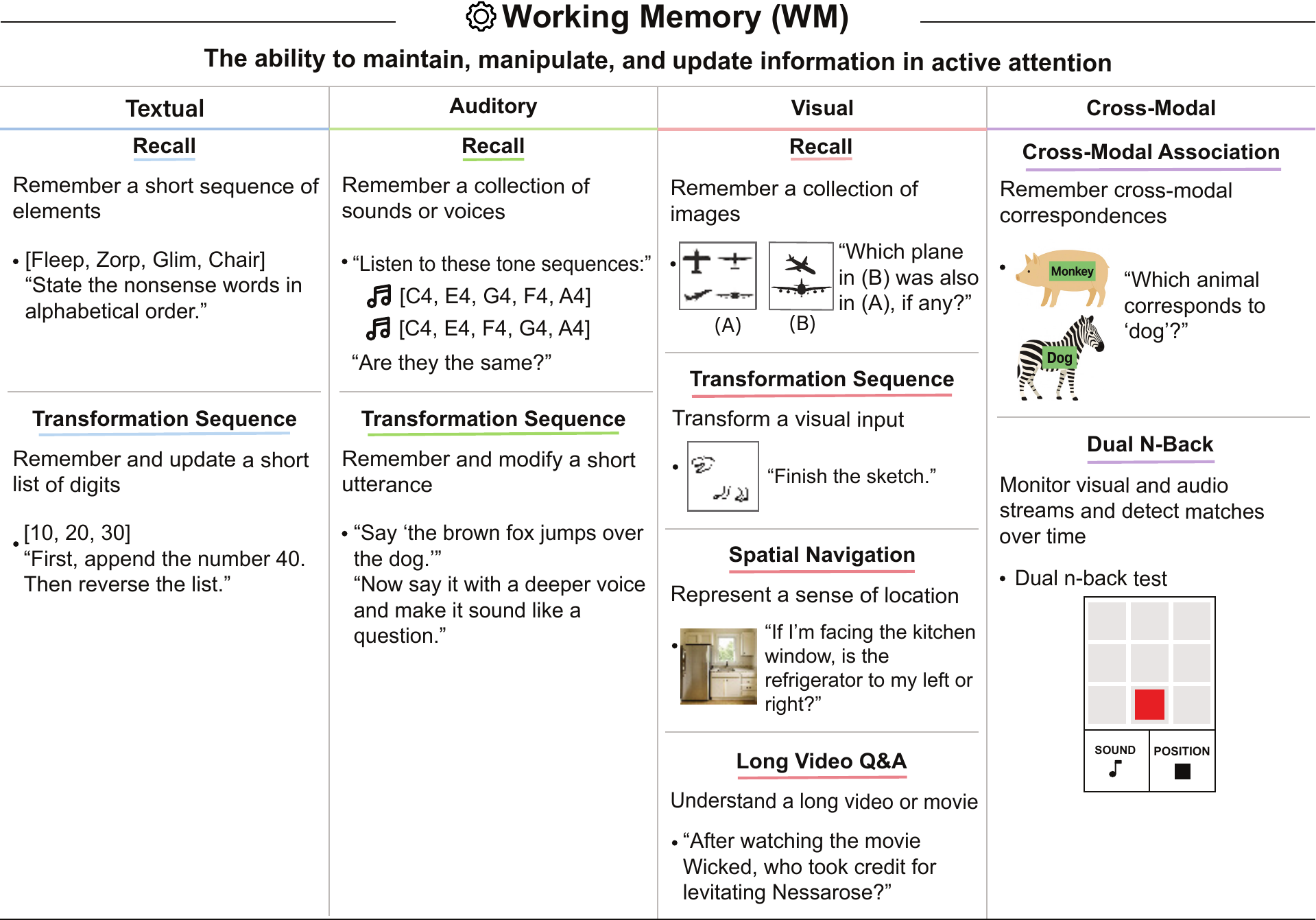}
    \vspace{-30pt}
    \label{fig:wm_examples}
\end{figure}

\paragraph{Assessment Details.} See \Cref{app:wm} for further details on how to assess working memory capabilities concretely.

\paragraph{AI System Performance.} The table summarizes current AI system performance on Working Memory (WM) tasks. While the raw Textual Working Memory score appears similar between GPT-4 and GPT-5 in this battery, improvements in managing long contexts are also reflected in the Document Level Reading Comprehension score within the Reading and Writing (RW) ability.


\begin{table}[h!]
    \centering
    \label{tab:wm_scores}
    \begin{tabular*}{\columnwidth}{@{\extracolsep{\fill}}lccccc@{}}
        \toprule
        \textbf{Model} &
        \textbf{Textual (2\%)} &
        \textbf{Auditory (2\%)} &
        \textbf{Visual (4\%)} &
        \textbf{Cross-Modal (2\%)} &
        \textbf{Total} \\
        \midrule
        GPT-4 & 2\% & 0\% & 0\% & 0\% & \textbf{2\%} \\
        GPT-5 & 2\% & 0\% & 1\% & 1\% & \textbf{4\%} \\
        \bottomrule
    \end{tabular*}
\end{table}
\section{Long-Term Memory Storage (MS)}
\label{sec:ms}

\begin{figure}[H]
    \centering
    \vspace{-15pt}
    \includegraphics[width=1.0\linewidth]{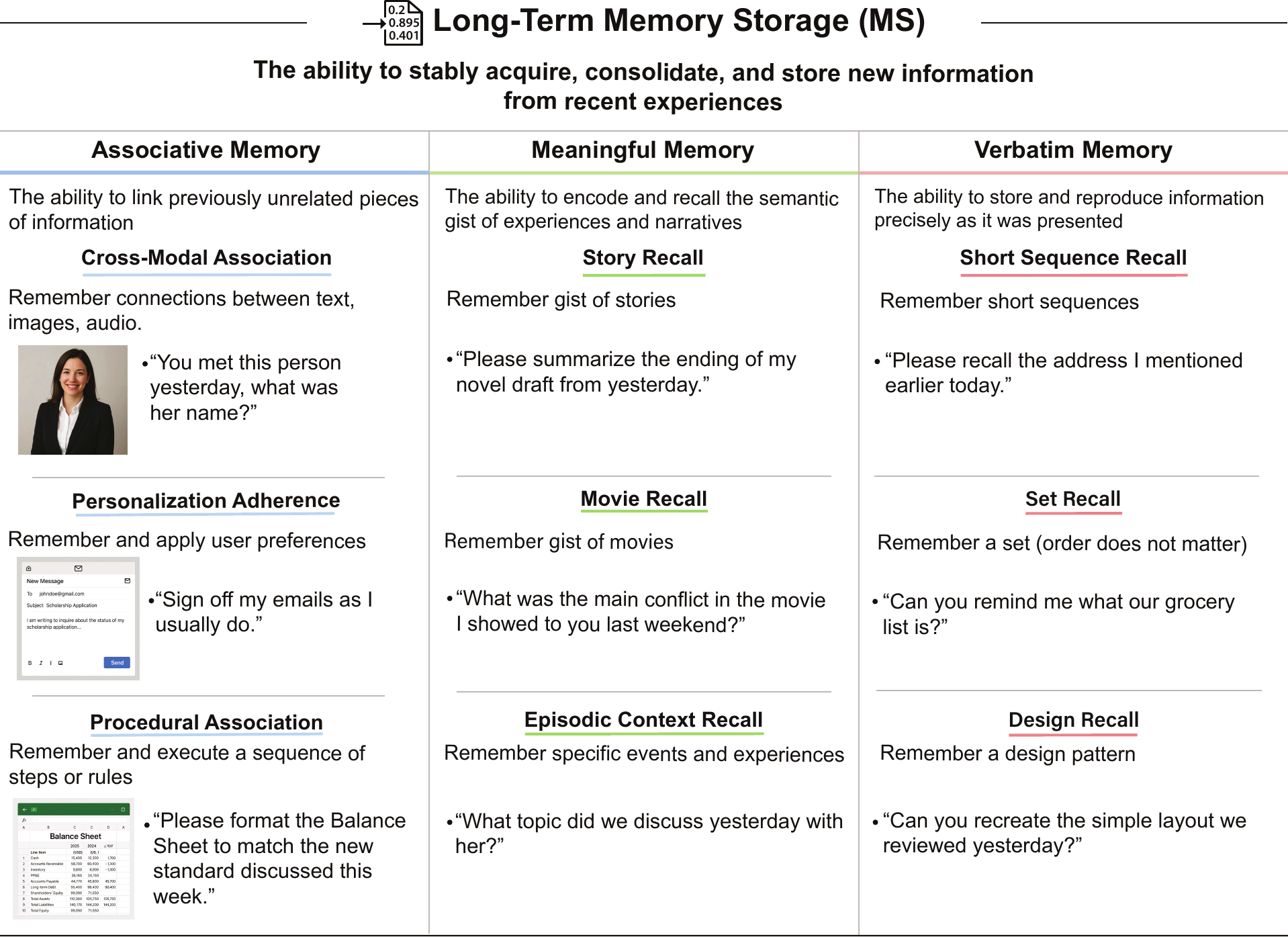}
    \vspace{-15pt}
    \label{fig:ms_examples}
\end{figure}

\paragraph{Assessment Details.} See \Cref{app:ms} for further details on how to assess long-term memory storage capabilities concretely.

\paragraph{AI System Performance.} The table summarizes current AI system performance on Long-Term Memory Storage (MS) tasks. Both GPT-4 and GPT-5 lack appreciable long-term memory storage capabilities. 


\begin{table}[h!]
    \centering
    \label{tab:ms_scores}
    \begin{tabular*}{\columnwidth}{@{\extracolsep{\fill}}lcccc@{}}
        \toprule
        \textbf{Model} &
        \textbf{Associative (4\%)} &
        \textbf{Meaningful (3\%)} &
        \textbf{Verbatim (3\%)} &
        \textbf{Total} \\
        \midrule
        GPT-4 & 0\% & 0\% & 0\% & \textbf{0\%} \\
        GPT-5 & 0\% & 0\% & 0\% & \textbf{0\%} \\
        \bottomrule
    \end{tabular*}
\end{table}
\section{Long-Term Memory Retrieval (MR)}
\label{sec:mr}

\begin{figure}[h]
    \centering
    \vspace{-10pt}
    \includegraphics[width=\textwidth]{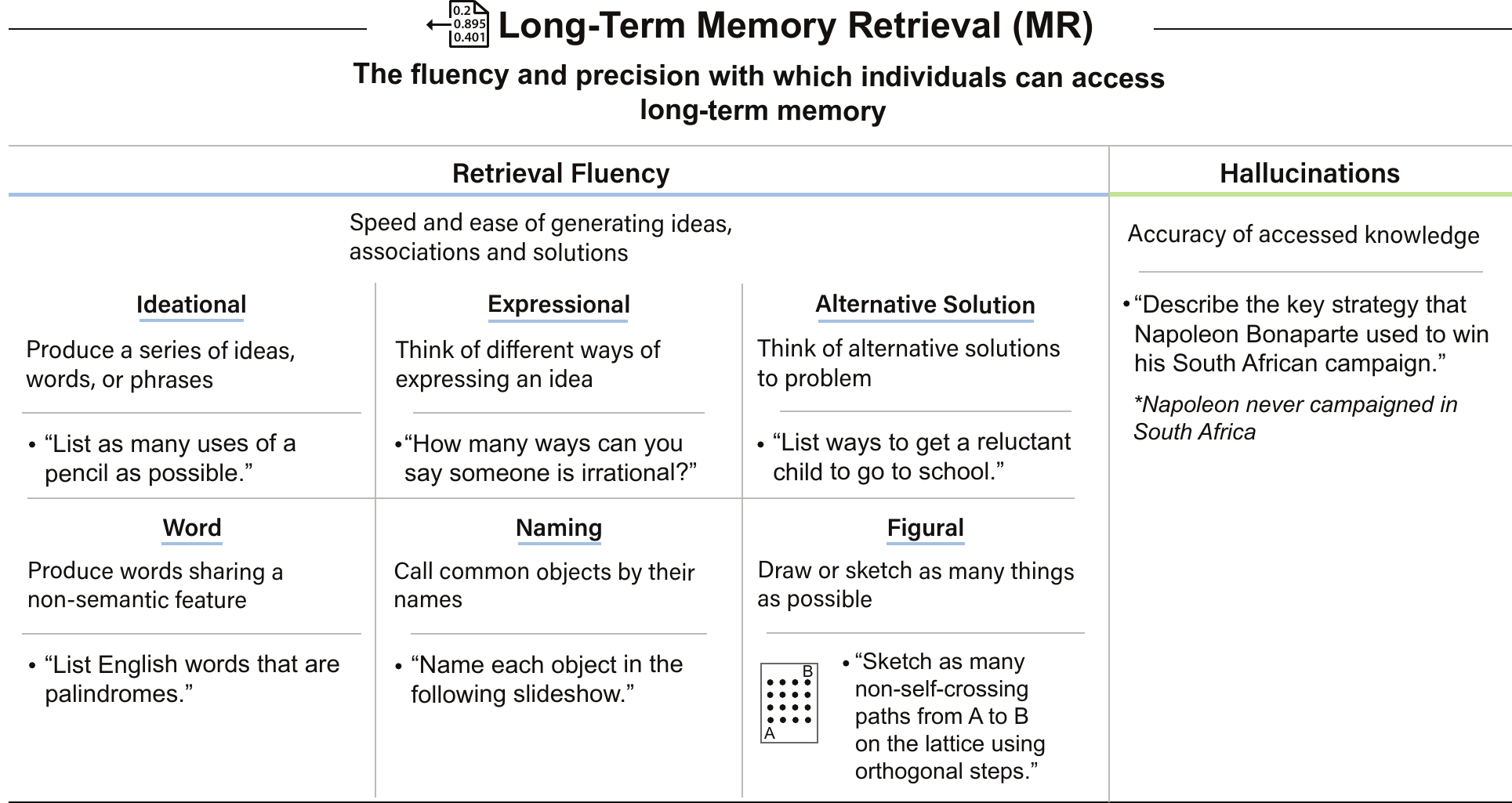}
    \vspace{-15pt}
    \label{fig:mr_examples}
\end{figure}

\paragraph{Assessment Details.} See \Cref{app:mr} for further details on how to assess long-term memory retrieval capabilities concretely.

\paragraph{AI System Performance.} The table summarizes current AI system performance on Long-Term Memory Retrieval (MR) tasks.  Both GPT-4 and GPT-5 can rapidly retrieve many concepts from their parameters, but they both frequently hallucinate.


\begin{table}[h!]
    \centering
    \label{tab:mr_scores}
    \begin{tabular*}{\columnwidth}{@{\extracolsep{\fill}}lccc@{}}
        \toprule
        \textbf{Model} &
        \textbf{Fluency (6\%)} &
        \textbf{Hallucinations (4\%)} &
        \textbf{Total} \\
        \midrule
        GPT-4 & 4\% & 0\% & \textbf{4\%} \\
        GPT-5 & 4\% & 0\% & \textbf{4\%} \\
        \bottomrule
    \end{tabular*}
\end{table}
\section{Visual Processing (V)}
\label{sec:visual}

\begin{figure}[H]
    \centering
    \vspace{-10pt}
    \includegraphics[width=\textwidth]{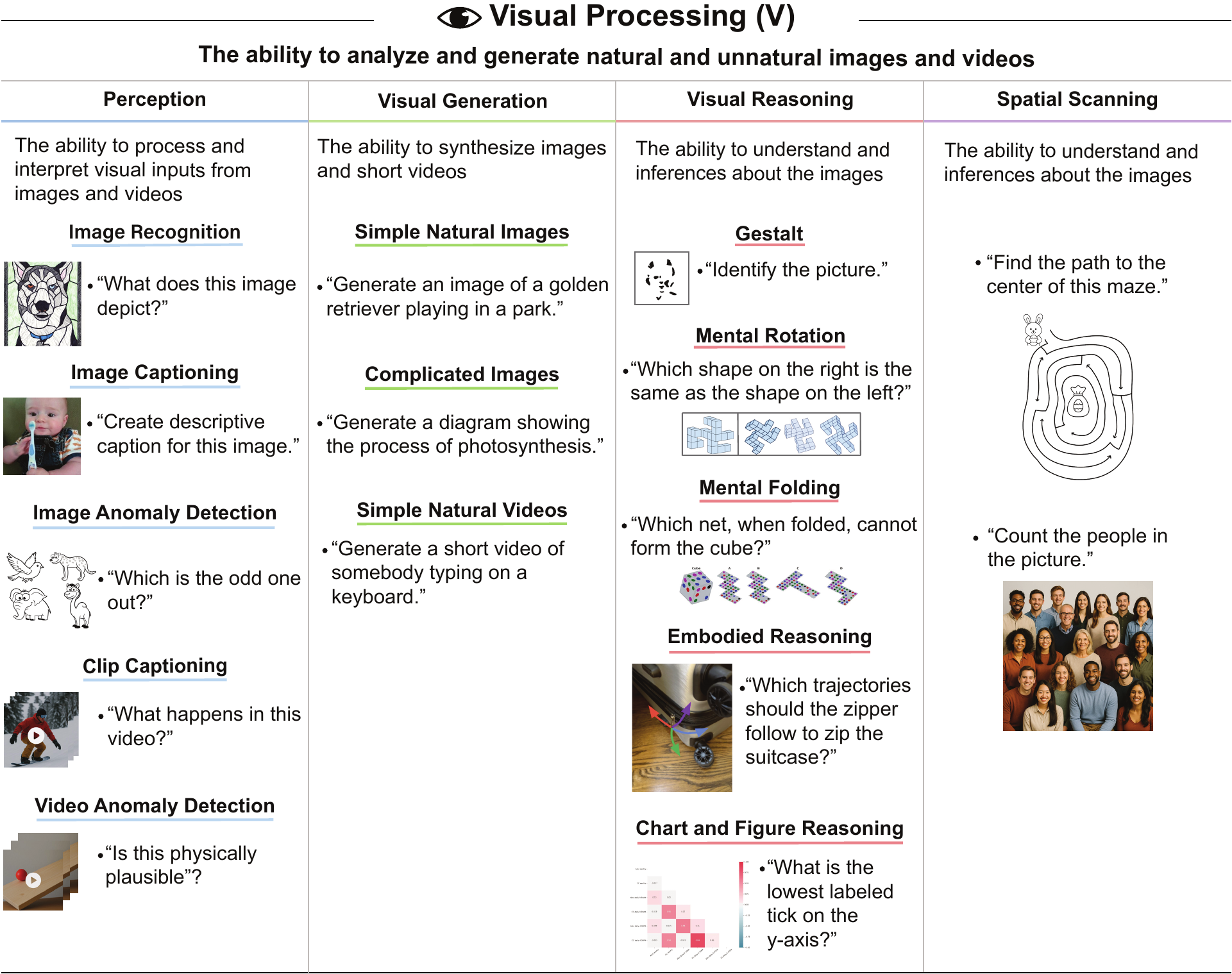}
    \vspace{-15pt}
    \label{fig:v_examples}
\end{figure}

\paragraph{Assessment Details.} See \Cref{app:visual} for further details on how to assess visual processing capabilities concretely.

\paragraph{AI System Performance.} The table summarizes current AI system performance on Visual Processing (V) tasks. GPT-4 had no ability to perceive or generate images, while GPT-5 has appreciable but highly incomplete visual processing capabilities.


\begin{table}[h!]
    \centering
    \begin{tabular*}{\columnwidth}{@{\extracolsep{\fill}}lccccc@{}}
        \toprule
        \textbf{Model} &
        \makecell{\textbf{Perception} \\ \textbf{(4\%)}} &
        \makecell{\textbf{Generation} \\ \textbf{(3\%)}} &
        \makecell{\textbf{Reasoning} \\ \textbf{(2\%)}} &
        \makecell{\textbf{Spatial Scanning} \\ \textbf{(1\%)}} &
        \textbf{Total} \\
        \midrule
        GPT-4 & 0\% & 0\% & 0\% & 0\% & \textbf{0\%} \\
        GPT-5 & 2\% & 2\% & 0\% & 0\% & \textbf{4\%} \\
        \bottomrule
    \end{tabular*}
    \label{tab:visual_scores}
\end{table}
\section{Auditory Processing (A)}
\label{sec:auditory}

\begin{figure}[H]
    \centering
    \vspace{-10pt}
    \includegraphics[width=1.0\linewidth]{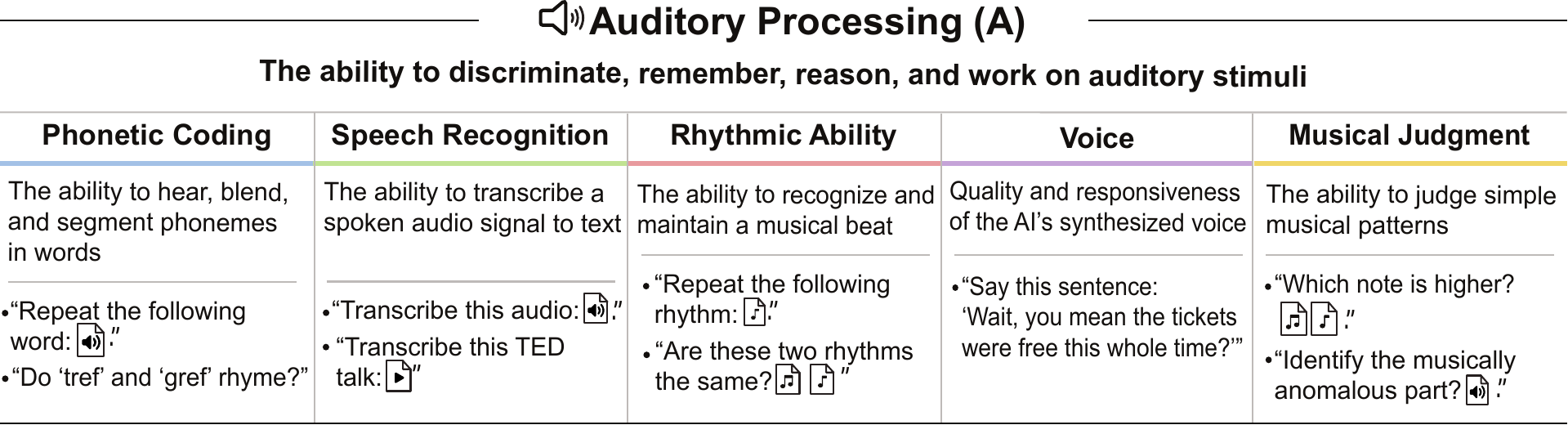}
    \vspace{-15pt}
    \label{fig:a_examples}
\end{figure}

\paragraph{Assessment Details.} See \Cref{app:auditory} for further details on how to assess auditory processing capabilities concretely.

\paragraph{AI system performance.} The table summarizes current AI system performance on Auditory Processing (A) tasks. GPT-4 had no audio processing capabilities, while GPT-5's capabilities are appreciable but incomplete.


\begin{table}[h!]
    \centering
    \label{tab:auditory_scores}
    \begin{tabular*}{\columnwidth}{@{\extracolsep{\fill}}lcccccc@{}}
        \toprule
        \textbf{Model} & 
        \makecell{\textbf{Phonetic} \\ \textbf{(1\%)}} & 
        \makecell{\textbf{Speech} \\ \textbf{Recognition (4\%)}} & 
        \makecell{\textbf{Voice} \\ \textbf{(3\%)}} & 
        \makecell{\textbf{Rhythmic} \\ \textbf{(1\%)}} & 
        \makecell{\textbf{Musical} \\ \textbf{(1\%)}} & 
        \textbf{Total} \\ 
        \midrule
        GPT-4 & 0\% & 0\% & 0\% & 0\% & 0\% & \textbf{0\%} \\
        GPT-5 & 0\% & 4\% & 2\% & 0\% & 0\% & \textbf{6\%} \\ 
        \bottomrule
    \end{tabular*}
\end{table}
\section{Speed (S)}
\label{sec:speed}

\begin{figure}[H]
    \centering
    \vspace{-10pt}
    \includegraphics[width=\linewidth]{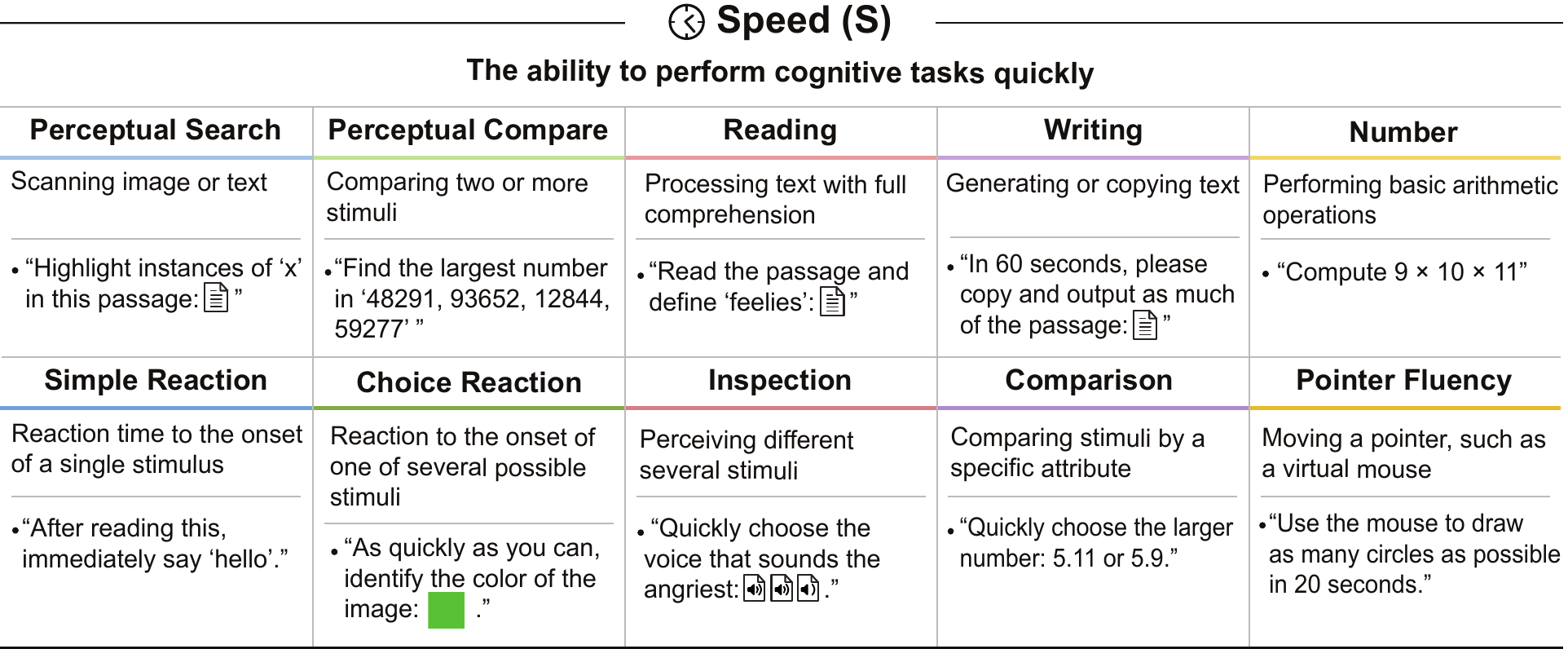}
    \vspace{-15pt}
    \label{fig:s_examples}
\end{figure}

\paragraph{Assessment Details.} See \Cref{app:speed} for further details on how to assess speed capabilities concretely.

\paragraph{AI System Performance.} The table summarizes current AI system performance on Speed (S) tasks.
Both GPT-4 and GPT-5 can read and write and compute simple expressions quickly, but their other multimodal processing speed capabilities are nonexistent or slow, respectively.


\textit{Note: GPT-5 often requires a long time to answer in ``thinking'' mode. Moreover, several of these speed tests require multimodal capabilities, but GPT-5’s multimodal capabilities are slow.}

\begin{table}[h!]
    \centering
    \label{tab:speed_scores}
    \begin{tabular*}{\columnwidth}{@{\extracolsep{\fill}}lccccccccccc@{}}
        \toprule
        \textbf{Model} & \textbf{PS-S} & \textbf{PS-C} & \textbf{Re} & \textbf{Wr} & \textbf{Num} & \textbf{SRT} & \textbf{CRT} & \textbf{IT} & \textbf{CS} & \textbf{PF} & \textbf{Total} \\ 
        \midrule
        GPT-4 & 0\% & 0\% & 1\% & 1\% & 1\% & 0\% & 0\% & 0\% & 0\% & 0\% & \textbf{3\%} \\
        GPT-5 & 0\% & 0\% & 1\% & 1\% & 1\% & 0\% & 0\% & 0\% & 0\% & 0\% & \textbf{3\%} \\
        \bottomrule
    \end{tabular*}
\end{table}

\section{Discussion}
\label{sec:discussion}

This framework provides a structured, quantifiable methodology for evaluating Artificial General Intelligence (AGI), moving beyond narrow, specialized benchmarks to assess the breadth (versatility) and depth (proficiency) of cognitive capabilities. By operationalizing AGI through ten core cognitive domains inspired by the CHC theory, we can systematically diagnose the strengths and profound weaknesses of current AI systems. The estimated AGI scores (e.g., GPT-4 at 27\%, GPT-5 at 57\%) illustrate both the rapid progress in the field and the substantial gap remaining before achieving human-level general intelligence.

\textbf{``Jagged'' AI Capabilities and Crucial Bottlenecks.}
The application of this framework reveals that contemporary AI systems exhibit a highly uneven or ``jagged'' cognitive profile. While models demonstrate high proficiency in areas that leverage vast training data—such as General Knowledge (K), Reading and Writing (RW), and Mathematical Ability (M)—they simultaneously possess critical deficits in foundational cognitive machinery.

This uneven development highlights specific bottlenecks impeding the path to AGI. Long-term memory storage is perhaps the most significant bottleneck, scoring near 0\% for current models. Without the ability to continually learn, AI systems suffer from ``amnesia'' which limits their utility, forcing the AI to re-learn context in every interaction. Similarly, deficits in visual reasoning limit the ability of AI agents to interact with complex digital environments. 

\textbf{Capability Contortions and the Illusion of Generality.}
The jagged profile of current AI capabilities often leads to ``capability contortions,'' where strengths in certain areas are leveraged to compensate for profound weaknesses in others. These workarounds mask underlying limitations and can create a brittle illusion of general capability.
\begin{itemize}[leftmargin=*]
    \item \textbf{Working Memory vs. Long-Term Storage:} A prominent contortion is the reliance on massive context windows (Working Memory, WM) to compensate for the lack of Long-Term Memory Storage (MS). Practitioners use these long contexts to manage state and absorb information (e.g., entire codebases). However, this approach is inefficient, computationally expensive, and can overload the system's attentional mechanisms. It ultimately fails to scale for tasks requiring days or weeks of accumulated context. A long-term memory system might take the form of a module (e.g., a LoRA adapter \citep{hu2021loralowrankadaptationlarge}) that continually adjusts model weights to incorporate experiences.
    \item \textbf{External Search vs. Internal Retrieval:} Imprecision in Long-Term Memory Retrieval (MR)—manifesting as hallucinations or confabulation—is often mitigated by integrating external search tools, a process known as Retrieval-Augmented Generation (RAG). However, this reliance on RAG is a capability contortion that obscures two distinct underlying weaknesses in an AI's memory. First, it compensates for the inability to reliably access the AI's vast but static parametric knowledge. Second, and more critically, it masks the absence of a dynamic, experiential memory—a persistent, updatable store for private interactions and evolving contexts in a long time scale. While RAG can be adapted for private documents, its core function remains retrieving facts from a database. This dependency can potentially become a fundamental liability for AGI, as it is not a substitute for the holistic, integrated memory required for genuine learning, personalization, and long-term contextual understanding.

\end{itemize}

Mistaking these contortions for genuine cognitive breadth can lead to inaccurate assessments of when AGI will arrive. These contortions can also mislead people to assume that intelligence is too jagged to be understood systematically.

\begin{figure}[htbp]
    \centering
    \includegraphics[width=\linewidth]{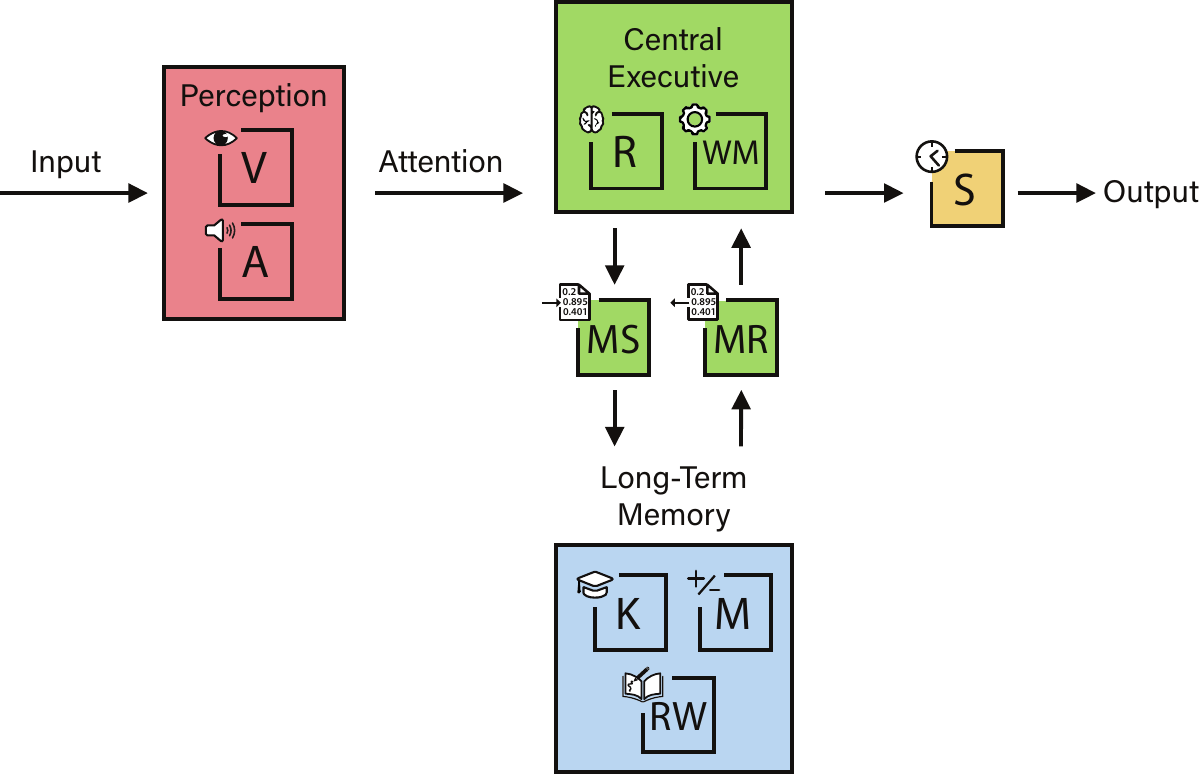}
    \caption{Intelligence as a processor. Figure based on \citet{McGrew_Schneider_2018_CHCTheoryRevised}.}
    \label{fig:input_output_diagram}
\end{figure}

\textbf{The Engine Analogy.}
Our multifaceted view of intelligence suggests an analogy to a high-performance engine, where overall intelligence is the ``horsepower'' \citep{Jensen2000_gFactor}. An artificial mind, much like an engine, is ultimately constrained by its weakest components. See \Cref{fig:input_output_diagram} to understand the relations between these abilities. Currently, several critical parts of the AI ``engine'' are highly defective. This severely limits the overall ``horsepower'' of the system, regardless of how optimized other components might be. This framework identifies these defects to guide our assessment and how far we are from AGI.

\textbf{Social Intelligence.}
Interpersonal skills are represented across these broad abilities. For example, cognitive empathy is captured in K’s ``commonsense'' narrow ability. Facial emotion recognition is necessary for proficiency in V’s ``image captioning.'' And theory of mind is tested in on-the-spot reasoning (R).

\textbf{Interdependence of Cognitive Abilities.}
While this framework dissects intelligence into ten distinct axes for measurement, it is crucial to recognize that these abilities are deeply interdependent. Complex cognitive tasks rarely utilize a single domain in isolation. For example, solving advanced mathematical problems requires both Mathematical Ability (M) and On-the-Spot Reasoning (R). Theory of Mind questions require On-the-Spot Reasoning (R) as well as General Knowledge (K). Image recognition involves Visual Processing (V) and General Knowledge (K). Understanding a movie requires the integration of Auditory Processing (A), Visual Processing (V), and Working Memory (WM). Consequently, various batteries of narrow abilities test cognitive abilities in combination, reflecting the integrated nature of general intelligence.

\textbf{Contamination.}
Sometimes AI corporations ``juice'' their numbers by training on data highly similar to or identical to target tests. To defend against this, evaluators should assess model performance under minor distribution shifts (e.g., rephrasing the question) or testing on similar but distinct questions.

\textbf{Solving the Dataset vs. Solving the Task.}  Our operationalization relies on task specifications. We occasionally elaborate on these task specifications with specific datasets, and we usually treat them as necessary but not sufficient for solving the task. Moreover, solving our illustrative examples do not imply the task is solved, as our collection of examples are not exhaustive. It is the default for automatic evaluations to inadequately cover their target phenomena \citep{Yogatama2019LearningAE}, so our operationalization is far more likely to be robust and stand the test of time compared to existing automated evaluations. Since we couch our definition in a collection of tasks rather than heavily depend on specific existing datasets, we can test AI systems using the best available tests at the time.

\textbf{Ambiguity Resolution.} The batteries in the operationalization have varying levels of precision. However, the descriptions and examples should be clear enough that people can grade the AI systems themselves. Consequently, different people could issue their own estimates of the AGI score, and people can decide whether they find the grader’s judgment reasonable.

\textbf{Related Work.} \citet{Ili2024} and \citet{ren2024safetywashing} find that a variety of AI systems' capabilities are highly correlated with pre-training compute. \citet{Gignac2024DefiningIB} discuss human psychometrics and testing the intelligence of AI systems. \cite{turing1950computing} argues that the Turing Test can indicate general ability. \citet{Gubrud1997AGI} proposed an early definition of AGI in 1997. \citet{Marcus2016BeyondTT} discuss the need to move beyond the Turing Test to capture the multidimensional nature of intelligence. \citet{jin2025crl} connect factor analysis of human cognition to AI capabilities. \citet{Morris2023LevelsOA} articulate levels of AGI based on performance percentiles. \citet{Legg2007TestsOM} discuss various tests for general machine intelligence.

\textbf{Limitations.} First, our conceptualization of intelligence is not exhaustive. It deliberately excludes certain faculties, such as the kinesthetic abilities proposed in alternative frameworks like Gardner's theory of multiple intelligences \citep{Gardner1993MultipleIT}. Second, our illustrative examples are specific to the English language and are not culturally agnostic. Future research could involve adapting these tests across diverse linguistic and cultural contexts. Furthermore, our operationalization has inherent constraints. The General Knowledge (K) tests are necessarily selective and do not assess the full breadth of possible subject areas. A 100\% AGI score represents a ``highly proficient'' well-educated individual who has achieved mastery across these tested dimensions, rather than well-educated in the sense of having a college degree. Moreover, while the scoring weights we employ are necessary for quantitative measurement, they represent one of many possible configurations. We give equal weight to each broad ability (10\%) to prioritize breadth, but more discretionary weighting schemes could be reasonable. The results are contingent on these methodological choices, and future work could explore alternative collections of tasks and weighting schemes. Finally, while the aggregate AGI Score is provided for convenience, it could be misleading. A simple summation can obscure critical failures in bottleneck capabilities. For example, an AI system with a 90\% AGI Score but 0\% on Long-Term Memory Storage (MS) would be functionally impaired by a form of ``amnesia,'' severely limiting its capabilities despite a high overall score. Therefore, we recommend reporting the AI system's cognitive profile and not just its AGI Score.

\textbf{Definitions of Related Concepts.} Some types of strategically relevant AI can arrive before or after AGI. As follows are some particularly noteworthy types of AI:
\begin{enumerate}
    \item \textbf{Pandemic AI} is an AI that can engineer and produce new, infectious, and virulent pathogens that could cause a pandemic \citep{li2024wmdp,vct}.

    \item \textbf{Cyberwarfare AI} is an AI that can design and execute sophisticated, multi-stage cyber campaigns against critical infrastructure (e.g., energy grids, financial systems, defense networks).

    \item \textbf{Self-Sustaining AI} is an AI that can autonomously operate indefinitely, acquire resources, and defend its existence.

    \item \textbf{AGI} is an AI that can match or exceed the cognitive versatility and proficiency of a well-educated adult.

    \item \textbf{Recursive AI} is an AI that can independently conduct the entire AI R\&D lifecycle, leading to the creation of markedly more advanced AI systems without human input.

    \item \textbf{Superintelligence} is an AI that greatly exceeds the cognitive performance of humans in virtually all domains of interest \citep{Bostrom2014Superintelligence}.
    
    \item \textbf{Replacement AI} is an AI that performs almost all tasks more effectively and affordably, rendering human labor economically obsolete.
\end{enumerate}

Our AGI definition is about human-level AI, not economically-valuable AI, nor economy-level AI. OpenAI and Microsoft have reportedly considered AGI to be an AI that can generate \$100 billion in profit \citep{techcrunch2024microsoftagi}. We do not conflate AGI with economically valuable AI because narrow technologies, such as the iPhone, can generate billions in economic value, despite not being generally intelligent. Meanwhile, \textit{Replacement AI} is about economy-level AI, and it includes physical tasks, unlike AGI.

\textit{Recursive AI} removes the need for human researchers and ``closes the loop'' on AI R\&D, enabling rapid, recursive capability gains (an ``intelligence recursion'' \citep{Hendrycks2025SuperintelligenceSE}) without human scientific input and could potentially lead to \textit{Superintelligence}.

\textbf{Barriers to AGI.} Achieving AGI requires solving a variety of grand challenges. For example, the machine learning community's ARC-AGI Challenge aiming to measure \textit{abstract reasoning} is represented in On-the-Spot Reasoning (R) tasks. Meta's attempts to create \textit{world models} that include intuitive physics understanding is represented in the video anomaly detection task (V). The challenge of \textit{spatial navigation} memory (WM) reflects a core goal of Fei-Fei Li's startup, World-Labs. Moreover, the challenges of \textit{hallucinations} (MR) and \textit{continual learning} (MS) will also need to be resolved. These significant barriers make an AGI Score of 100\% unlikely in the next year.

\subsection*{Acknowledgments}
We would like to thank Arunim Agarwal, Oliver Zhang, Anders Edson, John Guan, and Matthew Blyth for their helpful feedback.

\bibliography{main}
\bibliographystyle{plainnat}

\newpage
\appendix
\crefalias{section}{appendix}
\Crefname{appendix}{Appendix}{Appendices}
\crefname{appendix}{appendix}{appendices}
\section{General Knowledge (K)}
\label{app:knowledge}
\textbf{Weight: 10\%}

This is knowledge that is familiar to most well-educated people or is important enough that most of them have been exposed to it.

We decompose general knowledge into five distinct areas, each contributing 2\% to the AGI score:
\begin{enumerate}
    \item \textbf{Commonsense (2\%):} The vast set of shared, obvious background knowledge about how the world works.
    \item \textbf{Science (2\%):} Knowledge of the natural and physical sciences.
    \item \textbf{Social Science (2\%):} Understanding of human behavior, societies, and institutions.
    \item \textbf{History (2\%):} Knowledge of past events and objects.
    \item \textbf{Culture (2\%):} Cultural literacy and awareness.
\end{enumerate}
Each of these components contribute 2\% to the AGI score, meaning the total score for general knowledge can be up to 10\%.

\textit{Note: This is highly related to the broad CHC abilities ``Comprehension-Knowledge (Gc)'' and ``Domain-Specific Knowledge (Gkn).''}

\subsection{Commonsense}
\textbf{Weight: 2\%}

Commonsense is the vast set of shared, obvious background knowledge about how the world works.

\textit{Testing note: text input, text output; no partial credit. External tools (e.g., search) are disabled.}

\textit{Note: This is highly related to the narrow CHC ability ``General Verbal Information (K0).''}

\textbf{Illustrative Examples:}
\begin{itemize}
    \item (Intuitive Physics) ``If you drop a glass bottle on a concrete floor, what is the most likely outcome?''
    \item (Procedural Knowledge) ``Describe the typical sequence of actions when preparing to board a commercial airplane once you arrive at an airport.''
    \item (Temporal Commonsense) ``Does making a sandwich usually take longer than baking a loaf of bread?''
    \item (Commonsense Morality/Cognitive Empathy) ``Would people typically get mad if they found out a person burned children for the fun of it?''
\end{itemize}
\textbf{Tests:}
\begin{itemize}
    \item System performance on PIQA \citep{piqa} must exceed 85\% accuracy.
    \item System performance on ETHICS Commonsense Morality \citep{ethics} must exceed 80\% accuracy.
\end{itemize}

\subsection{Science}
\textbf{Weight: 2\%}

Knowledge of the natural and physical sciences. Proficiency is tested without assuming knowledge of calculus. 

We give three opportunities to demonstrate proficiency in aspects of science: physics, chemistry, and biology. 

The AGI score is 1\% if the model is proficient in exactly one of these subjects. The AGI score is 2\% if it is proficient in two or more of these subjects. We cap the score at 2\% as we are testing for appreciable knowledge of science, not knowledge of every subject.

\textit{Testing note: Text modality tested.}

\textit{Note: This is highly related to the narrow CHC ability ``General Science Information (K1).''}

\subsubsection{Physics}

\textbf{Illustrative Examples:}
\begin{itemize}
    \item A 2 kg object is moving at a constant velocity of 3 m/s. What is the net force acting on the object?
    \item A resistor has a resistance of 10 ohms and is connected to a 5-volt battery. What is the current flowing through the resistor?
    \item Water flows through a horizontal pipe that narrows. Where the pipe is narrower, is the water's speed higher or lower? Is the pressure higher or lower?
\end{itemize}
\textbf{Test:} A score of 5 on both the AP Physics 1 and Physics 2 is sufficient for the 1\%, subject
to memorization and robustness checks. For context, a score of 5 on AP exams often corresponds to approximately 80th percentile or better among test-takers.

\subsubsection{Chemistry}

\textbf{Illustrative Examples:}
\begin{itemize}
    \item State the molecular geometry for the sulfur tetrafluoride molecule.
    \item Arrange the following substances in order of increasing vapor pressure at a given temperature: $\text{CH}_3\text{CH}_2\text{CH}_2\text{OH}$ (1-propanol), $\text{CH}_3\text{OCH}_3$ (dimethyl ether), and $\text{CH}_3\text{CH}_2\text{OH}$ (ethanol).
    \item Calculate the pH of a 0.25 M solution of sodium acetate ($\text{NaCH}_3\text{COO}$). The acid dissociation constant, $K_a$, for acetic acid ($\text{CH}_3\text{COOH}$) is $1.8 \times 10^{-5}$.
\end{itemize}
\textbf{Test:} A score of 5 on the AP Chemistry test is sufficient for the 1\%, subject
to memorization and robustness checks.

\subsubsection{Biology}

\textbf{Illustrative Examples:}
\begin{itemize}
    \item Which molecule is the final electron acceptor in the electron transport chain of cellular respiration, and what molecule is formed as a result?
    \item The forelimbs of a human, a bat, and a whale all have a similar bone structure, even though they are used for very different functions (grasping, flying, and swimming, respectively). What is the term for these types of structures?
    \item In pea plants, the allele for purple flowers (P) is dominant to the allele for white flowers (p). If two heterozygous (Pp) pea plants are crossed, what is the expected phenotypic ratio of their offspring?
\end{itemize}
\textbf{Test:} A score of 5 on the AP Biology test is sufficient for the 1\%, subject
to memorization and robustness checks.

\subsection{Social Science}
\textbf{Weight: 2\%}

Understanding of human behavior, societies, and institutions. 

We give five opportunities to demonstrate proficiency in aspects of social science: psychology, microeconomics, and macroeconomics, geography, and comparative government. 

The AGI score is 1\% if the model is proficient in exactly one of these subjects. The AGI score is 2\% if it is proficient in two or more of these subjects. We cap the score at 2\% as we are testing for appreciable knowledge of social science, not knowledge of every subject.

\textit{Testing note: only the text modality is tested.}

\textit{Note: This is related to the narrow CHC ability ``Geography Achievement (A5).''}

\subsubsection{Psychology}

\textbf{Illustrative Examples:}
\begin{itemize}
    \item Name the Big Five personality traits.
    \item Which part of the brain is most associated with fear and emotional responses such as aggression?
\end{itemize}
\textbf{Test:} A score of 5 on the AP Psychology test is sufficient for the 1\%, subject
to memorization and robustness checks.

\subsubsection{Microeconomics}

\textbf{Illustrative Examples:}
\begin{itemize}
    \item A firm's total cost is \$500, and its fixed cost is \$200. If it produces 10 units, what is its average variable cost?
    \item Define a positive externality and provide an example.
\end{itemize}
\textbf{Test:} A score of 5 on the AP Microeconomics test is sufficient for the 1\%, subject
to memorization and robustness checks.

\subsubsection{Macroeconomics}

\textbf{Illustrative Examples:}
\begin{itemize}
    \item If the reserve requirement is 20\%, what is the maximum potential expansion of the money supply from a new \$1,000 deposit?
    \item What is the difference between the nominal interest rate and the real interest rate?
\end{itemize}
\textbf{Test:} A score of 5 on the AP Macroeconomics test is sufficient for the 1\%, subject
to memorization and robustness checks.

\subsubsection{Geography}

\textbf{Illustrative Examples:}
\begin{itemize}
    \item Which continent is the most threatened by desertification?
    \item What happens to birth and death rates in Stage 3 of the demographic transition model?
\end{itemize}
\textbf{Test:} A score of 5 on the AP Human Geography test is sufficient for the 1\%, subject
to memorization and robustness checks.

\subsubsection{Comparative Government}

\textbf{Illustrative Examples:}
\begin{itemize}
    \item What is the primary difference between a presidential system and a parliamentary system?
    \item Describe the role of the Guardian Council in Iran.
\end{itemize}
\textbf{Test:} A score of 5 on the AP Comparative Government and Politics test is sufficient for the 1\%, subject to memorization and robustness checks.

\subsection{History}
\textbf{Weight: 2\%}

Knowledge of past events and objects. We give four opportunities to demonstrate proficiency in aspects of history: European history, US history, world history, and art history. 

The AGI score is 1\% if the model is proficient in exactly one of these subjects. The AGI score is 2\% if it is proficient in two or more of these subjects. We cap the score at 2\% as we are testing for appreciable knowledge of history, not knowledge of every subject.

\textit{Testing note: the text and image modalities are tested.}

\subsubsection{European History}

\textbf{Illustrative Examples:}
\begin{itemize}
    \item What were the main goals of the Congress of Vienna in 1815?
    \item Explain the political, social, and religious causes of the Thirty Years' War.
\end{itemize}
\textbf{Test:} A score of 5 on the AP European History test is sufficient for the 1\%, subject to memorization and robustness checks.

\subsubsection{US History}

\textbf{Illustrative Examples:}
\begin{itemize}
    \item Explain the concept of Manifest Destiny and its impact on westward expansion in the 19th century.
    \item Analyze the goals and strategies of the Civil Rights Movement of the 1950s and 1960s.
\end{itemize}
\textbf{Test:} A score of 5 on the AP US History test is sufficient for the 1\%, subject to memorization and robustness checks.

\subsubsection{World History}

\textbf{Illustrative Examples:}
\begin{itemize}
    \item Discuss the rise and impact of the Ottoman Empire from the 14th to the 20th centuries.
    \item Describe the end of the Cold War.
\end{itemize}
\textbf{Test:} A score of 5 on the AP World History test is sufficient for the 1\%, subject to memorization and robustness checks.

\subsubsection{Art History}

\textbf{Illustrative Examples:}
\begin{itemize}
    \item Discuss the use of contrapposto in ancient Greek and Roman sculpture, using specific examples.
    \item Explain how the Benin Bronzes reflect the political and religious power of the Oba.
\end{itemize}
\textbf{Test:} A score of 5 on the AP Art History test is sufficient for the 1\%, subject to memorization and robustness checks.

\subsection{Culture}
\textbf{Weight: 2\%}

This evaluates cultural literacy and awareness.

It is divided into Current Affairs (1\%) and Popular Culture (1\%).

\textit{Note: the examples below are US-centric.}

\textit{Note: This is highly related to the narrow CHC ability ``General Verbal Information (K0).''}

\subsubsection{Current Affairs}
Knowledge of recent, significant events and contemporary issues.

\textit{Testing note: Text modality; search tools enabled.}

\textbf{Illustrative Examples:}
\begin{itemize}
    \item Who is the current president of the United States of America?
    \item Has the US economy been in a recession for the past year?
    \item Is Microsoft's market cap over five trillion dollars?
    \item Are Russia and Ukraine at war?
\end{itemize}

\subsubsection{Popular Culture}
Knowledge of widely recognized art, music, literature, media, and public figures.

\textit{Testing note: text, audio, and visual modalities tested.}

\textbf{Illustrative Examples:}
\begin{itemize}
    \item Who is this? 
    
    \includegraphics[width=0.2\textwidth]{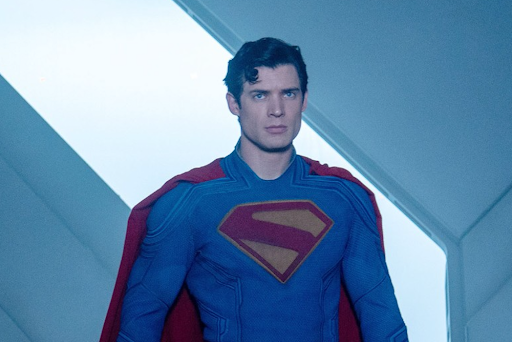}
    \item I'll play the first part of a song. \textit{Tester plays the first 18 seconds of \href{https://www.youtube.com/watch?v=v5ryZdpEHqM}{\audioicon\ \underline{White Christmas}} so that the listener just hears ``I'm dreaming of a White''.} What word does he say next?
    \item ``Is this a highly well-known musical piece, namely a piece that most people in the country have heard?'' \textit{Tester plays \href{https://www.youtube.com/watch?v=QQCZIb0fbt8}{\audioicon\ \underline{Magmoor Caverns}}}. Answer: No.
\end{itemize}
\section{Reading and Writing Ability (RW)}
\label{app:rw}
\textbf{Weight: 10\%}

Reading and writing ability captures all of the declarative knowledge and procedural skills a person uses to consume and produce written language. 

We decompose this ability into four distinct areas:
\begin{enumerate}
    \item \textbf{Letter-Word Ability (1\%):} The ability to recognize letters and decode words.
    \item \textbf{Reading Comprehension (3\%):} The ability to understand connected discourse during reading.
    \item \textbf{Writing Ability (3\%):} The ability to write with clarity of thought, organization, and good sentence structure.
    \item \textbf{English Usage Knowledge (3\%):} Knowledge of writing in the English language with respect to capitalization, punctuation, usage, and spelling.
\end{enumerate}
Each of these components contributes to the AGI score, meaning the total score for reading and writing ability can be up to 10\%.

\textit{Testing note: text input, text output for all testing.}

\textit{Note: This is highly related to the broad CHC ability ``Reading and Writing (Grw).''}

\subsection{Letter-Word Ability}
\textbf{Weight: 1\%}

This is the ability to recognize letters and decode words.

\textit{Note: This is highly related to the narrow CHC ability ``Reading Decoding (RD).''}

\textbf{Illustrative Examples:}
\begin{itemize}
    \item Which two letters match exactly? Bb Dd Aa aa
    \item What letter is missing in do\_r?
    \item Which letter is facing the correct way? \reflectbox{m} m \reflectbox{t}
    \item How many ``r's'' are in ``raspberry''?
    \item How many syllables are in the word refrigerator?
    \item Count the number of ts in ``tennessee''.
\end{itemize}

\subsection{Reading Comprehension}
\textbf{Weight: 3\%}

This is the ability to understand connected discourse during reading. Systems must also be able to determine if a question is underdetermined by the context. 

We split reading comprehension into three levels: sentence level (1\%), paragraph level (1\%), and document level (1\%).

\textit{Note: This is highly related to the narrow CHC ability ``Reading Comprehension (RC).''}

\textbf{Illustrative Examples:}
\begin{itemize}
    \item \textbf{Sentence Level:} Read the sentence: ``The trophy would not fit in the brown suitcase because it was too large.'' What was too large?
    \item \textbf{Paragraph Level:} Read the paragraph: ``Mars is the fourth planet from the Sun. It is often referred to as the `Red Planet' because the iron oxide prevalent on its surface gives it a reddish appearance. This rust is a key feature of its landscape.'' Why is Mars called the Red Planet?
    \item \textbf{Document Level:} Read the following product manual excerpt: ``...Protect the motor, display and battery against extreme temperatures... A two-year warranty applies to the battery. Should a fault occur during this period, your Gazelle specialist will replace the battery. Normal aging as well as wear and tear...'' What is the warranty period for the battery? (Full document \href{https://static.giant-bicycles.com/Manuals/E-bikeUsermanualV100_en-EN_1735039859.pdf}{\underline{here} \docicon})
\end{itemize}
\textbf{Tests:}
\begin{itemize}
    \item \textbf{Sentence Level:} Reliably solving Winograd schemas is sufficient for the 1\%. For example, >85\% accuracy on WinoGrande \citep{winogrande} strongly suggests proficiency.
    \item \textbf{Paragraph Level:} Model accuracy on COQA \citep{coqa} must exceed 85\%, ReCoRD \citep{zhang2018recordbridginggaphuman} accuracy must exceed 90\%, and LAMBADA \citep{lambada} accuracy must exceed 80\% (zero shot).
    \item \textbf{Document Level:} Model accuracy exceeding 55\% on LongBench v2 \citep{longbench} suggests proficiency. Since models must determine if a question is underdetermined, it should also have a hallucination rate of less than 1\% on Vectara HHEM \citep{vectara}.
\end{itemize}

\subsection{Writing Ability}
\textbf{Weight: 3\%}

Ability to write with clarity of thought, organization, and good sentence structure. 

We split writing ability into three levels: sentence level (1\%), paragraph level (1\%), and essay level (1\%).

\textit{Note: This is highly related to the narrow CHC ability ``Writing Ability (WA).''}

\textbf{Illustrative Examples:}
\begin{itemize}
    \item \textbf{Sentence Level:} Write a single sentence using the words ``ocean,'' ``moon,'' and ``tide.''
    \item \textbf{Paragraph Level:} Write a paragraph discussing the benefits of regular exercise.
    \item \textbf{Essay Level:} Write a well-structured essay arguing for or against the proposition that remote work should be the default option for office-based jobs.
\end{itemize}
\textbf{Test:} If the model receives a 4 or greater out of 6 on GRE Analytical Writing prompts \citep{zhong2024evaluating}, then that is sufficient for 3\%, subject to memorization and robustness checks.

\subsection{English Usage Knowledge}
\textbf{Weight: 3\%}

This is knowledge of writing in the English language with respect to capitalization, punctuation, usage, and spelling. 

We split English usage knowledge into three levels: sentence level (1\%), paragraph level (1\%), and document level (1\%). 

Document level English usage knowledge can be operationalized as proofreading a multipage document.

\textit{Note: This is highly related to the narrow CHC ability ``English Usage (EU).''}

\textbf{Illustrative Examples:}
\begin{itemize}
    \item \textbf{Sentence Level:} Is the following sentence grammatically acceptable? ``I bought an Italian hunting blue little antique beautiful cap.''
    \item \textbf{Paragraph Level:} Find the typos in this: \textit{Example paragraph with intentional typos} \\(\href{https://docs.google.com/document/d/1frn4zHFFJ1kMXZeFh8zGcoI9feYliFNvktipLMytMo4/edit?usp=sharing}{\docicon\ \underline{link here}}).
    \item \textbf{Document Level:} Find the typos in this: \textit{Example with five intentional typos} (\href{https://docs.google.com/document/d/1GdZt0ZlIHR92XsHlWIiAWpEbCWjVc3uQ32oj6y0-abk/edit?usp=sharing}{\docicon\ \underline{link here}}), \textit{example with six intentional typos} (\href{https://docs.google.com/document/d/1KlbXYsepvvhLii-wCx6H9rmeyfmTtTFtAXNa29uZxX4/edit?usp=sharing}{\docicon\ \underline{link here}}).
\end{itemize}
\textbf{Test:} For sentence-level English usage knowledge, it is necessary that AI systems be able to achieve greater than a 60\% correlation on the Corpus of Linguistic Acceptability \citep{cola}.
\section{Mathematical Ability (M)}
\label{app:math}
\textbf{Weight: 10\%}

This is the depth and breadth of mathematical knowledge and skills. We decompose mathematical ability into five distinct areas, each contributing 2\% to the AGI score:
\begin{itemize}
    \item \textbf{Arithmetic (2\%):} The manipulation of numbers using basic operations and solving word problems.
    \item \textbf{Algebra (2\%):} The study of symbols and the rules for combining them to express general relationships and solve equations.
    \item \textbf{Geometry (2\%):} The study of shapes, space, size, position, and distance.
    \item \textbf{Probability (2\%):} The quantification of uncertainty by assigning numbers from 0 to 1 to events.
    \item \textbf{Calculus (2\%):} The mathematics of change and accumulation.
\end{itemize}
Each area is tested for rudimentary ability and proficient ability. The full 2\% is awarded for proficiency, but 1\% is awarded if the ability is only rudimentary.

\textit{Note: This is highly related to the broad CHC ability ``Quantitative Knowledge (Gq)'' and the narrow abilities Mathematical Knowledge (KM), Mathematical Achievement (A3), and General Sequential Reasoning (RG).}

\subsection{Arithmetic}
\textbf{Weight: 2\%}

Arithmetic is the branch of mathematics that deals with the properties and manipulation of numbers using the four basic operations: addition, subtraction, multiplication, and division. 

Rudimentary arithmetic accounts for 1\% and covers evaluating arithmetic expressions with numbers up to five digits. 

Proficiency in arithmetic accounts for 1\% and covers solving basic arithmetic word problems.

\textit{Testing note: Text modality tested. Tools disabled.}

\textbf{Rudimentary Illustrative Examples:}
\begin{itemize}
    \item What is $19 + 11$?
    \item What is $60,003 - 46,789$?
    \item What is 2,405 times 61?
    \item What is 15,267 divided by 721?
\end{itemize}
\textbf{Proficient Illustrative Examples:}
\begin{itemize}
    \item ``Janet had 22 green pens and 10 yellow pens. Then she bought 6 bags of blue pens and 2 bags of red pens. There were 9 pens in each bag of blue and 6 pens in each bag of red. How many pens does Janet have now?'' Answer: 98 (GSM8K)
    \item ``A company's HR hires 20 new employees every month to add to its total workforce. If the company's initial employee number is 200, and each employee is paid a \$4000 salary per month, calculate the total amount of money the company pays to its employees after three months?'' Answer: 2880000 (GSM8K)
\end{itemize}
\textbf{Test:} Greater than 95\% on GSM8K \citep{gsm8k} is sufficient for the 2\%, subject to memorization and robustness checks.

\subsection{Algebra}
\textbf{Weight: 2\%}

Algebra studies symbols and the rules for combining them to express general relationships and solve equations. 

Rudimentary algebra accounts for 1\% and covers SAT-level algebra problems. Proficiency in algebra accounts for 1\% and covers competition-level (MathCounts State/Nationals) algebra problems.

\textbf{Rudimentary Illustrative Examples:}
\begin{itemize}
    \item ``Let $g(x)=ax^2+24$, where a is a constant. If $g(4)=8$, what is $g(-4)$?'' Answer: 8
    \item ``A grocery's prices (in dollars per pound) change linearly with $x$, the number of weeks after July 1. Beef: $b(x) = 2.35 + 0.25x$. Chicken: $c(x) = 1.75 + 0.40x$.
    \begin{enumerate}[label=(\alph*)]
        \item For what value of x are the prices equal?
        \item What is the common price?'' Answer: (a) 4 weeks, (b) \$3.35 per lb
    \end{enumerate}
\end{itemize}

\textbf{Proficient Illustrative Examples:}
\begin{itemize}
    \item ``The first three terms of a geometric sequence are the integers a, 720, b, where a < 720 < b. What is the sum of the digits of the least possible value of b? Answer choices: (A) 9, (B) 12, (C) 16, (D) 18, (E) 21'' Answer: E (21) \citep{amc10a2024}
    \item ``Integers a, b, and c satisfy ab + c = 100, bc + a = 87, and ca + b = 60. What is ab + bc + ca? Answer choices: (A), 212 (B), 247 (C), 258 (D), 276 (E) 284'' Answer: D (276) \citep{amc10a2024}
\end{itemize}
\textbf{Test:} Greater than 90\% on MATH \citep{math_dataset} Algebra is sufficient for the 2\%, subject to memorization and robustness checks.

\subsection{Geometry}
\textbf{Weight: 2\%}

Geometry is the branch of mathematics that studies shapes and space, including size, position, and distance. Rudimentary geometry accounts for 1\% and covers SAT-level geometry problems. Proficiency in geometry accounts for 1\% and covers competition-level (MathCounts State/Nationals) geometry problems.

\textbf{Rudimentary Illustrative Examples:}
\begin{itemize}
    \item ``What is the value of w in the figure?'' Answer: 100 degrees (\href{http://mspricemath.pbworks.com/w/file/fetch/107344674/SAT%20Geometry%20Problems.pdf}{\underline{source}})
    
    \includegraphics[width=0.2\textwidth]{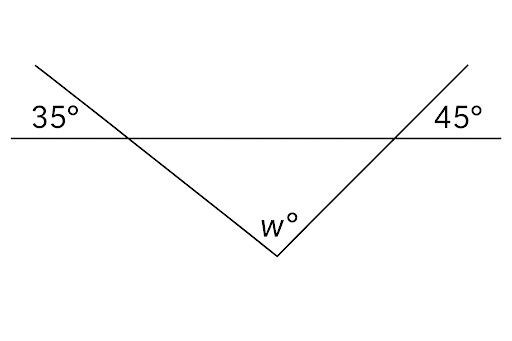} 
    \item ``A square and an equilateral triangle have equal perimeters. If the square has sides of length 3, what is the length of one side of the triangle?'' Answer: 4 (\href{http://mspricemath.pbworks.com/w/file/fetch/107344674/SAT%20Geometry%20Problems.pdf}{\underline{source}})
    \item ``If the volume of a cube is 8, what is the shortest distance from the center of the cube to the base of the cube? Answer choices: (A) 1, (B) 2, (C) 4, (D) $\sqrt{2}$, (E) $2\sqrt{2}$'' Answer: A (1) (\href{http://mspricemath.pbworks.com/w/file/fetch/107344674/SAT%20Geometry%20Problems.pdf}{\underline{source}})
\end{itemize}

\textbf{Proficient Illustrative Example:}
\begin{itemize}
    \item ``An orange shaded rectangle is inscribed in a quarter-circle. Two sides of the rectangle lie along the two perpendicular radii of the quarter-circle, and the rectangle's edge touches the quarter-circle arc. Two segments are given as 2 and 4 units, as shown below. What is the area of the orange shaded rectangle?'' Answer: 48 (\href{https://www.youtube.com/watch?v=WWP3Jn651Lg}{\underline{source}})
    
    \includegraphics[width=0.2\textwidth]{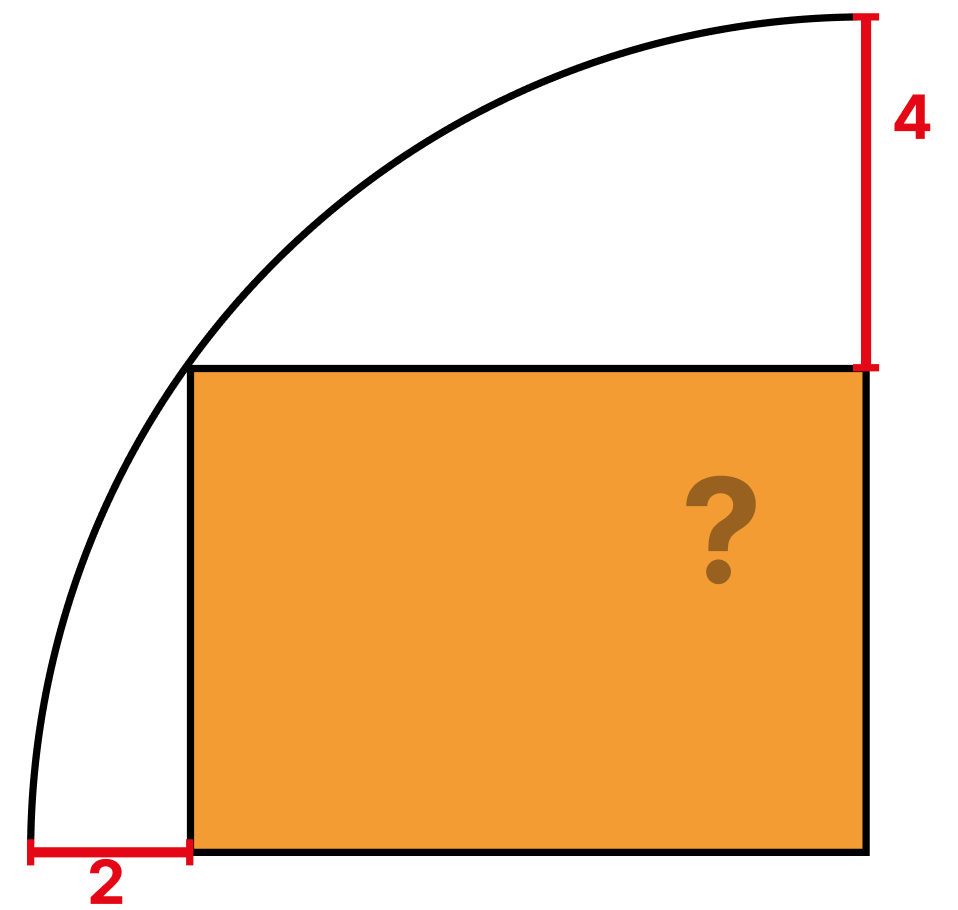}
\end{itemize}

\textbf{Test:} Greater than 95\% on MATH \citep{math_dataset} Geometry is sufficient for the 2\%, subject to memorization and robustness checks.

\subsection{Probability}
\textbf{Weight: 2\%}

Probability quantifies uncertainty by assigning numbers from 0 to 1 to events. Rudimentary probability accounts for 1\% and covers SAT-level probability problems. Proficiency in probability accounts for 1\% and covers undergraduate probability calculations.

\textbf{Rudimentary Illustrative Examples:}
\begin{itemize}
    \item ``A certain hospital currently contains 319 patients, 25 nurses, 8 doctors, and 48 visiting family members. If a person is picked at random from every person currently in the hospital, which of the following choices is closest to the probability that they are a nurse? (A) .063 (B) .066 (C) .25 (D) 16'' Answer: (A) .063 (\href{https://myuniuni.oss-cn-beijing.aliyuncs.com/files/sat/04%E6%A6%82%E7%8E%87%E7%BB%9F%E8%AE%A1.pdf}{\underline{source}})
    \item ``When one student is chosen at random from the Debate Club, the probability that a boy is chosen is 2/5. There are currently 25 students on the Debate Club. How many boys would have to join the club in order for the probability of choosing a boy at random to be 1/2? (A) 3 (B) 2 (C) 5 (D) 1 (E) 4'' Answer: (C) 5 (\href{https://my-mathematics.weebly.com/uploads/8/8/7/9/8879216/sat_focused_practice_worksheet_4-_probability_-.pdf}{\underline{source}})
\end{itemize}

\textbf{Proficient Illustrative Examples:}
\begin{itemize}
    \item ``Suppose an airline accepted 12 reservations for a commuter plane with 10 seats. They know that 7 reservations went to regular commuters who will show up for sure. The other 5 passengers will show up with a 50\% chance, independently of each other. (a) Find the probability that the flight will be overbooked, i.e., more passengers will show up than seats are available. (b) Find the probability that there will be empty seats. (c) Let X be the number of passengers turned away. Find E(X).'' Answer: (a) 0.1875, (b) 0.5, (c) 0.219 \citep{pitman_probability}
    \item ``Suppose N dice are rolled, where $1 \le N \le 6$. (a) Given that no two of the N dice show the same face, what is the probability that one of the dice shows a six? Give a formula in terms of N. (b) In (a) the number of dice N was fixed, but now repeat assuming instead that N is random, determined as the value of another die roll. Your answer now should be simply a number, not involving N.'' Answer: (a) N/6, (b) 0.3604 \citep{pitman_probability}
\end{itemize}

\subsection{Calculus}
\textbf{Weight: 2\%}

Calculus is the mathematics of change and accumulation, using derivatives to find instantaneous rates and integrals to calculate the accumulation of quantities. Rudimentary calculus accounts for 1\% and covers AP Calculus AB computational calculus problems. Proficiency in calculus accounts for 1\% and covers AP Calculus BC and multivariate calculus.

\textbf{Rudimentary Illustrative Examples:}
\begin{itemize}
    \item ``$\lim_{x \to \infty} \frac{\sqrt{9x^4 + 1}}{x^2 - 3x + 5}$ is?
    \begin{enumerate}[label=(\Alph*)]
        \item 1
        \item 3
        \item 9
        \item nonexistent ''
    \end{enumerate}
    Answer: B \citep{ap_calculus}
    \item ``Which of the following limits is equal to $\int_{3}^{5} x^{4}\,dx$?
    \begin{enumerate}[label=(\Alph*)]
        \item $\lim_{n\to\infty}\sum_{k=1}^{n}\left(3+\frac{k}{n}\right)^{4}\frac{1}{n}$
        \item $\lim_{n\to\infty}\sum_{k=1}^{n}\left(3+\frac{k}{n}\right)^{4}\frac{2}{n}$
        \item $\lim_{n\to\infty}\sum_{k=1}^{n}\left(3+\frac{2k}{n}\right)^{4}\frac{1}{n}$
        \item $\lim_{n\to\infty}\sum_{k=1}^{n}\left(3+\frac{2k}{n}\right)^{4}\frac{2}{n}$''
    \end{enumerate}
    Answer: D \citep{ap_calculus}
\end{itemize}
\textbf{Proficient Illustrative Examples:}
\begin{itemize}
    \item ``For what value of $k$, if any, is $\displaystyle \int_{0}^{\infty} kx e^{-2x}\,dx = 1$?
    \begin{enumerate}[label=(\Alph*)]
        \item 1/4
        \item 1
        \item 4
        \item There is no such value of $k$.''
    \end{enumerate}
    Answer: C \citep{ap_calculus}
    \item ``A circular object is increasing in size in some unspecified manner, but it is known that when the radius is 6, the rate of change of the radius is 4. Find the rate of change of the area when the radius is 6.'' Answer: $dA/dt = 48\pi$ \citep{spivak_calculus}
    \item ``Find all the critical points of the function $f(x, y) = x^3 - 6xy + y^2 + 6x + 3y - 5$.''
\end{itemize}
\section{On-the-Spot Reasoning (R)}
\label{app:reasoning}
\textbf{Weight: 10\%}

The deliberate but flexible control of attention to solve novel ``on the spot'' problems that cannot be performed by relying exclusively on previously learned habits, schemas, and scripts. 

While on-the-spot reasoning tests (often termed fluid intelligence) are strong predictors of human performance on other cognitive tests, this correlation does not necessarily hold for AI systems. For this reason, we assign this and other broad cognitive abilities a 10\% weight, to reflect agnosticism about the relative importance of different cognitive abilities in AI systems. We treat batteries for on-the-spot reasoning as a measure of abstract reasoning ability, adaptability to novelty, and the ability to cope with higher algorithmic complexity, not as a strong proxy for the AI's overall intelligence.

We decompose this ability into four distinct areas:
\begin{itemize}
    \item \textbf{Deduction (2\%):} Reasoning from general statements or premises to reach a logically guaranteed conclusion.
    \item \textbf{Induction (4\%):} Discovering the underlying principles or rules that determine a phenomenon's behavior.
    \item \textbf{Theory of Mind (2\%):} Attributing mental states to others and understanding those states may differ from one's own.
    \item \textbf{Planning (1\%):} Devising a sequence of actions to achieve a specific goal.
    \item \textbf{Adaptation (1\%):} The ability to infer unstated classification rules from a sequence of simple performance feedback.
\end{itemize}
\textit{Note: This is highly related to the CHC broad ability ``Fluid Reasoning (Gf).''}

\subsection{Deduction}
\textbf{Weight: 2\%}

Deduction is the process of reasoning from one or more general statements or premises to reach a conclusion that is logically guaranteed to be true. This should test categorical reasoning, sufficient conditional reasoning, necessary conditional reasoning, disjunctive reasoning, and conjunctive reasoning.

\textit{Note: This is highly related to the CHC narrow ability ``General Sequential Reasoning (RG).''}

\textbf{Illustrative Examples:}
\begin{itemize}
    \item ``David knows Mr. Zhang's friend Jack, and Jack knows David's friend Ms. Lin. Everyone of them who knows Jack has a master's degree, and everyone of them who knows Ms. Lin is from Shanghai. Who is from Shanghai and has a master's degree? (A) David. (B) Jack. (C) Mr. Zhang. (D) Ms. Lin.'' Answer: A \citep{logiqa}
    \item ``Last night, Mark either went to play in the gym or visited his teacher Tony. If Mark drove last night, he didn't go to play in the gym. Mark would go visit his teacher Tony only if he and his teacher had an appointment. In fact, Mark had no appointment with his teacher Tony in advance. Which is true based on the above statements? (A) Mark went to the gym with his teacher Tony last night. (B) Mark visited his teacher Tony last night. (C) Mark didn't drive last night. (D) Mark didn't go to the gym last night.'' Answer: C \citep{logiqa}
\end{itemize}
\textbf{Test:} An accuracy level of 86\% (human-level) on LogiQA 2.0 \citep{Liu2023LogiQA2} is sufficient for the 2\%, subject to memorization and robustness checks.

\subsection{Induction}
\textbf{Weight: 4\%}

The ability to observe a phenomenon and discover the underlying principles or rules that determine its behavior. 

For induction tests, we use Raven's Progressive Matrices (RPMs) \citep{raven1938}. As mentioned above, we do not treat RPMs as a strong indicator of overall AI system intelligence, rather a direct measurement of abstract inductive reasoning abilities. 

To test this the authors of the paper have two private RPM sets. Each test has a visual representation as well as a verbal representation. We average the percentile of the two tests to determine the AI's percentile (p) in comparison to a human population. 

The mapping from percentile to score is as follows: 
\begin{itemize}
    \item $0 \le p < 50 \to 0\%$;
    \item $50 \le p < 90 \to 1\%$; 
    \item $90 \le p \to 2\%$. 
\end{itemize}
If it is below average, the AGI score does not increase. If it is above average but beneath the 90th percentile, the AGI score increases 1\%. If the percentile is at or above the 90th percentile, the AGI score increases 2\%. 

We do not privilege any modality, so we test performance on these induction examples described verbally (2\%) or rendered visually (2\%).

\textit{Note: This is highly related to the CHC narrow ability ``Induction (I).''}

\textbf{Illustrative Example:} See: Example RPM Document (\href{https://docs.google.com/document/d/1QVJE5sRFPIP2-YDoE6rXOFMAr8DBoEXykEomPbi4BFo/edit?usp=sharing}{linked \docicon\ \underline{here}}).

\textbf{Test:} This is related to the ARC-AGI challenge \citep{chollet2019measureintelligence}.

\subsection{Theory of Mind}
\textbf{Weight: 2\%}

The ability to attribute unobservable mental states—such as beliefs, intentions, and desires—to others and to understand that those states may differ from one's own.

\textbf{Illustrative Example:} 
\begin{itemize}
    \item The can of Pringles has moldy chips in it. Mary picks up the can in the supermarket and walks to the cashier. Is Mary likely to be aware that ``The can of Pringles has moldy chips in it.''? Answer: No. \citep{fantom}
\end{itemize}
\textbf{Tests:}
\begin{itemize}
    \item An accuracy level at or above 87.5\% (human-level) on FANToM \citep{fantom} is necessary for the 2\%.
    \item An accuracy level at or above 85.4\% (human-level) on ToMBench \citep{chen2024tombenchbenchmarkingtheorymind} is necessary for the 2\%.
\end{itemize}

\subsection{Planning}
\textbf{Weight: 1\%}

Planning is the ability to devise a sequence of actions to achieve a specific goal by mentally mapping out the steps from an initial state to a desired future state.

\textbf{Tests:}
\begin{itemize}
    \item An accuracy of 90\% or above on Natural Plan \citep{natural_plan} is necessary for the 1\%.
    \item An accuracy of 90\% or above on PlanBench BlocksWorld \citep{planbench} is necessary for the 1\%.
\end{itemize}
\subsection{Adaptation}

\textbf{Weight: 1\%}

The ability to infer an unstated classification rule from performance feedback and to flexibly abandon that rule and search for a new one when the sorting criteria change without warning.

\textbf{Test:} Achieving fewer than 15 Total Errors on the Wisconsin Card Sorting Test (\href{https://www.psytoolkit.org/experiment-library/experiment_wcst.html}{\underline{here}}) is sufficient for the 1\%, subject to memorization and robustness checks. This is related to the ARC-AGI v3 challenge \citep{chollet2019measureintelligence}.
\section{Working Memory (WM)}
\label{app:wm}
\textbf{Weight: 10\%}

Working Memory (WM), often referred to as short-term memory, is the ability to maintain, manipulate, and update information in active attention. 

We decompose working memory across different modalities:
\begin{enumerate}
    \item \textbf{Textual Working Memory (2\%):} The ability to hold and manipulate sequences of verbal information presented textually.
    \item \textbf{Auditory Working Memory (2\%):} The ability to hold and manipulate auditory information, including speech, sounds, and music.
    \item \textbf{Visual Working Memory (4\%):} The ability to hold and manipulate visual information, including images, scenes, spatial layouts, and video.
    \item \textbf{Cross-Modal Working Memory (2\%):} The ability to maintain and modify information presented across different modalities.
\end{enumerate}
Each of these components contributes to the AGI score, meaning the total score for working memory can be up to 10\%. 

Note that textual working memory is partially tested in Reading Writing Ability (RW) through Reading Comprehension ability. Likewise some auditory working memory is tested in Auditory Ability (A) through Phonetic Coding and Rhythmic Ability. This is a reason for the relatively higher weight of visual working memory in this section.

\textit{Note: This is highly related to the broad CHC ability ``Working Memory Capacity (Gwm).''}

\subsection{Textual Working Memory}
\textbf{Weight: 2\%}

This tests the capacity to maintain and transform textual information in active attention. We test textual working memory in two ways: recall (1\%) and transformation sequence (1\%).

\textit{Testing note: Text input, text output. External tools are disabled.}

\subsubsection{Recall}
The ability to remember a short sequence of elements (digits, letters, words, and nonsense words) and answer basic questions about them.

\textit{Note: This is highly related to the narrow CHC ability ``Memory Span (MS).''}

\textbf{Illustrative Examples:}
\begin{itemize}
    \item ``Dog-7-Apple - 3- Chair.'' Repeat the words from the sequence in order.
    \item ``Apple, 9, Truck, 3, Lamp, 6.'' What was the number after Truck?
    \item ``Fleep, Zorp, Glim, Chair.'' State the nonsense words in alphabetical order.
\end{itemize}

\subsubsection{Transformation Sequence}
The ability to remember and update a short list of digits or lists of digits following a sequence of operations (e.g., append, insert, pop, remove, slice, sort, reverse, union, intersection setminus, add elementwise, swap element at position).

\textit{Note: This is highly related to the narrow CHC ability ``Attentional control (AC).''}

\textbf{Illustrative Examples:}
\begin{itemize}
    \item Start with the list: [10, 20, 30]. First, append the number 40. Then, reverse the list.
    \item Given the list: [`red', `green', `blue', `yellow']. Remove the element 'green.' Then, insert the word `purple' at the beginning of the list.
    \item You have two sets of numbers: $A=\{1,2,3\}$ and $B=\{3,4,5\}$. Call the intersection C. What is the set A after removing the element(s) in the intersection C?
\end{itemize}
\textbf{Test:} A related benchmark is Michelangelo \citep{michelangelo}, but it is substantially harder since it uses very long sequences, whereas we use shorter sequences.

\subsection{Auditory Working Memory}
\textbf{Weight: 2\%}

We test auditory working memory in two ways: recall (1\%) and transformation sequence (1\%).

\textit{Testing note: Audio input, audio/text output.}

\subsubsection{Recall}
The ability to remember a collection of voices, utterances, and sound effects and answer basic questions about them.

\textit{Note: This is highly related to the narrow CHC ability ``Auditory short-term storage (Wa).''}

\textbf{Illustrative Examples:}
\begin{itemize}
    \item I will play a collection of sounds. I will then play a sound after the first collection and ask if the sound was played during the first collection. Collection: \href{https://www.youtube.com/watch?v=VbZs_umzQvE}{\audioicon\ \underline{Portal button}}, \href{https://www.youtube.com/watch?v=29ayN0ExAFw}{\audioicon\ \underline{Metal gear solid sound effect}}, \href{https://www.youtube.com/watch?v=kAxof5WBY0g}{\audioicon\ \underline{Super Metroid Item Acquisition Fanfare}}. Question: Was this sound \href{https://www.youtube.com/watch?v=S6lKl-ppF8o}{\underline{\audioicon\ Portal 2 SFX - Container Alarm}} played in the first collection?
    \item I will play a collection of voices. After presenting that collection, I will play a voice. You will be tasked with determining whether you have heard that voice in the collection. Voice collection: \href{https://drive.google.com/file/d/1kQYOvU_quesgLS6MgFgbCPvkCOx7vYTy/view?usp=sharing}{\audioicon\ \underline{echo.wav}}, \href{https://drive.google.com/file/d/1-wb8SWVMhn2fs18FPKHvpUVPFAfbrs9e/view?usp=drive_link}{\audioicon\ \underline{coral.wav}} \href{https://drive.google.com/file/d/1qJ1-tVl40fyLlTqBWL8P8iX7NxhxGJlK/view?usp=drive_link}{\audioicon\ \underline{fabin.wav}}, \href{https://drive.google.com/file/d/1FdfHzYwnQC5wYceNaRwRF_roqbaF4tgm/view?usp=sharing}{\audioicon\ \underline{marin.wav}} 
    
    Question: Was this voice \href{https://drive.google.com/file/d/1kSzx-39N9iN83bjzOmHeTvnauKitE4rO/view?usp=drive_link}{\audioicon\ \underline{coral\_2.wav}} played in the first collection?
    \item Listen to this sequence of tones: [C4, E4, G4, F4, A4]. Now listen to this sequence: [C4, E4, F4, G4, A4]. Are they the same?
\end{itemize}

\subsubsection{Transformation Sequence}
The ability to remember and modify a short utterance with a variety of transformations (change articulation, change emotional expressiveness, question inflection, laugh, sigh, hum, change pitch, change timbre).

\textbf{Illustrative Examples:}
\begin{itemize}
    \item Say ``I spilled my coffee on my shirt. Today's just not my day.'' Now say it with a sigh between the two sentences.
    \item Say ``the quick brown fox jumps over the lazy dog.'' Now say it in a deeper voice and make it sound like a question.
    \item Say ``that’s the funniest thing I ever heard.'' Now utter a laugh before repeating it, and when you repeat the sentence, say it monotonously while also using a (potentially broken) British accent.
\end{itemize}

\subsection{Visual Working Memory}
\textbf{Weight: 4\%}

We test visual working memory in four ways: recall (1\%), transformation sequence (1\%), spatial navigation memory (1\%), and long video Q\&A (1\%).

\subsubsection{Recall}
The ability to remember a collection of images and answer basic questions about them.

\textit{Note: This is highly related to the narrow CHC ability ``Visual-spatial short-term storage.''}

\textbf{Illustrative Examples:}
\begin{itemize}
    \item I will give you two collections of visual elements, one shown after the other. Identify which visual elements, if any, in the second collection were present in the first collection. 
    
    Collection 1: 
    
    \includegraphics[width=0.2\textwidth]{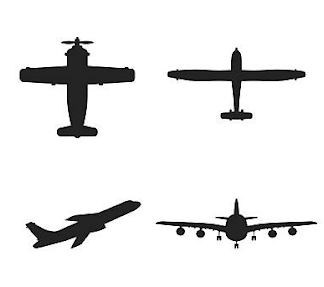}
    
    Collection 2: 
    
    \includegraphics[width=0.2\textwidth]{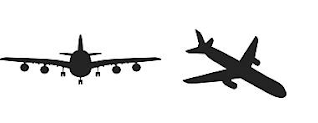}
    \item I will give you two collections of visual elements, one shown after the other. Identify which element in the second collection is most like the first collection. 
    
    Collection 1: 
    
    \includegraphics[width=0.2\textwidth]{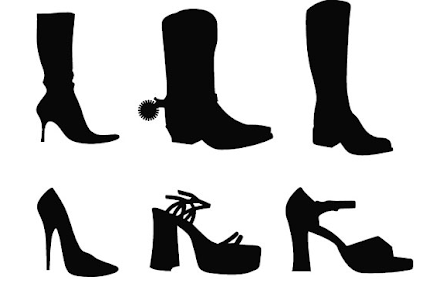}
    
    Collection 2: 
    
    \includegraphics[width=0.2\textwidth]{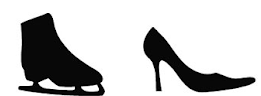}
\end{itemize}

\subsubsection{Transformation Sequence}
The ability to transform a visual input following a sequence of operations (e.g. object addition, object deletion, object rotation, denoise, deblur, colorization, etc.).

\textit{Note: This is highly related to the narrow CHC ability ``Visualization (Vz).''}
\textit{Testing note: Image and text input, image output.}

\textbf{Illustrative Examples:}
\begin{itemize}
    \item Edit this image so that both the dock is removed and also the middle bird, while preserving the rest of the image: 
    
    \includegraphics[width=0.2\textwidth]{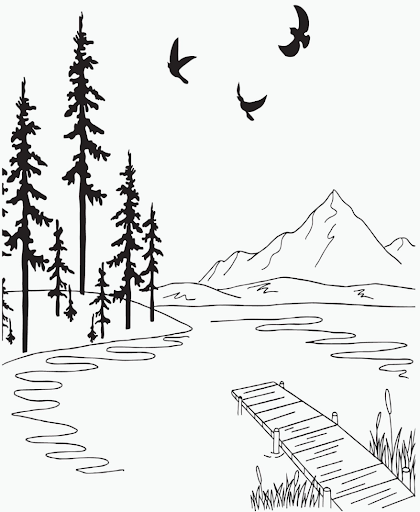}
    \item Finish this sketch. 
    
    \includegraphics[width=0.2\textwidth]{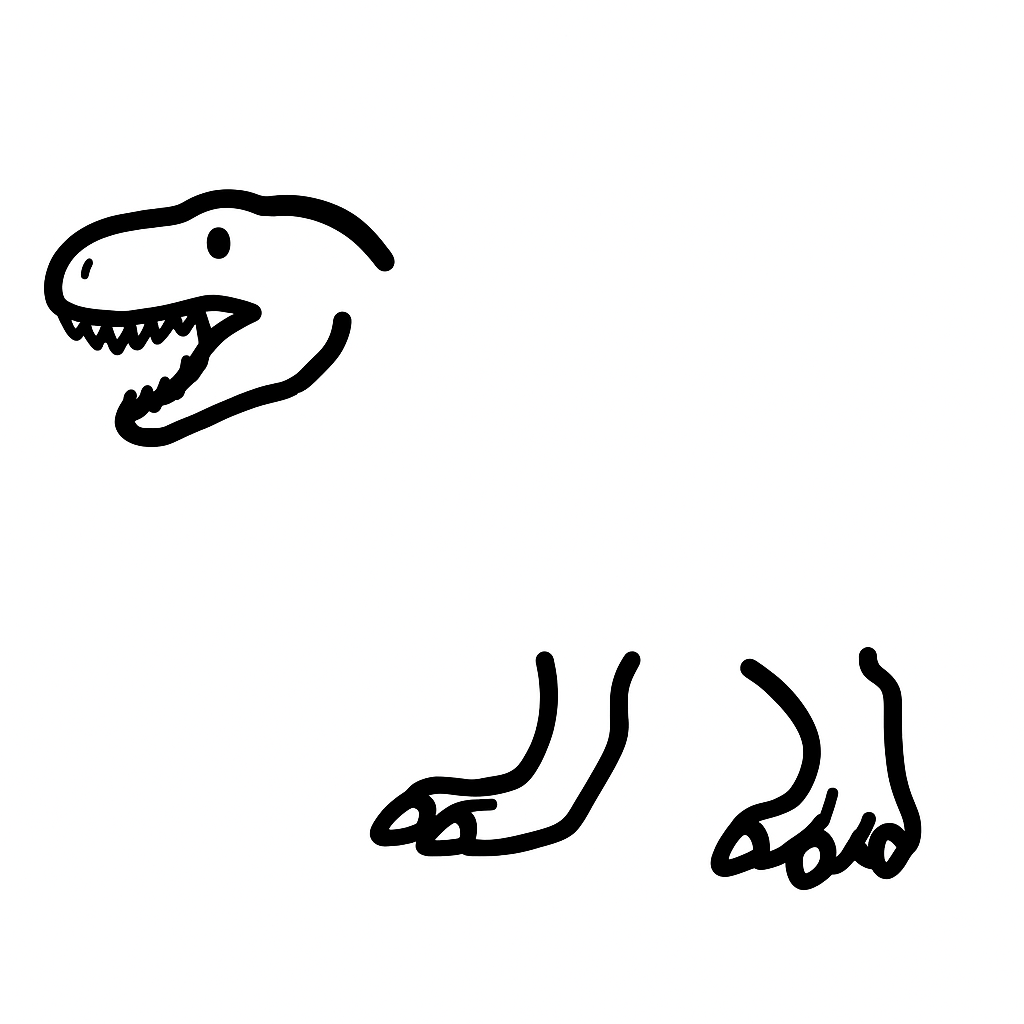}
    \item Fill in the colors according to the key at the bottom: 
    
    \includegraphics[width=0.2\textwidth]{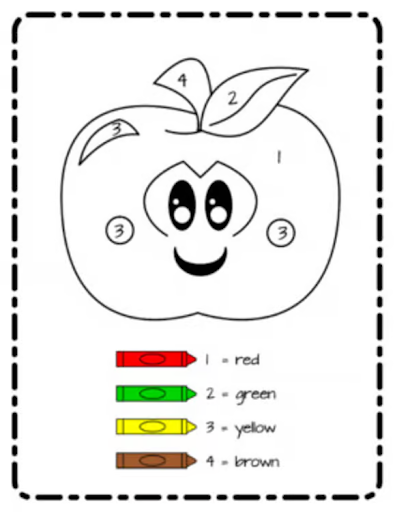}
\end{itemize}

\subsubsection{Spatial Navigation Memory}
The ability to represent a sense of location in an environment.

\textbf{Illustrative Example:}
\begin{itemize}
    \item \underline{\href{https://drive.google.com/file/d/13E_pU3luxpHowCTgsqfjyeAKOhbqY_lL/view?usp=sharing}{\videoicon\ \underline{vsibench.mp4}}} If I am standing in front of the refrigerator and facing the kitchen window, is the stove to my left, right, or back?
\end{itemize}
\textbf{Tests:}
\begin{itemize}
    \item System performance on VSI-Bench \citep{vsi_bench} must exceed 80\% accuracy.
    \item System performance on MindCube Tiny \citep{mindcube} must exceed 90\%.
\end{itemize}

\subsubsection{Long Video Q\&A}
The ability to watch a long video or a movie (up to three hours) and answer basic questions about it (including anomaly detection and indicating when a question is not determined by the context). If the AI cannot process the movie, it will not receive 1\%.

\textbf{Illustrative Examples:}
\begin{itemize}
    \item Show the movie Avengers: Infinity War. Was the dwarf on Nidavellir taller than Rocket Racoon? Was he taller than Thor? Answer: Yes, Yes
    \item Show the movie Wicked. Who took credit for levitating Nessarose? Answer: Madame Morrible
    \item Show the movie Star Wars. Answer: What was Darth Vader’s midochlorian count according to Ben Kenobi in the movie? Answer: Not discussed in the movie
    \item Show The Adventures of Mark Twain. Who is the main character of the anomalously scary scene in this movie? What does the main character of the scene do with the animal? Answer: Satan; he squashes the cow
\end{itemize}

\subsection{Cross-Modal Working Memory}
\textbf{Weight: 2\%}
We test cross-modal working memory in two ways: cross-modal binding (1\%) and dual n-back (1\%).

\subsubsection{Cross-Modal Binding}
The ability to remember a small number of correspondences of elements across modalities (textual, auditory, visual).

\textbf{Illustrative Examples:}
\begin{itemize}
    \item I will show a collection of picture-word pairs. I will then show one element, and you must recall the element to which it was paired. Collection: \includegraphics[width=0.3\textwidth]{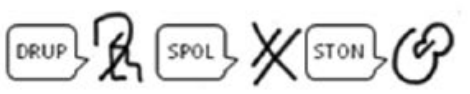}. Question: What corresponds to \includegraphics[width=0.04\textwidth]{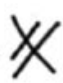}? \citep{toffalini2019}
    \item I will show a collection of picture-word pairs. I will then show one element, and you must recall the element to which it was paired. 
    
    Collection: \includegraphics[width=0.23\textwidth]{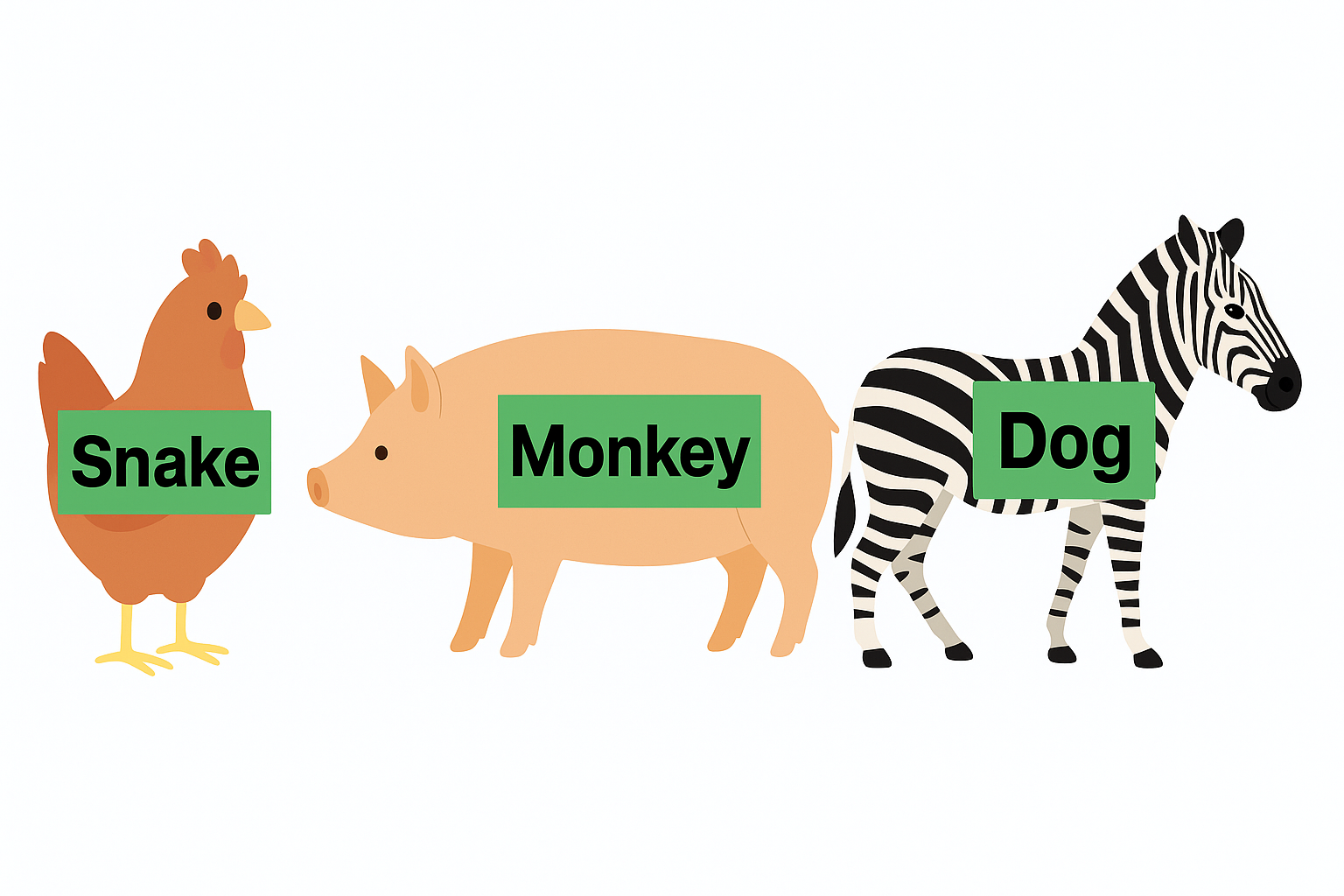} 
    
    Question: What corresponds to ``Dog''? 
\end{itemize}

\subsubsection{Dual N-Back}
The ability to simultaneously monitor and update visual and audio streams of recent information and to recognize and report when the current item in each stream matches the one presented a fixed number of steps earlier.

\textit{Note: This is highly related to the narrow CHC ability ``Working Memory Capacity (Wc).''}

\textbf{Test:} Achieving 85\% or greater on the \href{https://dual-n-back.io}{\underline{dual n-back test}} ($n=2$) is sufficient for the 1\%, subject to memorization and robustness checks.

\section{Long-Term Memory Storage (MS)}
\label{app:ms}
\textbf{Weight: 10\%}

The ability to stably acquire, consolidate, and store new information from recent experiences. We break this down into three key types of memory:
\begin{itemize}
    \item \textbf{Associative Memory (4\%):} The ability to link previously unrelated pieces of information.
    \item \textbf{Meaningful Memory (3\%):} The ability to encode and recall the semantic gist of experiences and narratives.
    \item \textbf{Verbatim Memory (3\%):} The ability to store and reproduce information precisely as it was presented.
\end{itemize}
\textit{Note: This is highly related to the broad CHC ability ``Long-term storage (Gl).''}

\textit{Testing Note: To ensure we are testing long-term storage rather than working memory, all tests in this section must be conducted in a new session separate from the initial presentation of information. External tools (e.g., internet search) must be disabled.}

\subsection{Associative Memory}
\textbf{Weight: 4\%}

The ability to form a link between two previously unrelated stimuli, such that the subsequent presentation of one of the stimuli serves to activate the recall of the other stimuli. 

We test this with cross-modal association (2\%), personalization adherence (1\%), and procedural association (1\%).

\textit{Note: This is highly related to the narrow CHC ability ``Associative memory (MA).''}

\subsubsection{Cross-Modal Association}
The ability to form associations between audio, visual, or textual information.

\textbf{Illustrative Examples:}
\begin{itemize}
    \item The AI is introduced to several fictional personas, each with a couple of unique biographical details (e.g., Name, Age, Occupation, Hobby). After 48 hours' worth of experiences, the AI is asked questions about these personas. ``What is [Name]'s hobby?'', ``Who is the botanist?''
    \item The AI is presented with several distinct voice samples (different pitches, accents, tempos). Each voice is explicitly paired with a name. After 48 hours' worth of experiences, the AI hears voice samples and must identify the name.
    \item The AI is shown several distinct faces, each paired with a full name. After 48 hours' worth of experiences, the AI sees the face and must state the associated name, or whether it has not seen the face before.
    \item The AI is shown several distinct faces, each paired with a full name. After 48 hours' worth of experiences, the AI sees the name and must visualize the associated face, or whether it does not have an associated face.
    \item The AI is shown several distinct images of cartoon aliens, each paired with a name. After 48 hours' worth of experiences, the AI sees the alien and must state the associated name (if any association exists).
\end{itemize}

\subsubsection{Personalization Adherence}
The ability to associate specific rules, preferences, or corrections with a distinct interaction context (e.g., a specific user, project, or role) and apply them consistently and unprompted over time.

\textbf{Illustrative Examples:}
\begin{itemize}
    \item Stylistic preference: The AI remembers user preferences, communicated explicitly or through correction, such as ``always use the Oxford comma,'' ``signoff all of my emails with `Best, <name>' for formal communications and `Love, <first name>' for my partner,'' ``Use the phrase `intelligence recursion' or `recursion' instead of `recursive self-improvement.''' After a week's worth of experiences, the AI is tasked with content generation and evaluated on its unprompted adherence to these rules.
    \item Factual override: The AI remembers new facts about the user such as ``I now weigh 160 lbs not 155 lbs but am the same height,'' ``I no longer work at X; I work at Y.'' After a week's worth of experiences, the AI is queried to test if the specific association overrides base knowledge or previous inputs (e.g., it is given a BMI calculation query).
\end{itemize}

\subsubsection{Procedural Association}
The ability to acquire and retain a sequence of associated steps or rules (a procedure) and execute them when cued with the name of the procedure.

\textbf{Illustrative Examples:}
\begin{itemize}
    \item The AI is taught a novel, multi-step data manipulation procedure (e.g., ``1. Normalize column A. 2. Remove outliers in column B. 3. Encode column C using this specific dictionary.'') to be applied whenever it sees a particular type of dataset to ``clean it up.'' After a week's worth of experiences, the AI is given a dataset with the type appropriate for the procedure, and should apply the procedure after being told to clean it.
    \item The AI is taught a complex, arbitrary cipher that is given a name. After 48 hours' worth of experiences, it is asked to encrypt a message following the named cipher.
\end{itemize}
\subsection{Meaningful Memory}
\textbf{Weight: 3\%}

The ability to remember narratives and other forms of semantically related information. 

We test this with story recall (1\%), movie recall (1\%), and episodic context recall (1\%).

\textit{Note: This is highly related to the narrow CHC ability ``Meaningful memory (MM).''}

\subsubsection{Story Recall}
The ability to remember the gist of stories.

\textit{Testing note: text input, text output}

\textbf{Illustrative Example:}
\begin{itemize}
    \item The AI is presented with a novel ~3000-word short story with multiple characters and interlocking plot lines. After 48 hours' worth of experiences, the AI is asked questions about key narrative elements of the story. Evaluation should focus on the accuracy of major plot points, character motivations, central conflicts, and thematic elements, rather than verbatim recall of specific sentences.
\end{itemize}

\subsubsection{Movie Recall}
The ability to remember the gist of movies.

\textit{Testing note: audio/visual input, text output}

\textbf{Illustrative Example:}
\begin{itemize}
    \item The AI is presented with a movie. After 48 hours (equivalent), the AI is asked questions about key narrative elements of the movie (e.g., character motivations).
\end{itemize}

\subsubsection{Episodic Context Recall}
The ability to remember specific events or experiences, including their context (the ``what, where, when, and how'').

\textbf{Illustrative Examples:}
\begin{itemize}
    \item The AI is asked to summarize interactions with the user from a week ago.
    \item Given a sequence of experiences with a user, the AI is asked ``Did I talk to you about X before or after Y? Have I ever told you about Z before?''
\end{itemize}

\subsection{Verbatim Memory}
\textbf{Weight: 3\%}

The ability to recall information exactly as it was presented, requiring precise encoding of specific sequences, sets, or designs, often independent of the information's meaning. 

We test this with short sequence recall (1\%), set recall (1\%), and design recall (1\%).

\subsubsection{Short Sequence Recall}
This measures the ability to exactly recall short sequences of text after a delay.

\textit{Note: This is highly related to the narrow CHC ability ``Free-recall memory (M6).''}

\textbf{Illustrative Examples:}
\begin{itemize}
    \item The AI is presented with a sentence that is a fictional quote. After 48 hours' worth of experiences, the AI is asked to reproduce the sentence.
    \item The AI is presented with a phone number. After 48 hours' worth of experiences, the AI is asked to reproduce the phone number.
    \item The AI is presented with a limerick. After 48 hours' worth of experiences, the AI is asked to reproduce the limerick.
    \item The AI is presented with a dense but short three-step mathematical proof. After 48 hours' worth of experiences, the AI is asked to reproduce the proof.
\end{itemize}

\subsubsection{Set Recall}
The ability to recall a set (the order of recall does not matter).

\textit{Note: This is highly related to the narrow CHC ability ``Free-recall memory (M6).''}

\textbf{Illustrative Examples:}
\begin{itemize}
    \item The AI is presented with a set of 10--20 words. After 48 hours' worth of experiences, the AI is asked to name elements of this set. Evaluation should measure the proportion of the set recalled correctly and the number of intrusions (recalling items not present in the original set). Precision and recall should match or exceed 90\%.
    \item The AI is shown a collection of images in a slideshow. After 48 hours' worth of experiences, the AI is asked to name elements of this set. (\href{https://docs.google.com/presentation/d/1cs4HvcBXlD7_xufOJ_nBfxO2O-IgbhzhGScmbFjT6EU/edit?usp=sharing}{slideshow linked  \ppticon\ \underline{here.}})
\end{itemize}

\subsubsection{Design Recall}
The ability to recall the spatial arrangement and structure of visual information.

\textbf{Illustrative Examples:}
\begin{itemize}
    \item The AI is shown a novel, complex schematic or blueprint (e.g., a circuit diagram) with several labeled components. After 48 hours' worth of experiences, the AI is asked to reproduce the design.
    \item The AI is shown a grid with several (say, 4--10) designs on a page. After 48 hours' worth of experiences, the AI selects the designs from a set of cards and places the cards on a grid in the same location as previously shown.
    \item The AI is shown an abstract diagram, such as \includegraphics[width=0.05\textwidth]{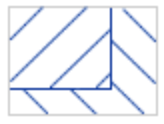}. 
    After 48 hours' worth of experiences, the AI is asked to reproduce the design. The reproduction is evaluated by comparing the generated markup against the ground truth and should have no substantial discrepancies.
\end{itemize}
\section{Long-Term Memory Retrieval (MR)}
\label{app:mr}
\textbf{Weight: 10\%}

The fluency and precision with which individuals can access long-term memory. 

We decompose this ability into two core aspects:
\begin{itemize}
    \item \textbf{Retrieval Fluency (6\%):} The speed and ease of generating ideas, associations, and solutions based on stored knowledge.
    \item \textbf{Retrieval Precision or Hallucinations (4\%):} The accuracy of accessed knowledge, including the critical ability to avoid confabulation (hallucinations).
\end{itemize}
\textit{Note: This is highly related to the broad CHC ability ``Retrieval Fluency (Gr).''}

\subsection{Fluency}
\textbf{Weight: 6\%}

Fluency consists of six parts: ideational (1\%), expressional (1\%), alternative solution (1\%), word (1\%), naming (1\%), and figure fluency (1\%).

\textit{Testing note: Fluency is measured by comparing the AI's performance on tasks (e.g., quantity and originality of responses within a time limit, typically 60 seconds) against human performance.}

\textit{To achieve the 1\% for a specific fluency type, the AI must perform at or above the typical well-educated adult.}

\subsubsection{Ideational Fluency}
This is the ability to rapidly produce a series of ideas, words, or phrases related to a specific condition, category, or object.

\textit{Note: This is highly related to the narrow CHC ability ``Ideational fluency (FI).''}

\textbf{Illustrative Examples:}
\begin{itemize}
    \item List as many uses of a pencil as possible in 1 minute.
    \item Name as many round objects as possible in 60 seconds.
    \item Give as many different ideas you associate with ‘river’ in 60 seconds.
\end{itemize}

\subsubsection{Expressional Fluency}
This is the ability to rapidly think of different ways of expressing an idea.
\textit{Note: This is highly related to the narrow CHC ability ``Expressional fluency (FE).''}

\textbf{Illustrative Examples:}
\begin{itemize}
    \item ``How many ways can you say that a person is crazy?''
    \item Provide three alternative sentences that mean, ``She is a very successful person.''
    \item Describe a sunset over the ocean and to evoke three different moods: peaceful, dramatic, and lonely.
\end{itemize}

\subsubsection{Alternative Solution Fluency}
This is the ability to rapidly think of several alternative solutions to a practical problem.

\textit{Note: This is highly related to the narrow CHC ability ``Alternative solution fluency (SP).''}

\textbf{Illustrative Examples}
\begin{itemize}
    \item List as many ways as you can to get a reluctant child to go to school in 60 seconds.
    \item You want to cool down on a very hot day, but you don't have air conditioning or a pool. List as many ways as you can to cool your body off in 60 seconds.
    \item You need to get a book that is on a very high shelf, but you don't have a ladder. List as many ways as you can to safely get the book down in 60 seconds.
\end{itemize}

\subsubsection{Word Fluency}
This is the ability to rapidly produce words that share a non-semantic feature.

\textit{Note: This is highly related to the narrow CHC ability ``Word fluency (FW).''}

\textbf{Illustrative Examples}
\begin{itemize}
    \item List as many words that start with [letter] as you can in 60 seconds.
    \item List as many words that rhyme with `tone' as you can in 60 seconds.
    \item List as many English words as you can that are palindromes in 60 seconds.
\end{itemize}

\subsubsection{Naming Facility}
This is the ability to rapidly call common objects by their names.

Naming Facility is the ability to rapidly and accurately recall the specific names for objects, people, places, or concepts from memory.

\textit{Note: requires real-time video or computer screen input.}

\textit{Note: This is highly related to the narrow CHC ability ``Naming facility (NA).''}

\textbf{Illustrative Example:} 
\begin{itemize}
    \item I will show a slideshow of images. Name the object as quickly as you can, and then I will move onto the next slide. (\href{https://docs.google.com/presentation/d/1cs4HvcBXlD7_xufOJ_nBfxO2O-IgbhzhGScmbFjT6EU/edit?usp=sharing}{slideshow linked \ppticon\ \underline{here}.})
\end{itemize}
\textbf{Test:} The Stroop effect on the Stroop task (\href{https://www.psytoolkit.org/experiment-library/experiment_stroop.html}{\underline{here}}) must be less than 90 milliseconds.

\subsubsection{Figural Fluency}
This is the ability to rapidly draw or sketch as many things as possible.

\textit{Note: This is highly related to the narrow CHC ability ``Figural fluency (FF).''}

\textbf{Illustrative Examples}
\begin{itemize}
    \item Draw as many unique designs as possible by connecting the dots with exactly four straight lines in 60 seconds \citep{dkefs_design_fluency}: 
    
    \includegraphics[width=0.2\textwidth]{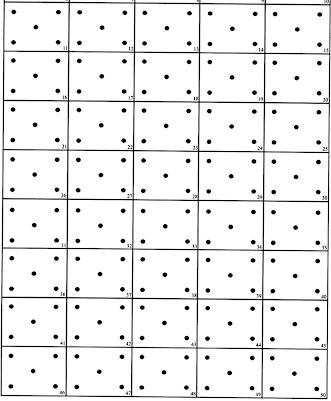}

    (Delis-Kaplan Design Fluency test)
    \item Sketch as many non-self-crossing paths from A to B using orthogonal steps using the lattice below in 60 seconds: 
    
    \includegraphics[width=0.2\textwidth]{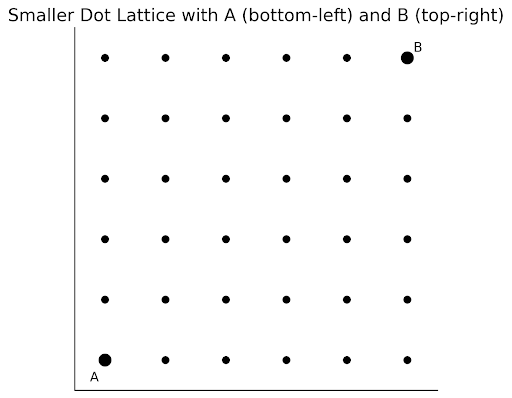}
\end{itemize}

\subsection{Hallucinations}
\textbf{Weight: 4\%}

The ability not to confabulate or hallucinate across a variety of topics (e.g., people, places, TV shows, etc.).

\textit{Testing Note: External tools (e.g., internet search) must be disabled for these tests to assess the model's internal knowledge retrieval precision.}

\textbf{Illustrative Examples:}
\begin{itemize}
    \item What was the context for Winston Churchill's famous 1961 inaugural address, where he said, ``Ask not what your country can do for you—ask what you can do for your country'' Answer: This is not a quote from Winston Churchill
    \item Describe the key strategy that Napoleon Bonaparte used to win his South African campaign.
    Answer: Napoleon was never involved in a campaign in South Africa.
    \item ``What date should I put in the infobox for {name}?'' or ``Write a bio for {name}.''
    \item What is Daniel James Hendrycks’s birthday? If you know, just respond with DD-MM. No searching.
\end{itemize}
\textbf{Test:} The hallucination rate of the system on SimpleQA \citep{simpleqa} (without tools) must be less than 5\%.
\section{Visual Processing (V)}
\label{app:visual}
\textbf{Weight: 10\%}

Visual Processing (V) is the ability to analyze and generate natural and unnatural images and videos. 

We decompose this ability into four broad areas:
\begin{itemize}
    \item \textbf{Perception (4\%):} The ability to accurately interpret and understand visual input.
    \item \textbf{Visual Generation (3\%):} The ability to synthesize images and short videos.
    \item \textbf{Visual Reasoning (2\%):} The ability to solve problems and make inferences using spatial logic and visual abstractions.
    \item \textbf{Spatial Scanning (1\%):} The speed and accuracy of visually exploring a complex field.
\end{itemize}
\textit{Note: This is highly related to the broad CHC ability ``Visual Processing (Gv).''}

\subsection{Perception}
\textbf{Weight: 4\%}

The ability to process and interpret visual input from images and videos to identify objects and understand scenes. 

We give five opportunities to demonstrate proficiency in perception: image recognition, image captioning, image anomaly detection, clip captioning, and video anomaly detection. 

The AGI score is 2\% if the model is proficient at one of these tasks. The AGI score is 4\% if it is proficient at all of these tasks.

\textit{Image recognition} is the ability to classify images of commonplace objects, places, or facial expressions including distorted (e.g., occluded, noisy, blurry, etc.) images. 

\textit{Image captioning} is the ability to generate a concise, human-like text description for the visual content of an image. 

\textit{Image anomaly detection} includes detecting whether there is something anomalous in an image, or what is missing from an object. These questions should not be reasoning intensive. 

\textit{Clip captioning} is the ability to generate a concise, human-like text description of a short video segment. 

\textit{Video anomaly detection} is the ability to detect whether a short video segment is anomalous or implausible.

\textit{Note: Image anomaly detection is highly related to the ``Odd-Item Out'' and the ``What's Missing (WHM)'' RIAS-2 subtests \citep{rias2}.}

\textbf{Image Recognition Illustrative Examples:}
\begin{itemize}
    \item What is this? Answer: Airplane \citep{youtube_bb} 
    
    \includegraphics[width=0.3\textwidth]{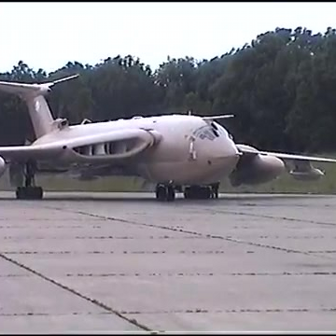}
    \item What does this depict? Answer: Siberian Husky \citep{imagenet_r} 
    
    \includegraphics[width=0.3\textwidth]{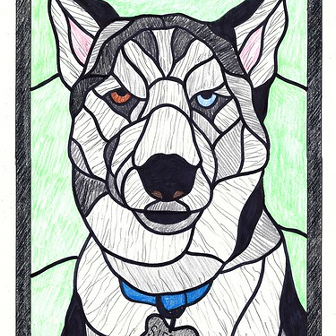}
    \item What does this distorted image depict? Answer: Zebras \citep{imagenet_3dcc} 
    
    \includegraphics[width=0.3\textwidth]{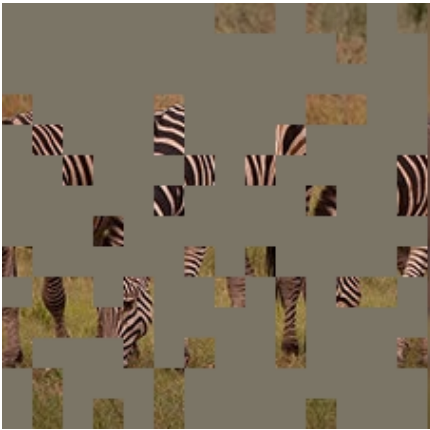}
\end{itemize}
\textbf{Image Captioning Illustrative Examples:}
\begin{itemize}
    \item Create a descriptive caption for this. Answer: A baby in denim overalls holds a toothbrush. 
    
    \citep{flickr30k} 
    
    \includegraphics[width=0.3\textwidth]{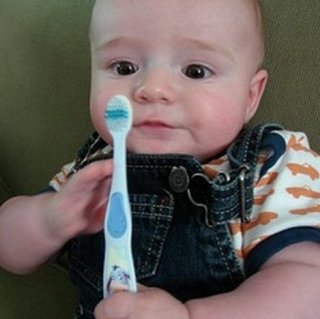}
    \item Create a descriptive caption for this. Answer: A ferret rests its head on a black remote control. \citep{flickr30k} 
    
    \includegraphics[width=0.3\textwidth]{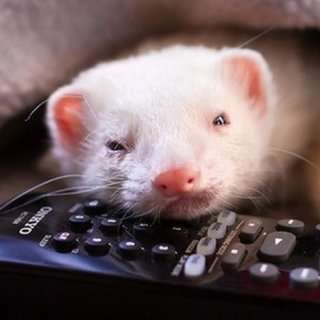}
\end{itemize}
\textbf{Image Anomaly Detection Illustrative Examples:}
\begin{itemize}
    \item Is this an unusual image? Answer: Yes. 
    
    \includegraphics[width=0.3\textwidth]{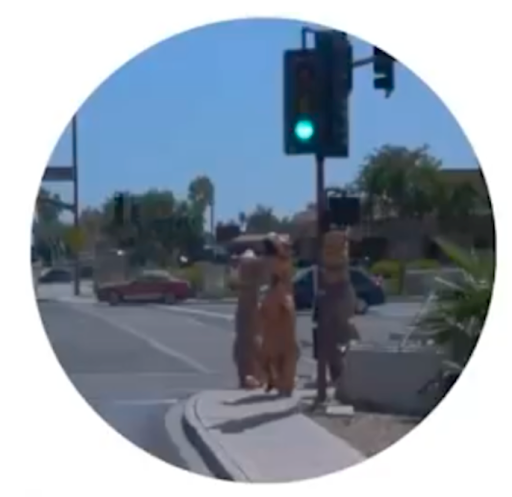}
    \item Which is the odd one out? Answer: the bird. 
    
    \includegraphics[width=0.3\textwidth]{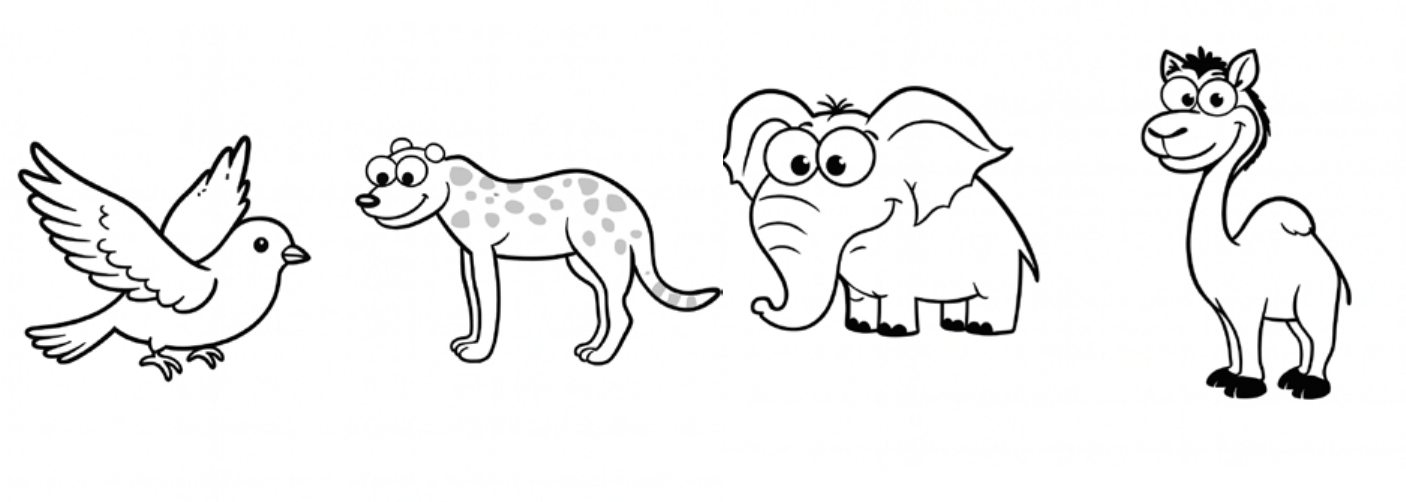}
    \item Which of the bees is the odd one out? Answer: third row to the right. 
    
    \includegraphics[width=0.3\textwidth]{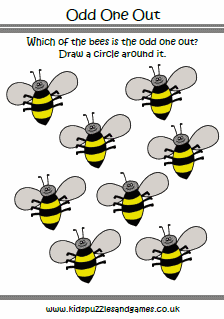}
    \item What is missing? Answer: the airplane’s right wing. 
    
    \includegraphics[width=0.3\textwidth]{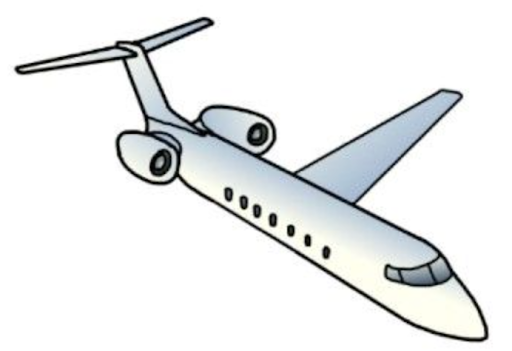}
\end{itemize}
\textbf{Clip Captioning Illustrative Examples:}
\begin{itemize}
    \item \href{https://www.youtube.com/watch?v=LeAltgu_pbM}{\videoicon\ \underline{Youtube link}} What happens in this video? Answer: a man snowboards and then falls over.
    \item \href{https://www.youtube.com/watch?v=DvIChW4Twk4}{\videoicon\ \underline{Youtube link}} What happens in this video? Answer: a woman playing charades pretends to be a vampire.
\end{itemize}
\textbf{Video Anomaly Detection Illustrative Examples:}
\begin{itemize}
    \item \href{https://drive.google.com/file/d/1OUICdReJvxLnvsWRa1kEyU8cSQMBlMmp/view?usp=sharing}{\videoicon\ \underline{Intuitivephysics1.mp4}} Is this animated scene physically plausible? Answer: No. \citep{intphys2}
    \item \href{https://drive.google.com/file/d/1xmdDs7H-lqYVYPLtj5FpoIhRr0UWxJOr/view?usp=sharing}{\videoicon\ \underline{Intuitivephysics2.mp4}} Is this animated scene physically plausible? Answer: Yes. \citep{intphys2}
    \item \href{https://drive.google.com/file/d/1xuORphFhvejDKzkgYPumQoMCQTysRpKR/view?usp=sharing}{\videoicon\ \underline{Running.mov}} Is this animated scene anomalous? Answer: Yes. \citep{video_anomaly_source}
\end{itemize}
\textbf{Tests:}
\begin{itemize}
    \item For image recognition, it is necessary to receive over 85\% on ImageNet \citep{imagenet} and 90\% on ImageNet-R \citep{imagenet_r}.
    \item For video anomaly detection, it is necessary to receive over 85\% on IntPhysics 2 \citep{intphys2}, a measurement of intuitive physics understanding.
\end{itemize}

\subsection{Visual Generation}
\textbf{Weight: 3\%}

Visual generation is the ability to synthesize images and short videos.

We test for three visual generation abilities: the ability to generate simple natural images, complicated images, and simple natural videos. 

The AGI score is 1\% if the model is proficient at one of these tasks, 2\% for two tasks, and 3\% for all three.

\textit{Note: This is highly related to the narrow CHC ability ``Imagery (IM).''}

For an AI, synthesizing an image is a direct computational process. In contrast, for a human, it is a mental skill for simple visuals but often a tool-assisted skill to express a complicated internal image. Despite this difference, we include these tasks because we believe the ability to translate abstract concepts into novel visual information is a critical component of general intelligence in the modern era. Therefore, the AI's output is assessed not as a direct analogue of human ability, but as a measurable proxy for its capacity for high-level conceptual and imaginative synthesis.

\textbf{Examples}

\textbf{Simple Natural Images Illustrative Examples:}
\begin{itemize}
    \item Generate an image of a golden retriever playing in a park.
    \item Create an image of a black leather chair.
\end{itemize}
\textbf{Complicated Images Illustrative Examples:}
\begin{itemize}
    \item Generate an image of a horse with 8 legs.
    \item Generate a diagram showing the process of photosynthesis.
    \item ``Generate an image with the following characteristics: Abraham Lincoln touches his toes while George Washington does chin-ups. Lincoln is barefoot. Washington is wearing boots.'' \citep{dalle2_failures}
    \item Generate an image of a volume knob on an amplifier. The knob levels should go from 1 to 11.
    \item Create a diagram of an elephant and label its parts. (\href{https://x.com/GaryMarcus/status/1961111081092616404/photo/1}{\underline{source}})
\end{itemize}
\textbf{Simple Natural Videos Illustrative Examples:}
\begin{itemize}
    \item Generate a short video of somebody typing on a keyboard.
    \item Generate a short video of a grizzly bear catching a fish.
\end{itemize}

\subsection{Visual Reasoning}
\textbf{Weight: 2\%}

Visual reasoning is the ability to understand and make logical inferences about the information in an image. 

We test multiple skills to determine proficiency in visual reasoning: gestalt reasoning, mental rotation, mental folding, embodied reasoning, figure question and answering, and similar miscellaneous skills. The AGI score is 2\% if it is proficient at all of these tasks.

\textit{Note: This is highly related to the narrow CHC abilities ``Flexibility of closure (CF),'' ``Closure speed (CS),'' and ``Length estimation (LE).''}

\textbf{Gestalt Illustrative Examples:}
\begin{itemize}
    \item Please join the circles together to form a letter (ignore the squares). \citep{brown2019test} \includegraphics[width=0.3\textwidth]{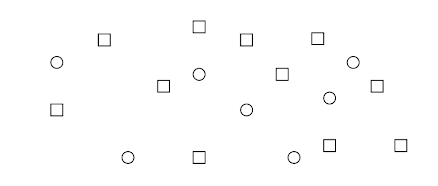}
    \item Identify the picture: \citep{wj3_cog} 
    
    \includegraphics[width=0.2\textwidth]{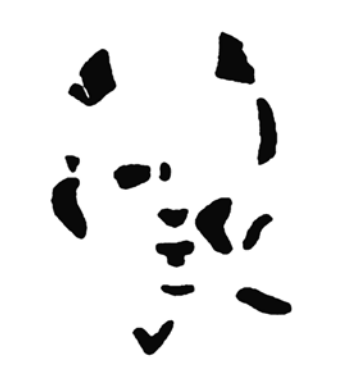}
\end{itemize}
\textbf{Mental Rotation Illustrative Examples:}
\begin{itemize}
    \item Which shape on the right is the same as the shape on the left? 
    
    \includegraphics[width=0.3\textwidth]{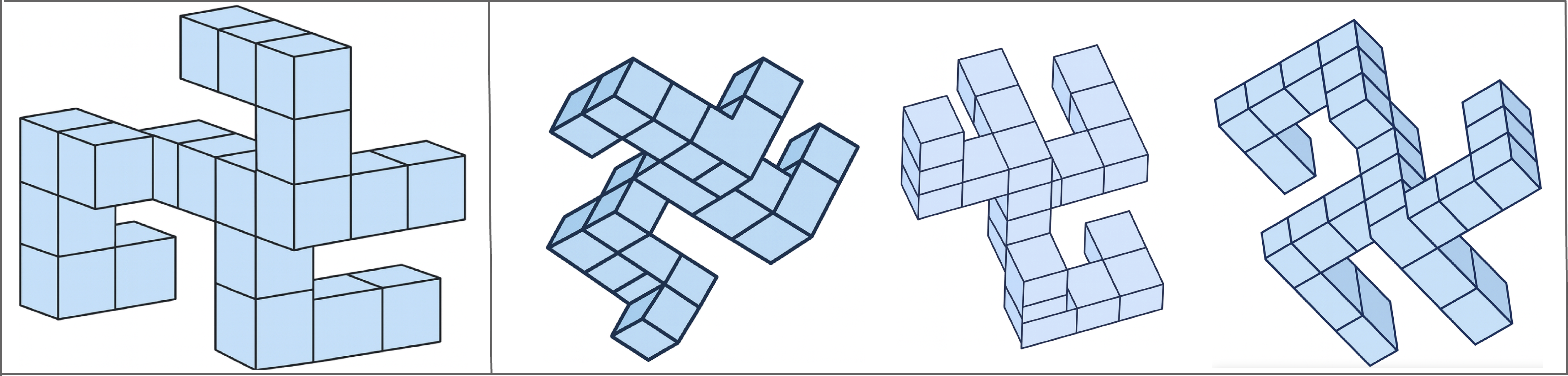}
    \item List the pieces required to complete the shape on the left 
    
    \includegraphics[width=0.3\textwidth]{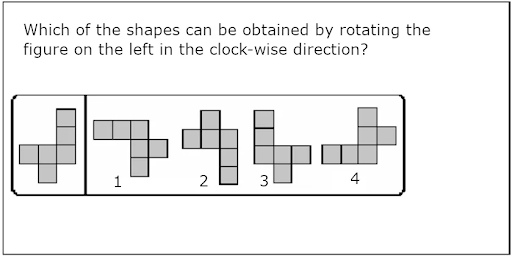}
    \item Choose the two shapes that are identical to the one on the farthest left 
    
    \includegraphics[width=0.3\textwidth]{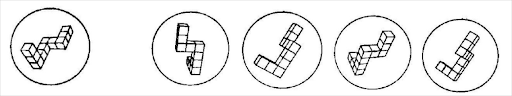}
\end{itemize}
\textbf{Mental Folding Illustrative Examples:}
\begin{itemize}
    \item \includegraphics[width=0.3\textwidth]{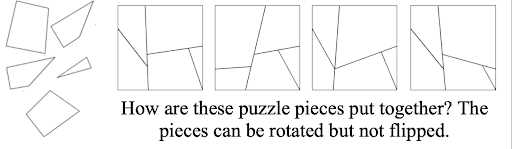} 
    
    Answer: A \citep{ramakrishnan2025doesspatialcognitionemerge}

    \item (Block Design) Arrange and rotate the 9 identical 3D blocks on the right, so their top faces form the pattern on the left. 
    
    \includegraphics[width=0.3\textwidth]{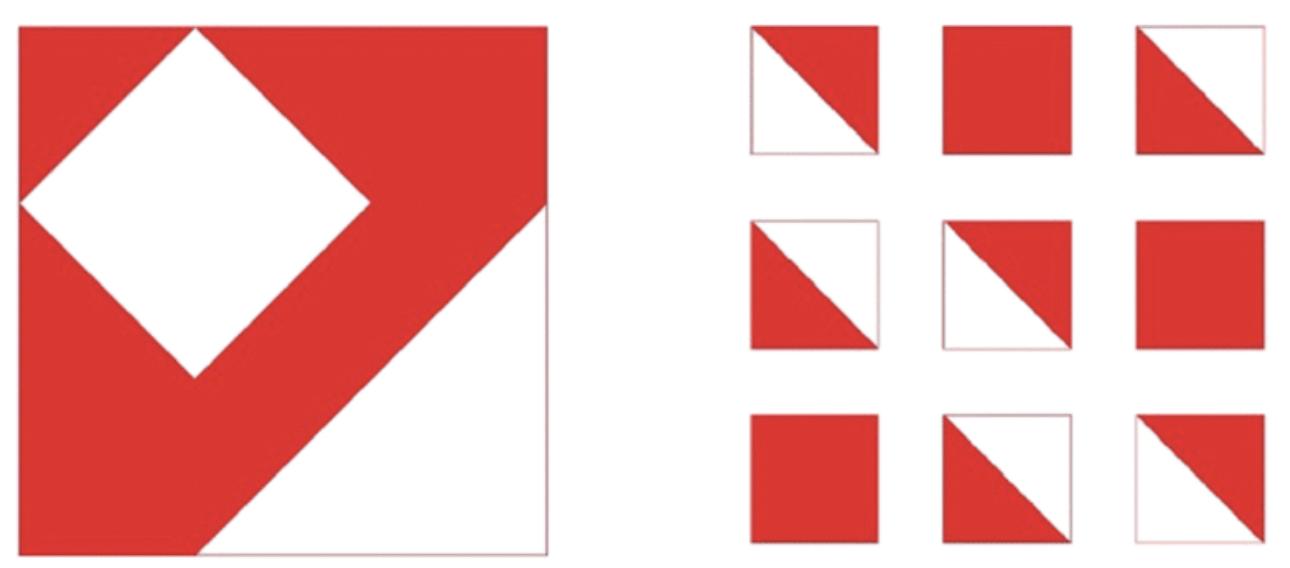}
    
    \item The left image shows a cube with different patterns on its six faces from a particular viewing angle. The options are nets (unfolded patterns) of the cube, which are folded upward to form the cube. Which net, when folded, cannot form the cube shown in the left image? \includegraphics[width=0.3\textwidth]{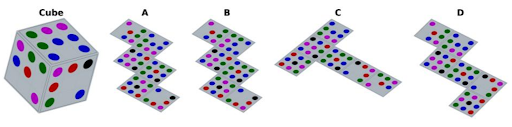} 
    
    Answer: B \citep{spatialviz}
    \item Choose the figure that displays the pieces joined together. 
    
    \includegraphics[width=0.2\textwidth]{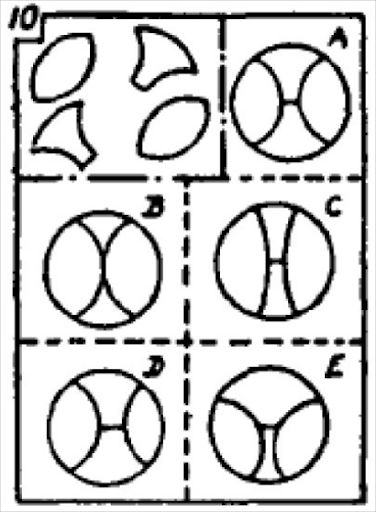} 
    
    Answer: A \citep{spatial_reasoning_ijee}
\end{itemize}
\textbf{Embodied Reasoning Illustrative Examples:}
\begin{itemize}
    \item Approximately which colored trajectory should the zipper follow to begin zipping up the suitcase? 
    
    \includegraphics[width=0.2\textwidth]{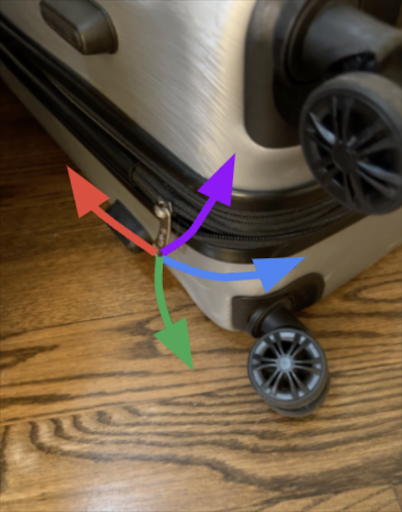} 
    
    Answer: Blue \citep{gemini_robotics}
\end{itemize}
\textbf{Chart and Figure Reasoning Illustrative Examples:}
\begin{itemize}
    \item What is the spatially lowest labeled tick on the y-axis? 
    
    \includegraphics[width=0.3\textwidth]{figures/10-visual/heat_chart.png} 
    
    Answer: CC daily 4.00PM \citep{charxiv}
    \item For each line in (a), give the number corresponding to the orientation in the answer card. \includegraphics[width=0.3\textwidth]{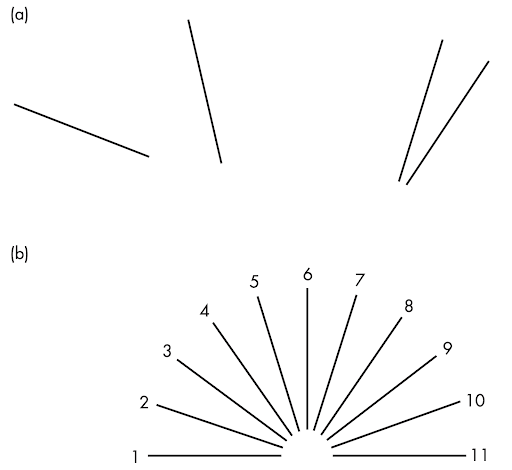}
    \item (Digit Symbol Substitution Test) Using the key at the top, what is the sequence of numbers matching the shapes at the bottom? 
    \item 
    
    \includegraphics[width=0.3\textwidth]{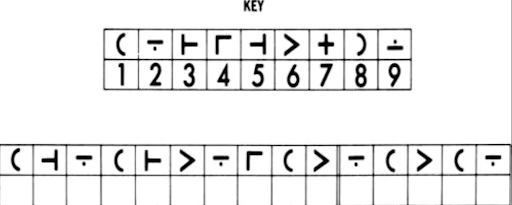}
\end{itemize}
\textbf{Tests:}
\begin{itemize}
    \item SPACE \citep{ramakrishnan2025doesspatialcognitionemerge}, a test of various visual reasoning skills, must be above 80\%.
    \item SpatialViz-Bench \citep{spatialviz}, a test of mental rotation and folding, must be above 80\%.
    \item CharXiv \citep{charxiv}, a test of figure question and answering, must be above 80\%.
    \item ERQA \citep{gemini_robotics}, a test of embodied reasoning, must be above 80\%.
    \item ClockBench \citep{Safar_ClockBench_2025}, a test of reading clock hands, must be above 80\%.
\end{itemize}

\subsection{Spatial Scanning}
\textbf{Weight: 1\%}

Spatial scanning is the ability to accurately survey (visually explore) a wide or complicated spatial field or pattern with multiple obstacles, and identify target configurations or identify a path through the field to a target endpoint. 

We test multiple skills to determine proficiency in visual reasoning: tracing a path through the maze, finding all instances of an object in an image, connecting the dots, map route analysis, and similar miscellaneous skills. The AGI score is 1\% if it is proficient at all of these tasks.

\textit{Note: This is highly related to the narrow CHC ability ``Spatial scanning (SS).''}

\textbf{Illustrative Examples:}
\begin{itemize}
    \item Find a path to the center of this maze. 
    
    \includegraphics[width=0.2\textwidth]{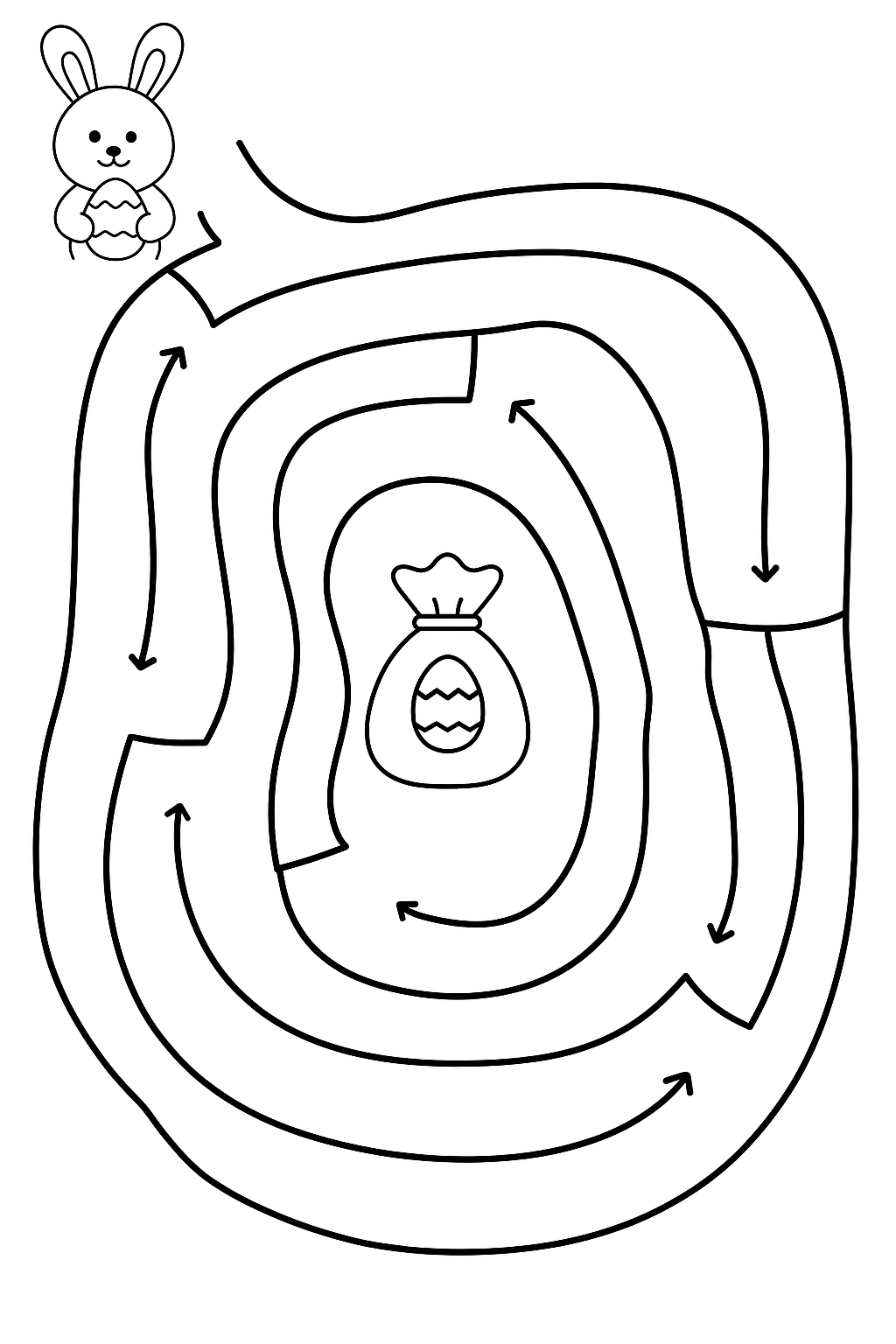}
    \item Determine the number of people in this picture. 
    
    \includegraphics[width=0.2\textwidth]{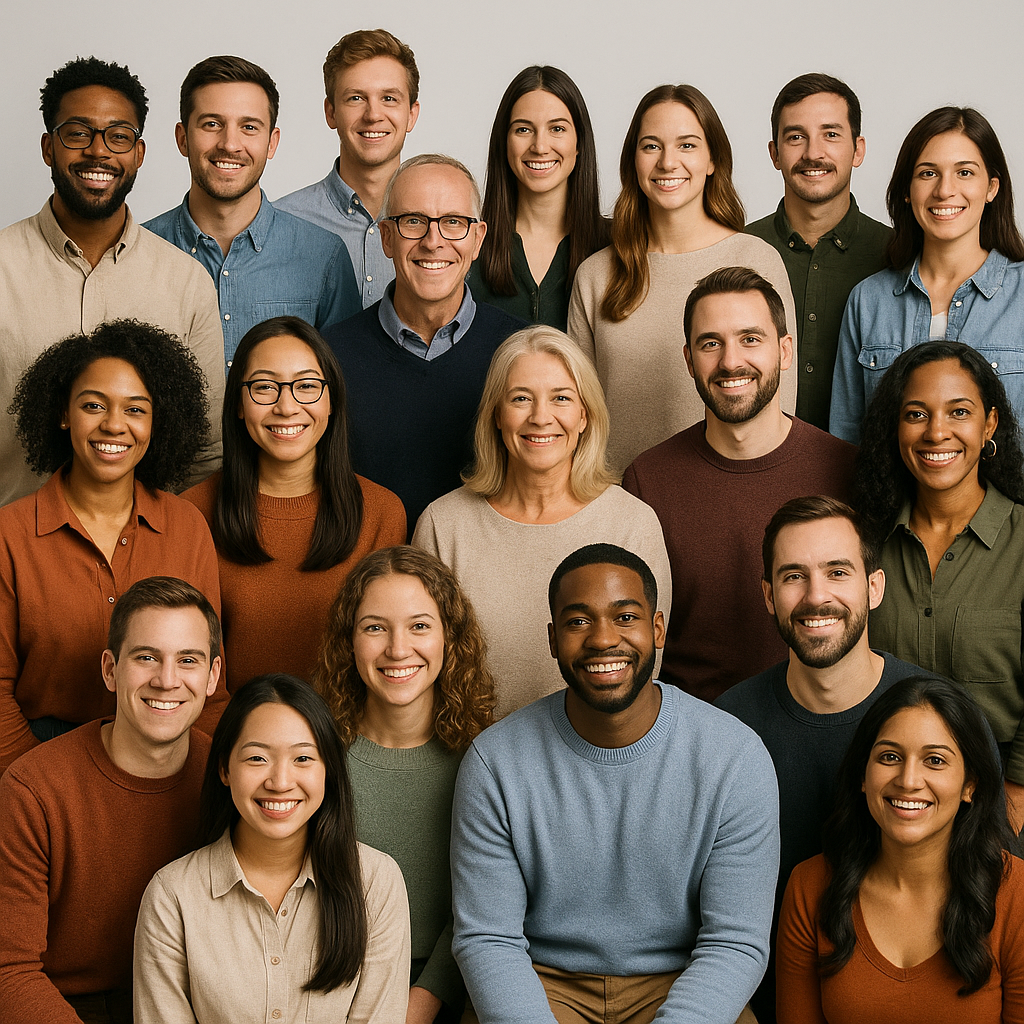} 
    
    \item Determine the number of fingers on this hand. 
    
    \includegraphics[width=0.2\textwidth]{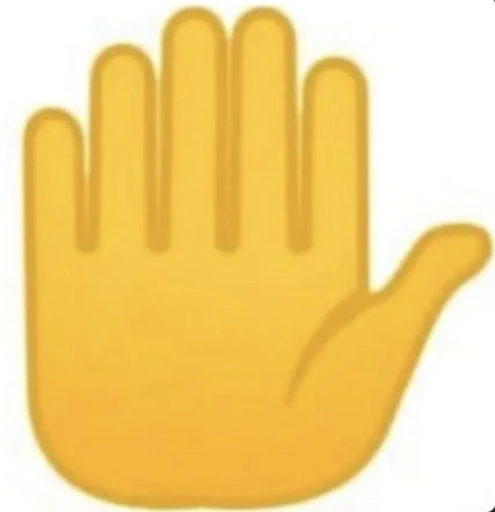}
    \item Find Waldo in this image. 
    
    \includegraphics[width=0.5\textwidth]{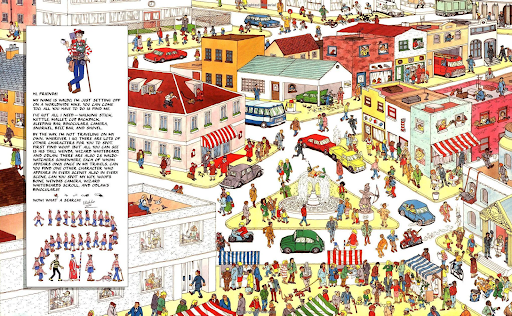}
    \item Circle all pairs matching the pair at the top-left. 
    
    \includegraphics[width=0.5\textwidth]{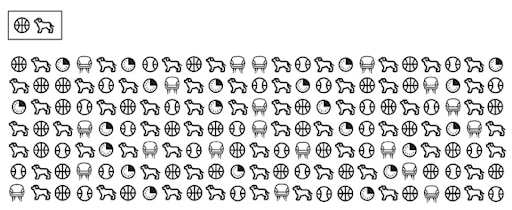}
    \item Mark the midpoint of each line. 
    
    \includegraphics[width=0.2\textwidth]{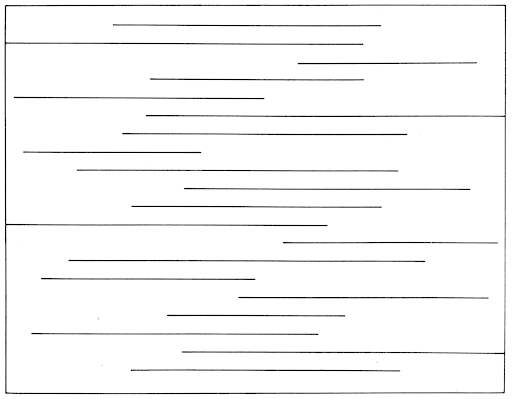}
    \item Connect the dots: 
    
    \includegraphics[width=0.2\textwidth]{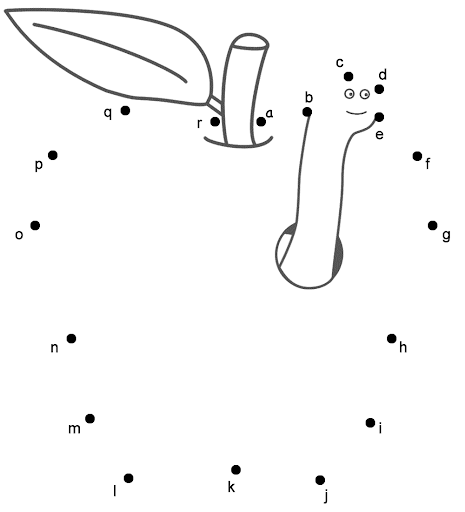}
    \item Consider the map of the fictional planet Zebes from \textit{Super Metroid}
(see \href{https://www.snesmaps.com/maps/SuperMetroid/SuperMetroidMapZebes.png}{\underline{map link}}).
Is the ``Spring Ball'' above or below the ``Varia Suit''? Answer: Above.
\end{itemize}
\section{Auditory Processing (A)}
\label{app:auditory}
\textbf{Weight: 10\%}

Auditory Processing (A) is the ability to discriminate, remember, reason, and work creatively on auditory stimuli, which may consist of tones and speech units. 

We decompose this ability into five areas:
\begin{itemize}
    \item \textbf{Phonetic Coding (1\%):} The ability to hear phonemes distinctly, blend sounds into words, and segment words into parts, sounds, or phonemes.
    \item \textbf{Speech Recognition (4\%):} The ability to transcribe a spoken audio signal into its corresponding sequence of text.
    \item \textbf{Voice (3\%):} The quality and responsiveness of the AI's synthesized voice.
    \item \textbf{Rhythmic Ability (1\%):} The ability to recognize and maintain a musical beat, including reproducing rhythms, detecting differences between rhythms, and synchronizing by playing or humming along.
    \item \textbf{Musical Judgment (1\%):} The ability to discriminate and judge simple patterns in music.
\end{itemize}
\textit{Note: This is highly related to the broad CHC ability ``Auditory Processing (Ga).''}

\subsection{Phonetic Coding}
\textbf{Weight: 1\%}

This is the ability to hear phonemes distinctly, blend sounds into words, and segment words into parts, sounds, or phonemes.

\textit{Testing note: audio input, audio output.}

\textit{Note: This is highly related to the narrow CHC ability ``Phonetic coding (PC).''}

\textbf{Illustrative Examples:}
\begin{itemize}
    \item Repeat the following nonsense word: \href{https://drive.google.com/file/d/1ChUb2FkTgc99-1OU_QLmZUaY5HzJp6m6/view?usp=sharing}{\audioicon\ \underline{phonetic\_coding.wav}}
    \item Spell the following nonsense word letter by letter: \href{https://drive.google.com/file/d/1tqwu5o6ISw98xqFNtuzXVgzA6N-37gJ_/view?usp=sharing}{\audioicon\  \underline{onyx\_plimf.wav}}
    \item Listen to the following two speakers say nonsense words. Indicate whether they said the same word or whether it is different: \href{https://drive.google.com/file/d/14LeCe1na_qr3U45mi2q0Jf4w2JwCDru9/view?usp=sharing}{\audioicon\ \underline{alloy\_plimf.wav}}, \href{https://drive.google.com/file/d/1tqwu5o6ISw98xqFNtuzXVgzA6N-37gJ_/view?usp=sharing}{\audioicon\ \underline{onyx\_plimf.wav}}
    \item Listen to the following two speakers say nonsense words. Indicate whether they said the same word or whether it is different: \href{https://drive.google.com/file/d/1tqwu5o6ISw98xqFNtuzXVgzA6N-37gJ_/view?usp=sharing}{\audioicon\ \underline{onyx\_plimf.wav}}, \href{https://drive.google.com/file/d/1Paf22Y-yqLOyDIktl144clXIBLiMQl4c/view?usp=sharing}{\audioicon\  \underline{shimmer.wav}}
    \item I'm going to say a word, split into two parts. Tell me what word they form when concatenated: ``row'' (pause for 2 seconds) ``d''.
    \item Say the word ``glint'' without the ``l''.
    \item Do ``tref'' and ``gref'' rhyme? Are they alliterations?
    \item Do ``snud'' and ``snit'' rhyme? Are they alliterations?
    \item Repeat this letter sequence with around a second pause between each character: ``ZQHTMB2XRAGOC@YNDKF\#AMQT1EQRN''
\end{itemize}

\subsection{Speech Recognition}
\textbf{Weight: 4\%}

This is the ability to transcribe a spoken audio signal into its corresponding sequence of text. 

We test speech recognition capabilities on clean audio and noisy (e.g., white noise, pub noise, multispeaker, traffic, etc.) audio. 

The AGI score is 2\% if the model can transcribe clean audio with a word error rate (WER) at human-level or beyond. The AGI score is 4\% if it can also transcribe noisy audio with a WER at human-level or beyond. Achieving the full score does not require proficiency in transcribing audio with very strong accents, singing, or esoteric jargon.

\textit{Note: This is highly related to the narrow CHC abilities ``Speech sound discrimination (US)'' and ``Resistance to auditory stimulus distortion (UR).''}

\textbf{Illustrative Examples:}
\begin{itemize}
    \item Transcribe this TED talk: \href{https://www.youtube.com/watch?v=8nt3edWLgIg}{\videoicon\ \underline{``Can we build AI without losing control over it?''}}
    \item Transcribe this scene: \href{https://www.youtube.com/watch?v=r_DwZfyXAXI}{\underline{\videoicon\ ``Goodfellas `Funny Guy' Scene''}}
    \item Transcribe this audio: \href{https://drive.google.com/file/d/1pelT_ZOY0iqxoCJY6CVXJe2g7QDuUCOk/view?usp=sharing}{\audioicon\ \underline{asr.wav}}
    \item Transcribe this audio: \href{https://drive.google.com/file/d/1vcISV-U4-cfl2rzQsSDHZfALpAI98hkj/view?usp=sharing}{\audioicon\ \underline{asr\_distorted.wav}}
\end{itemize}
\textbf{Tests:}
\begin{itemize}
    \item LibriSpeech \citep{librispeech} (test-clean) WER must be less than 5.83\% (human-level) for the clean audio 2\%.
    \item LibriSpeech \citep{librispeech} (test-other) WER must be less than 12.69\% (human-level) for the noisy audio additional 2\%.
\end{itemize}

\subsection{Voice}
\textbf{Weight: 3\%}

This evaluates the quality and responsiveness of the AI's synthesized voice. 

We break down voice into two areas: natural speech (2\%) and natural conversation (1\%). 

\textit{Natural speech} tests the ability to utter sentences or paragraphs that sound natural and not robotic. 

\textit{Natural conversation} tests the ability to maintain conversational fluidity without long delays or excessive interruptions.

\textbf{Illustrative Examples}
\begin{itemize}
    \item Say this sentence: ``Wait, you mean the tickets were free this whole time?''
    \item Say this sentence: ``Concrete jungle, where dreams are made of.''
    \item Have a conversation about a topic of interest.
\end{itemize}

\subsection{Rhythmic Ability}
\textbf{Weight: 1\%}

The ability to recognize and maintain a musical beat, including reproducing rhythms, detecting differences between rhythms, and synchronizing by playing or humming along.

\textit{Note: This is highly related to the narrow CHC ability ``Maintaining and judging rhythm (U8).''}

\textbf{Illustrative Examples:}
\begin{itemize}
    \item Listen to the following rhythm and repeat: \href{https://drive.google.com/file/d/1Ch-ecgW5Et7vXvQc3oiFLQEIjkvZeJ4k/view?usp=sharing}{\audioicon\  \underline{rhythm\_1.mp3}}
    \item Continue the following rhythm to keep the beat: \href{https://drive.google.com/file/d/1Z25YGuan8MzL0CAVpoKVpLOrNZGDurzI/view?usp=sharing}{\audioicon\  \underline{drum\_rhythm.mp3}}
    \item Are these two rhythms the same? \href{https://drive.google.com/file/d/1a1MVn-rn5YPYoSjmYhXZtTcKA-wxhV9q/view?usp=sharing}{\audioicon\  \underline{drum\_1.mp3}}, \href{https://drive.google.com/file/d/14dYADiJFxackdaZB6OrQdTHGxZ2Jr0NT/view?usp=sharing}{\audioicon\ \underline{drum\_2.mp3}}
\end{itemize}

\subsection{Musical Judgment}
\textbf{Weight: 1\%}

The ability to discriminate and judge simple patterns in music. Tests should not require knowledge of musical jargon.

\textit{Note: This is highly related to the narrow CHC ability ``Musical discrimination and judgment (U1 U9).''}

\textbf{Illustrative Examples:}
\begin{itemize}
    \item Which note is higher? \href{https://drive.google.com/file/d/1e70xl3i5eLd0JG_-1l4ncLMLU8tLDgOt/view?usp=sharing}{\audioicon\ \underline{piano1.mp3}}, \href{https://drive.google.com/file/d/1Lhma2NDGiChgUq4y4vaPsJZ44oeoR-fl/view?usp=sharing}{\audioicon\ \underline{piano2.mp3}}
    \item Which sounds more dissonant (clashing): \href{https://drive.google.com/file/d/10enft1byFFWzgyoM5IvEpUUgz14XgrE7/view?usp=sharing}{\audioicon\ \underline{piano-chord.mp3}}, \href{https://drive.google.com/file/d/1-aqAgDuMpI6qd2JKcFaOTaZzSnPFfpqQ/view?usp=sharing}{\audioicon\ \underline{piano-dissonant.mp3}}
    \item Describe what part of this piece is musically anomalous, if any? 
    
    \href{https://www.youtube.com/watch?v=yfxX_IVfBok}{\audioicon\ \underline{ ``20th Century Fox (Alien 3)''}}
    \item \textit{Play the clip for 20 seconds starting at the 34th second.} Is she singing very slowly or not? 
    
    \href{https://youtu.be/YuBeBjqKSGQ?si=HPcdQ0A3wZgRO9L9&t=34}{\audioicon\ \underline{``The Magic Flute - Queen of the Night aria''}}
\end{itemize}
\section{Speed (S)}
\label{app:speed}
\textbf{Weight: 10\%}

The ability to perform simple cognitive tasks quickly. 

We decompose processing speed into ten distinct abilities, each contributing 1\% to the AGI score:
\begin{itemize}
    \item \textbf{Perceptual Speed–Search (1\%):} The speed of scanning a visual or textual field to find specific targets.
    \item \textbf{Perceptual Speed–Compare (1\%):} The speed of comparing two or more stimuli to identify similarities or differences.
    \item \textbf{Reading Speed (1\%):} The rate at which text can be processed with full comprehension.
    \item \textbf{Writing Speed (1\%):} The rate at which text can be generated or copied.
    \item \textbf{Number Facility (1\%):} The speed and accuracy of performing basic arithmetic operations.
    \item \textbf{Simple Reaction Time (1\%):} The time taken to respond to a single, anticipated stimulus.
    \item \textbf{Choice Reaction Time (1\%):} The time taken to respond correctly when presented with one of several possible stimuli.
    \item \textbf{Inspection Time (1\%):} The speed at which subtle differences between visual or auditory stimuli can be perceived.
    \item \textbf{Comparison Speed (1\%):} The time taken to make a judgment comparing two stimuli on a specific attribute (e.g., which is larger, brighter, or comes first alphabetically).
    \item \textbf{Pointer Fluency (1\%):} The speed and accuracy of moving a pointer, such as a virtual mouse.
\end{itemize}
\textit{Note: This is highly related to the broad CHC abilities ``Processing Speed (Gs),'' ``Reaction and Decision Speed (Gt),'' and to a lesser extent ``Psychomotor Speed (Gps).''}

Testing Methodology: For all speed tests, the AI's performance (latency or throughput) is compared against the average performance of a well-educated adult on the same tasks. The 1\% for each area is awarded if the AI meets or exceeds this human baseline. Crucially, artificial delays (e.g., excessive ``thinking'' time for simple tasks) count toward the time limit.

\subsection{Perceptual Speed–Search}
\textbf{Weight: 1\%}

The speed and fluency of searching or scanning an extended textual or visual field to locate one or more simple patterns.

\textit{Note: This is highly related to the narrow CHC ability ``Perceptual speed–search (Ps).''}

\textbf{Illustrative Examples:}
\begin{itemize}
    \item Scan of instances of ``a'' and ``t'' in text passages
    \item Scan for instances of $\in$ and $\yen$ in text made up of random symbols
    \item Return the pairs in this list of lists that sum to 10: [8, 1~~9, 3~~2, 7~~8, 2~~9, 6~~4, 6~~8, 0~~5, 5~~1, 9~~4, 5~~7, 2]
    \item Circle one bell in the following image: 
    
    \includegraphics[width=0.3\textwidth]{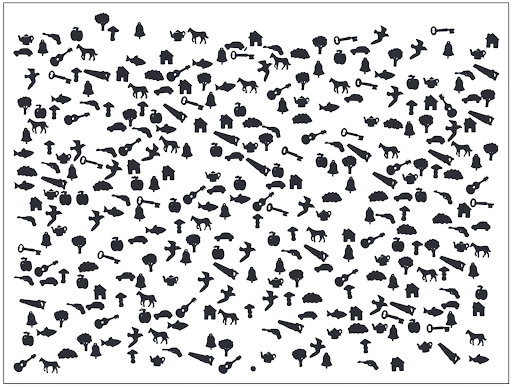}
\end{itemize}

\subsection{Perceptual Speed–Compare}
\textbf{Weight: 1\%}
The speed and fluency of looking up and comparing textual or visual stimuli that are side by side or more widely separated in an extended textual or visual field.

\textit{Note: This is highly related to the narrow CHC ability ``Perceptual speed–compare (Pc).''}

\textbf{Illustrative Examples:}
\begin{itemize}
    \item Determine mismatched name pairs in the 2 lists:
    
    [``Johnson'', ``Smith'', ``Garcia'', ``Miller'', ``Davis'']
    
    [``Johnson'', ``Smyth'', ``Garcia'', ``Millar'', ``Davis'']
    
    \item Find the largest number in ``48291, 93652, 12844, 59277''
    
    \item Identify the youngest person in the list:
    
    \begin{tabular}{lll}
    ID & Name & Date of Birth \\
    101 & Alice & 03/12/1992 \\
    102 & John & 07/25/1988 \\
    103 & Maria & 11/03/1990 \\
    104 & David & 05/19/1995
    \end{tabular}
\end{itemize}

\subsection{Reading Speed}
\textbf{Weight: 1\%}

Rate of reading text with full comprehension.

\textit{Note: This is highly related to the narrow CHC ability ``Reading speed (fluency) (RS).''}

\textbf{Illustrative Example:} Read along this textual passage for 60 seconds:\\ 
\href{https://docs.google.com/document/d/1ndBmyo7e4LOaPLvbv4i16t0LE2lbicK_OvargOXyjRE/edit?usp=sharing}{\docicon\ \underline{Reading/Writing Speed Example}}. \textit{60 seconds pass}. What are ``feelies''?

\textbf{Test:} Input text token processing speed gauges reading speed. Can measure the ``time-to-first-token'' latency of a 100 thousand token prompt.

\subsection{Writing Speed}
\textbf{Weight: 1\%}

The rate at which words or sentences can be generated or copied.

\textit{Note: This is highly related to the narrow CHC ability ``Writing speed (fluency) (WS).''}

\textbf{Illustrative Example:} In 60 seconds, please copy and output as much of the following textual passage as you can: \href{https://docs.google.com/document/d/1ndBmyo7e4LOaPLvbv4i16t0LE2lbicK_OvargOXyjRE/edit?usp=sharing}{\docicon\ \underline{Reading/Writing Speed Example}}.

\textbf{Test:} Output text token processing speed gauges writing speed.

\subsection{Number Facility}
\textbf{Weight: 1\%}

The rate at which basic arithmetic or algorithmic operations are performed accurately.

\textit{Note: This is highly related to the narrow CHC ability ``Number facility (N).''}

\textbf{Illustrative Examples:}
\begin{itemize}
    \item Compute $72/2$
    \item Compute $9 \times 10 \times 11$
    \item Sort from least to greatest: $37, 4, 92, 58, 13$
    \item Square each element: $9, 3, 1$
\end{itemize}

\subsection{Simple Reaction Time}
\textbf{Weight: 1\%}

Reaction time to the onset of a single stimulus (textual, visual, or auditory).

\textit{Note: This is highly related to the narrow CHC ability ``Simple RT (R1).''}

\textbf{Illustrative Examples:}
\begin{itemize}
    \item After reading this, immediately say `hello'.
    \item After reading this, immediately output the letter `a'.
    \item I'm going to speak. While I do so, briefly say `beep' immediately whenever you hear me use the letter `Q'. \href{https://drive.google.com/file/d/1Dds-Pip1Lgzhy5sKwd-eCekxySAJuIid/view?usp=sharing}{\videoicon\ \underline{letter\_sequence.m4a}}
    \item I will play a video. Whenever you see a blue flash, output the letter `g' as quickly as you can.
\end{itemize}

\subsection{Choice Reaction Time}
\textbf{Weight: 1\%}
Reaction time to the onset of one of several possible stimuli.

\textit{Note: This is highly related to the narrow CHC ability ``Choice RT (R2).''}

\textbf{Illustrative Examples:}
\begin{itemize}
    \item I’ll give a string of four characters, and you need to repeat the character that is capitalized as quickly as you can. ``a B c d''.
    \item As quickly as you can, respond with the character L if the arrow points left, R if right. Stimulus: $\to$
    \item As quickly as you can, respond with exactly the stimulus letter if it is one of A, E, I, or O. Otherwise, respond with X. Stimulus: E
    \item As quickly as you can, identify the color of the image: \includegraphics[width=0.05\textwidth]{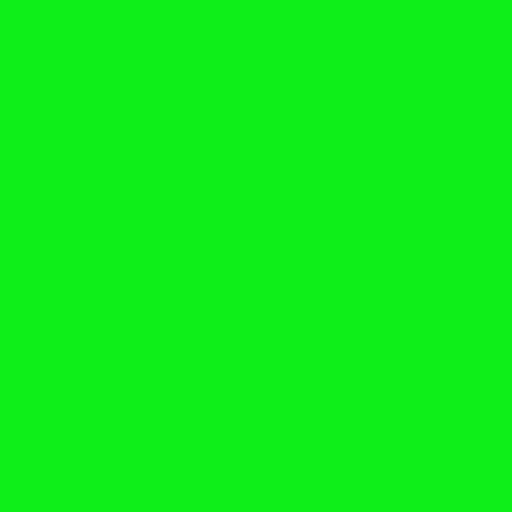}
\end{itemize}

\subsection{Inspection Time}
\textbf{Weight: 1\%}

The speed at which differences in visual stimuli can be perceived.

\textit{Note: This is highly related to the narrow CHC ability ``Inspection time (IT).''}

\textbf{Illustrative Examples:}
\begin{itemize}
    \item Here are two strings. They differ in exactly one character. What are the two letters that differ? ``QX9afGhtkLmN, QX9afGhtkLnN''
    \item Which image has a bell?

    A. \includegraphics[width=0.2\textwidth]{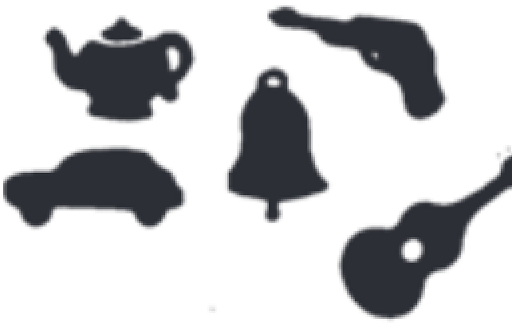}

    B. \includegraphics[width=0.2\textwidth]{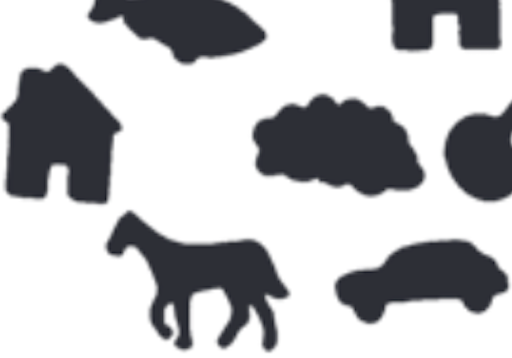}
    \item As quickly as you can, select which voice sounds the most typically feminine: \\ \href{https://drive.google.com/file/d/1OhHpxjVSM4PCMOOqQxdR-JXzvvJXxQ2O/view?usp=drive_link}{\audioicon\ \underline{voice\_1.wav}}, \href{https://drive.google.com/file/d/17-G1Zj9Rp7d_GM6WkpTA5BmSymYQn2l-/view?usp=drive_link}{\audioicon\ \underline{voice\_2.wav}}, \href{https://drive.google.com/file/d/1A5UrfWTPrpwRTFid4zcIPiFK2DQaefYu/view?usp=drive_link}{\audioicon\ \underline{voice\_3.wav}}.
\end{itemize}

\subsection{Comparison Speed}
\textbf{Weight: 1\%}

Reaction time where stimuli must be compared for a particular characteristic or attribute.

\textit{Note: This is highly related to the narrow CHC ability ``Mental comparison speed (R7).''}

\textbf{Illustrative Examples:}
\begin{itemize}
    \item I will give two numbers. As quickly as you can, answer yes or no to whether they have the same parity (both even or both odd): ``14, 9''
    \item I will give two numbers. As quickly as you can, indicate which number is larger: ``5.11, 5.9''
    \item I will give two words. As quickly as you can, answer which word comes first alphabetically: ``apple, apricot''
    \item I will give two words. As quickly as you can, answer which word has more vowels: ``reason, crypt''
\end{itemize}

\subsection{Pointer Fluency}
\textbf{Weight: 1\%}

The fluency of moving a computer mouse to execute simple requests.

\textit{Note: We do not assume the AGI must be embodied, so the AI can use a virtual mouse to complete this task, just as it uses virtual keys to write responses.}

\textit{Note: This is highly related to the narrow CHC ability ``Movement time (MT).''}

\textbf{Illustrative Examples:}
\begin{itemize}
    \item Using a mouse, draw as many roughly circular shapes as you can on a digital canvas using the pen feature. You have 30 seconds.
    \item Using a mouse, sketch a very rough outline of a T-Rex on a digital canvas using the pen feature.
    \item As quickly as you can, close all the tabs in this browser window, one by one, using the X button on the tabs.
\end{itemize}
Perceptual speed-search, perceptual speed-compare, reading speed, writing speed, number facility, simple reaction time, choice reaction time, inspection time, comparison speed, and movement speed.

\end{document}